\PassOptionsToPackage{table}{xcolor}
\documentclass[10pt,twocolumn,letterpaper]{article}

\usepackage{cvpr}              %

\definecolor{cvprblue}{rgb}{0.21,0.49,0.74}
\usepackage[pagebackref,breaklinks,colorlinks,allcolors=cvprblue]{hyperref}

\usepackage[accsupp]{axessibility}  %

\usepackage{graphicx}

\usepackage{multirow}
\usepackage{relsize}
\usepackage{xcolor}
\definecolor{ForestGreen}{RGB}{34,139,34}
\definecolor{apricot}{rgb}{0.98, 0.81, 0.69}
\definecolor{cadmiumorange}{rgb}{0.93, 0.53, 0.18}
\definecolor{burntorange}{rgb}{0.8, 0.33, 0.0}
\definecolor{brown(web)}{rgb}{0.65, 0.16, 0.16}
\definecolor{cornellred}{rgb}{0.7, 0.11, 0.11}
\definecolor{burgundy}{rgb}{0.5, 0.0, 0.13}
\definecolor{flame}{rgb}{0.89, 0.35, 0.13}
\definecolor{fireenginered}{rgb}{0.81, 0.09, 0.13}
\definecolor{firebrick}{rgb}{0.7, 0.13, 0.13}
\definecolor{mossgreen}{rgb}{0.68, 0.87, 0.68}
\definecolor{mountainmeadow}{rgb}{0.19, 0.73, 0.56}
\definecolor{olivine}{rgb}{0.6, 0.73, 0.45}
\definecolor{olivedrab}{rgb}{0.42, 0.56, 0.14}
\definecolor{oucrimsonred}{rgb}{0.6, 0.0, 0.0}
\definecolor{darkpastelpurple}{rgb}{0.59, 0.44, 0.84}
\definecolor{babyblueeyes}{rgb}{0.63, 0.79, 0.95}

\usepackage{chngcntr}

\usepackage{scalerel} %

\usepackage{listings}
\definecolor{darkblue}{RGB}{46,25, 110}

\newcommand{\edit}[1]{#1}
\newcommand{\dssectionheader}[1]{%
   \noindent\framebox[\columnwidth]{%
      {\fontfamily{phv}\selectfont \textbf{\textcolor{darkblue}{#1}}}
   }
}

\newcommand{\dsquestion}[1]{%
    {\noindent \fontfamily{phv}\selectfont \textcolor{darkblue}{\textbf{#1}}}
}

\newcommand{\dsquestionex}[2]{%
    {\noindent \fontfamily{phv}\selectfont \textcolor{darkblue}{\textbf{#1} #2}}
}

\def\paperID{8164} %
\def\confName{CVPR}
\def\confYear{2025}

\title{Unbiasing through Textual Descriptions: \\Mitigating Representation Bias in Video Benchmarks}

\author{
Nina Shvetsova$^{1,2,3}$\thanks{The work was done during PhD visit to the University of Oxford within the ELLIS PhD program.}, Arsha Nagrani$^{4}$, Bernt Schiele$^{3}$, Hilde Kuehne$^{1,2,5}$,  Christian Rupprecht$^{4}$\\
\small $^{1}$ Goethe University Frankfurt,
\small $^{2}$ Tuebingen AI Center/University of Tuebingen, 
\small $^{3}$ MPI for Informatics, Saarland Informatics Campus, \\
\small $^{4}$ University of Oxford,
$^{5}$ MIT-IBM Watson AI Lab \\ \small{
    \texttt{shvetsov@uni-frankfurt.de}} 
}

\begin{document}
\maketitle

\begin{abstract}

We propose a new ``Unbiased through Textual Description (UTD)'' video benchmark based on unbiased subsets of existing video classification and retrieval datasets to enable a more robust assessment of video understanding capabilities.
Namely, we tackle the problem that current video benchmarks may suffer from different representation biases, e.g., object bias or single-frame bias, where mere recognition of objects or utilization of only a single frame is sufficient for correct prediction. 
We leverage VLMs and LLMs to analyze and debias benchmarks from such representation biases. 
Specifically, we generate frame-wise textual descriptions of videos, filter them for specific information (e.g. only objects) and leverage them to examine representation biases across three dimensions: 1) concept bias\,---\,determining if a specific concept (e.g., objects) alone suffice for prediction; 2) temporal bias\,---\,assessing if temporal information contributes to prediction; and 3) common sense vs. dataset bias\,---\,evaluating whether zero-shot reasoning or dataset correlations contribute to prediction. 
We conduct a systematic analysis of 12 popular video classification and retrieval datasets and create new object-debiased test splits for these datasets. Moreover, we benchmark 30 state-of-the-art video models on original and debiased splits and analyze biases in the models. 
To facilitate the future development of more robust video understanding benchmarks and models, we release: ``UTD-descriptions'', a dataset with our rich structured descriptions for each dataset, and ``UTD-splits'', a dataset of object-debiased test splits.\footnote{Project page: \url{https://utd-project.github.io/} \\ To be published at CVPR 2025. 
When citing this work, please refer to the final version published in IEEE Xplore. Cite as:
Nina Shvetsova, Arsha Nagrani, Bernt Schiele, Hilde Kuehne, Christian Rupprecht. 
``Unbiasing through Textual Descriptions: Mitigating Representation Bias in Video Benchmarks''.
In: \textit{Proceedings of the IEEE/CVF conference on computer vision and pattern recognition}, 2025.
}

\end{abstract}
\section{Introduction}

\begin{figure}[t]
\begin{center}
\includegraphics[width=1\linewidth]{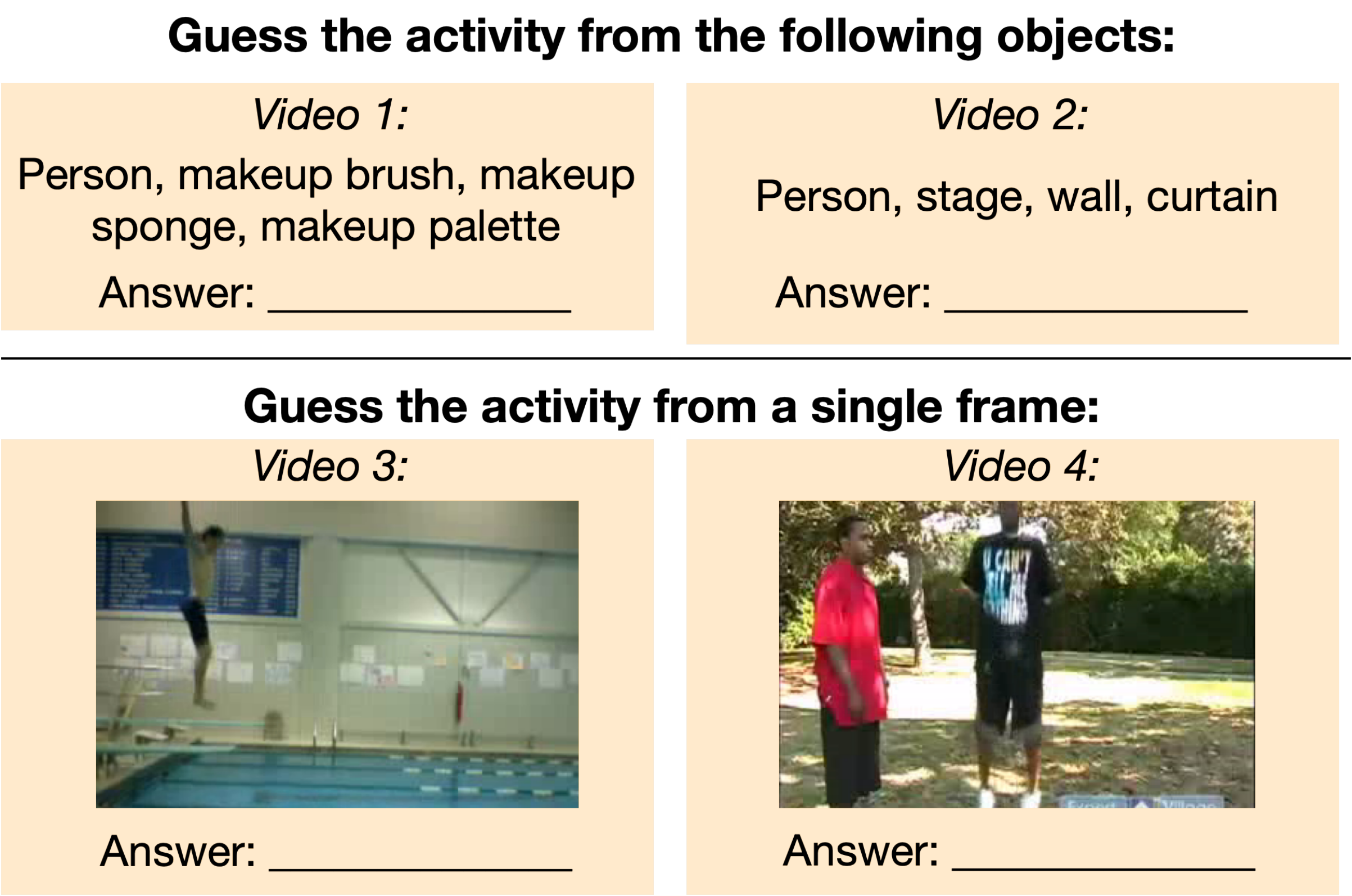}
\end{center}
\vspace{-0.5cm}
\caption{ \small{
Can you guess activity on a video based solely on the objects or a single frame? (The answer is on the next page in~\cref{fig:answer}.)
}
\label{fig:teaser}}
\vspace{-0.8cm}
\end{figure}

Ever since action recognition moved from simple, staged clip classification to real-life video understanding \cite{marszalek09Actions, SoomroUCF101, kuehne2011hmdb}, there has been an ongoing discussion about how much of a system's performance can be attributed to the recognition of the action itself compared to related or unrelated visual cues such as objects or scene information in the data~\cite{He2016Human,Chung2022HAT,Weinzaepfel2021Mimetics}, namely \textit{object/scene representation bias} (see~\cref{fig:teaser}). 
Different strategies have been proposed to separate action recognition from scene and object influences     
such as segmenting the human figure in a video and placing it against a gray or black background~\cite{He2016Human, Chung2022HAT} as well as miming actions in unrelated contexts~\cite{Weinzaepfel2021Mimetics}.
Some works focused on collecting datasets designed to minimize scene or object representation bias~\cite{Li_2018_RESOUND_ECCV,BenShabat2021IKEA,Goyal_2017_ICCV}.
However, debiased solutions have so far rarely been adopted for benchmarking. Reasons for that may be that modified videos in case of~\cite{He2016Human, Chung2022HAT} often introduce artifacts and benchmarks with explicitly low scene bias~\cite{Li_2018_RESOUND_ECCV,Goyal_2017_ICCV}, might be considered out-of-domain relative to standard training setups.
Another line of research questions whether current video benchmarks truly require temporal understanding, as single-frame models have shown comparable performance on several benchmarks~\cite{buch2022revisiting, lei2022revealing}. While new benchmarks are emerging to assess temporal understanding~\cite{bagad2023test, mangalam2023egoschema, cai2024temporalbench}, \textit{temporal representation bias} in standard benchmarks remains underexplored.

In this work, we propose a new scalable method, \textit{Unbiasing through Textual Descriptions} (\textit{UTD}), to systematically analyze and mitigate different types of representation bias, such as object bias, in existing video benchmarks through automatic textual descriptions. 
Since most datasets lack textual narrations of videos, we leverage Vision-Language Models (VLMs) to generate detailed descriptions of video frames. We then use Large Language Models (LLMs) to extract different concepts from those captions, namely \textit{objects}, \textit{activities}, and \textit{verbs} (an overview of this process is shown in~\cref{fig:method}).
Based on generated descriptions, we examine representation biases across three independent dimensions. First, \textit{concept representation bias}\,---\,assessing if a specific concept (e.g., objects) alone is sufficient for accurate prediction of the video. Second, \textit{temporal representation bias}\,---\,evaluating if temporal information is needed for correct prediction or if single-frame information is sufficient. 
Lastly, we distinguish between \textit{common sense} and \textit{dataset bias}. Common sense bias refers to the implicit association of certain concepts with specific activities, such as associating an object ``a piano'' with the label ``playing piano.'' This differs from dataset bias, where spurious correlations arise within the dataset, e.g., ``apply makeup'' label frequently occurs with objects like ``mirror'' and ``flowers'' in the background. We propose to leverage the zero-shot reasoning capabilities of LLMs to assess common sense bias and train a linear classifier for the dataset bias. 

In our study, we found that many samples of popular benchmarks have a significant \textit{object representation bias}, meaning that recognizing only objects on the video is already enough to predict the correct label. Therefore, using such data for benchmarking video backbones may give models with strong object recognition ability an advantage (i.e.~a shortcut), \emph{unintentionally} hindering video understanding evaluation with respect to temporal reasoning or action recognition. 
Therefore, by excluding test samples with the highest object bias, we introduce \textit{UTD-splits}, debiased evaluation splits for considered datasets, which we believe provide a more robust assessment of video understanding capabilities.
Importantly, compared to other approaches that rely on modifying videos~\cite{He2016Human, Chung2022HAT}, this approach neither requires any additional processing of evaluation data, as no new videos are generated nor does it require any fine-tuning to adjust for any domain bias that could be introduced by debiasing video frames at a pixel level.

\begin{figure}[t]
\begin{center}
\includegraphics[width=1\linewidth]{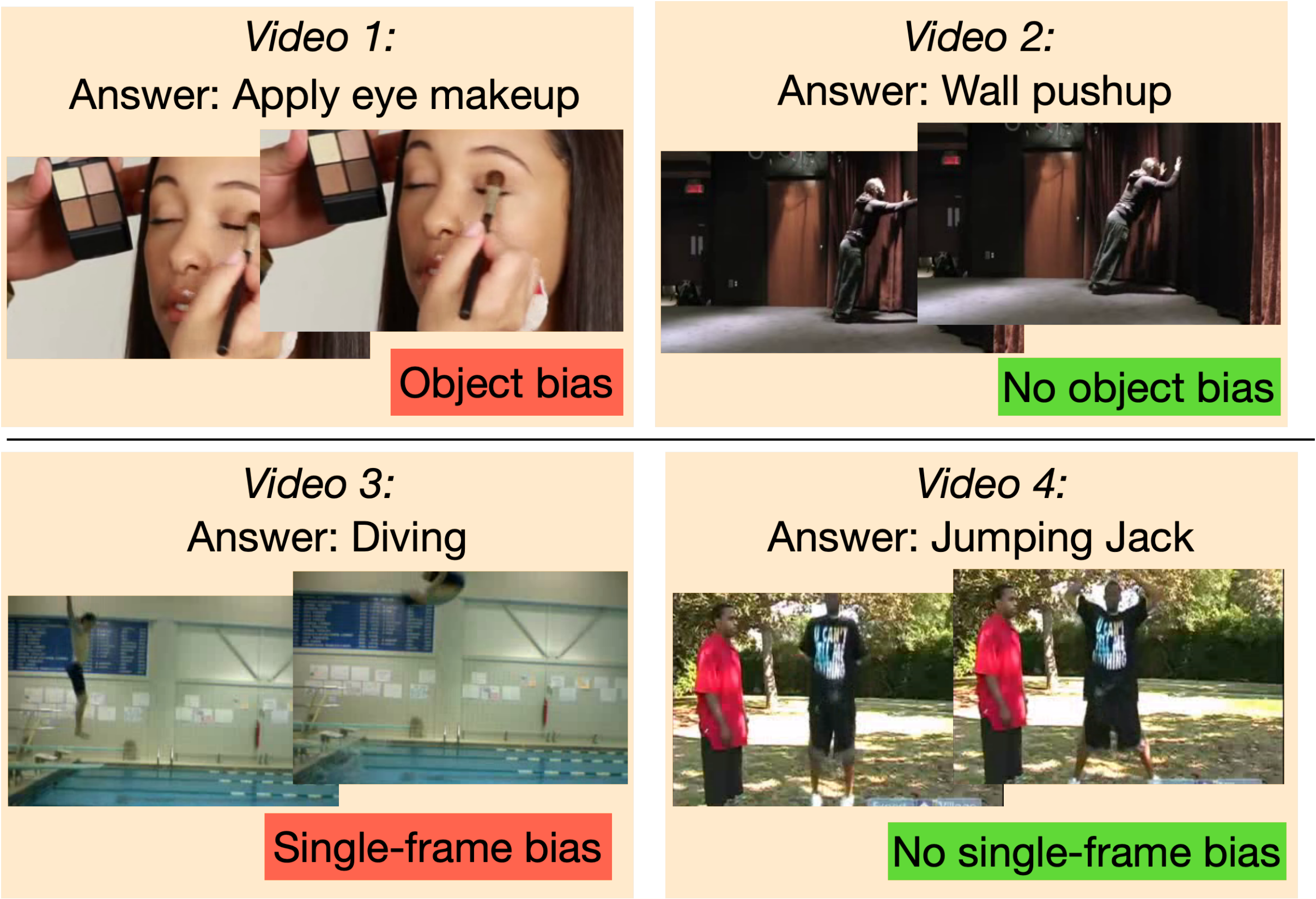}
\end{center}
\vspace{-0.5cm}
\caption{ \small{
The answer to~\cref{fig:teaser}. Some videos may exhibit an object representation bias (allowing predictions based solely on objects) or single-frame representation bias (solely on a single frame), while others require more information for prediction.
}
\label{fig:answer}}
\vspace{-0.5cm}
\end{figure}

\begin{figure*}[t]
\begin{center}
\includegraphics[width=1\linewidth]{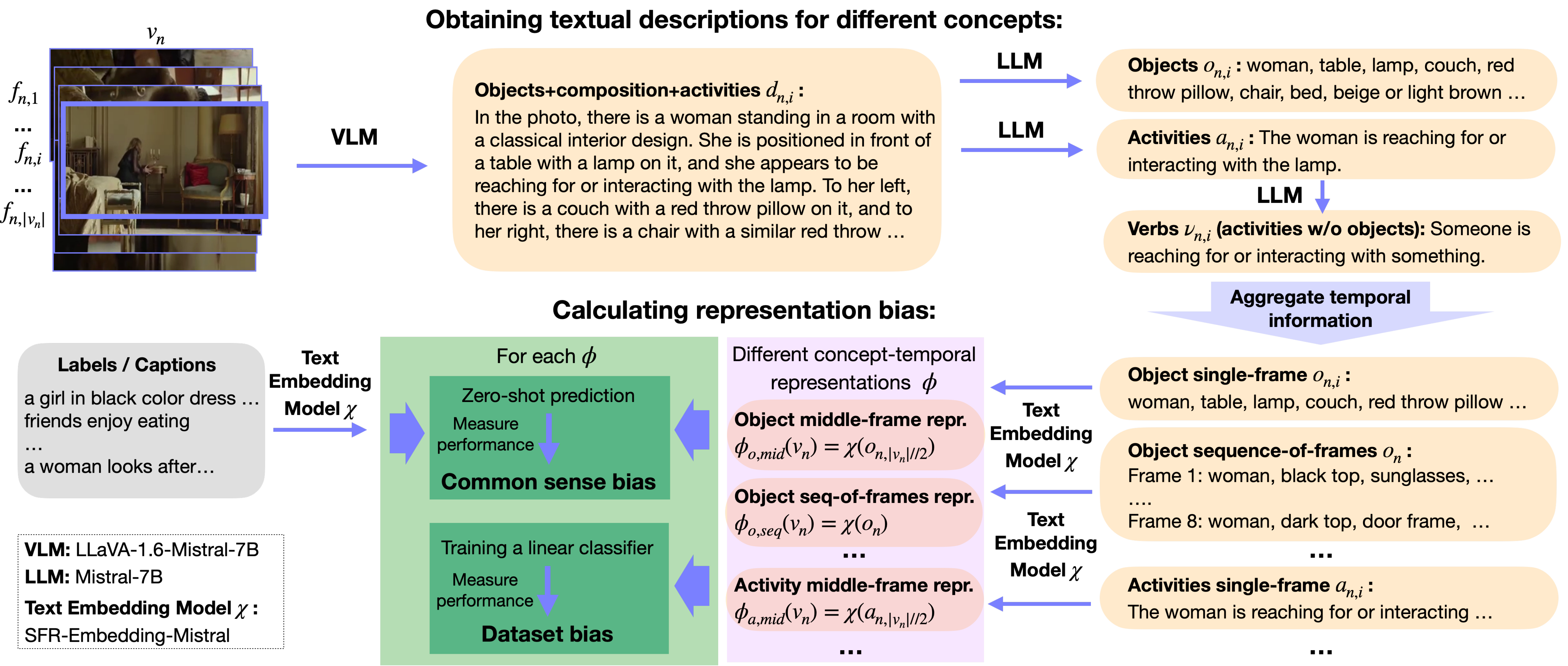}
\end{center}
\vspace{-0.5cm}
\caption{ \small{
\textbf{The proposed UTD method} involves generating textual descriptions of different concepts in video frames using VLMs and LLMs, combining them in various temporal configurations, and evaluating the performance of these concept-temporal representations with strong text embedding models. For each representation, we distinguish between common sense bias, which relies on zero-shot reasoning by text embedding models, and dataset bias, assessed using a linear model trained on the dataset’s training set.
\label{fig:method}
}}
\vspace{-0.5cm}
\end{figure*}

We apply the proposed UTD method to analyze and debias six action classification datasets, namely UCF101~\cite{SoomroUCF101}, SomethingSomethingv2~\cite{Goyal_2017_ICCV}, Kinetics-400~\cite{Kay2017Kinetics}, -600~\cite{Carreira2018Kinetics600}, and -700~\cite{Carreira2019Kinetics700}, and Moments In Time~\cite{monfort2019moments}, and six text-to-video retrieval datasets, namely MSRVTT~\cite{xu2016msr}, DiDeMo~\cite{anne2017localizing}, ActivityNet~\cite{Heilbron_2015_CVPR}, LSMDC~\cite{Rohrbach_2015_CVPR}, YouCook2~\cite{Das_2013_YouCook}, and Spoken Moments In Time~\cite{Monfort_2021_CVPR}. 
Further, we benchmark 30 state-of-the-art video models of six different model designs, namely VideoMAE~\cite{tong2022videomae}, VideoMAE2~\cite{wang2023videomae}, All-in-one~\cite{wang2023all}, Unmasked Teacher~\cite{li2023unmasked}, VideoMamba~\cite{li2024videomamba}, and InternVid~\cite{wang2023internvid}, on the original and UTD-debiased splits. 
We show that debiasing reduces performance saturation on current benchmarks. Considering model performance drops between original and debiased splits also allows us to analyze models on object bias, e.g., in action classification, larger backbones or those trained on more data appear more robust to object bias.

We summarize the contributions of this work as follows:  
(1) We introduce \textit{Unbiasing through Textual Descriptions (UTD)}, a scalable, automated method to analyze and mitigate representation biases in video benchmarks without the need for human annotators in the loop.
(2) Using UTD, we analyze representation biases across 12 popular classification and retrieval video benchmarks and release UTD-debiased test splits for each, \textit{UTD-splits}, promoting a more robust evaluation of video understanding.
(3) We benchmark 30 state-of-the-art video models on both the original and UTD-debiased splits to assess their robustness against object bias under varying conditions. 
(4) Finally, we release \textit{UTD-descriptions}, a dataset of detailed structured descriptions (including objects, activities, and verb categories) for all 12 considered datasets, to support further dataset analysis and the development of less biased benchmarks.

\section{Related Work}

\noindent\textbf{Debiasing Video Datasets.}
While most visual datasets are prone to various biases~\cite{FABBRIZZI2022survey}, one unique effect in video understanding is the impact of object and scene information on the overall performance. 
Different strategies have so far been explored to mitigate this bias.
He et al.~\cite{He2016Human} and HAT~\cite{Chung2022HAT} proposed to separate the performance of scene appearance from human motion by segmenting human figures and placing them into black/gray background, or alternatively by cropping or inpainting human figures and classifying videos based on background only. 
Weinzaepfel et al.~\cite{Weinzaepfel2021Mimetics} proposed the Mimetics dataset by extending certain classes from the Kinetics dataset \cite{Kay2017Kinetics} with videos of activities in unrelated background.
Zhai et al. \cite{Zhai_2023_SOAR_ICCV} proposed an approach for scene-debiased action recognition via adversarial scene reconstruction and classification. 
Another line of work focused on capturing datasets with explicitly low object and scene bias. SomethingSomethingv1 and later v2 dataset~\cite{Goyal_2017_ICCV} were proposed as an almost scene-free dataset, focusing on simple manipulative actions only. 
Li et al. \cite{Li_2018_RESOUND_ECCV} proposed a procedure for designing representation unbiased datasets and introduced Diving48 as an example.
However, this specialization usually also implies that those cases are rather suitable for specialized approaches but not for, e.g., zero-shot evaluation of generalist video models. 
Other studies~\cite{buch2022revisiting, lei2022revealing} showed that for several benchmarks, using single frames during training or inference can achieve performance comparable to models that incorporate temporal information. However, recent efforts have mainly focused on developing new benchmarks to assess temporal understanding in videos~\cite{bagad2023test}, e.g., 
visual question-answering benchmarks~\cite{mangalam2023egoschema, cai2024temporalbench}, while temporal representation bias in standard benchmarks has not been thoroughly investigated.
In contrast, our UTD method allows for a systematic analysis of representation biases across various dimensions, including different concept and temporal configurations. Additionally, we are the first to propose mitigating object representation bias in popular benchmarks by discarding object-biased samples, which helps us to avoid adding artifacts to test samples (as with cropping or inpainting in~\cite{He2016Human, Chung2022HAT}) and keeps the dataset domain unconstrained, unlike~\cite{Goyal_2017_ICCV, Li_2018_RESOUND_ECCV}.

\noindent\textbf{LLMs for Video Understanding.}
The progress of LLMs prompted a wide range of applications in various fields, including video understanding. Here, one line of work utilizes pre-trained LLMs in video-language tasks, such as video captioning~\cite{sun2019videobert,zhao2024videoprism,wang2023gpt4video}, video question answering~\cite{zhao2024videoprism,li2023unmasked,Maaz2023VideoChatGPT,wang2023gpt4video}, text-video retrieval~\cite{zhao2024videoprism,shvetsova2023style}, or dialog-based user interaction~\cite{wang2023gpt4video,damonlpsg2023videollama}. 
Some works~\cite{wang2023internvid,chen2024vast,shvetsova2023howtocaption} create new video-language datasets using image/audio captioning models or subtitles and using LLM summarization and reasoning ability to post-process these captions. Other works~\cite{zeng2022socratic,2023videochat,wang2022language,wang2023chatvideo} 
use expert perception models
to extract information from the video that is later processed by the LLM for tasks such as video captioning.
In our work, we leverage VLM- and LLM-generated frame descriptions to evaluate different representation biases in video datasets.

\section{Discovering Representation Biases}
\label{sec:method}

In this section, we define representation bias in~\cref{sec:preliminaries}, outline the considered representation biases in~\cref{sec:RepresentationBiases}, and detail our UTD method in~\cref{sec:utd}.

\subsection{Definition: Representation Bias}
\label{sec:preliminaries}

To formalize the definition of representation bias in datasets, we adopt the notation from RESOUND~\cite{Li_2018_RESOUND_ECCV}. 

Let $D = \{v_{n}\}_{n \leq N}$ be a video dataset $D$, where each video $v_{n}$ consists of a sequence of frames $v_{n} =\{f_{n,i}\}_{ i \leq |v_n|}$. A representation $\phi$ is a function $\phi(v)$ of the video input $v$. Let $R = \{\phi\}$ be a family of representations that share specific characteristics. For example, a family of single-frame representations includes representations that encode information from only a single frame of a video $\phi(v_n) = \phi(f_{n,i})$ where ${i \leq |v_n|}$ or a family of object-based representations encodes only the objects in the video without considering motion or activity information.

The representation $\phi$ can be used to train different models $\gamma_{\phi}$ for desired tasks, e.g. video classification. Let $M(v_n, \gamma_{\phi})$ be the performance (e.g., accuracy) of a model $\gamma_{\phi}$ on a video $v_n$, then the performance of $\gamma_{\phi}$ on $D$ is 
\begin{equation}
\label{eq:performance}
    M(D, \gamma_{\phi}) = \frac{1}{N}\sum_{n=1}^{N} M(v_n, \gamma_{\phi}).
\end{equation}
We define the performance of a representation $\phi$ on $D$ as 
\begin{equation}
\label{eq:bias}
    M(D, \phi) = \max_{\gamma_{\phi}} M(D, \gamma_{\phi}) ,
\end{equation}
where the maximum is taken over all models $\gamma_{\phi}$ that are based on the representation $\phi$.

When the performance $M(D, \phi)$ of a representation $\phi$ is high, the dataset exhibits a preference for this representation, indicating that the dataset is more biased toward this representation. In this work, we use $M(D, \phi)$ as a measure of bias towards representation $\phi$. However, a high bias toward a specific representation is not necessarily a negative thing. Ideally, a dataset should exhibit the highest bias toward the \textit{``ground truth representation''}: the representation truly required to solve the task. For instance, in video understanding, one could argue that the optimal representation should capture both static (e.g., objects) and dynamic (e.g., actions) properties, as well as encode temporal aspects.
However, if a video dataset used to evaluate video understanding exhibits a strong object or single-frame representation bias, it may unintentionally favor models with strong object recognition or image understanding capabilities, thus hindering the evaluation of true video understanding.
Therefore, in this work, we aim to estimate representation biases in popular video benchmarks. 

\subsection{Different Representation Biases}
\label{sec:RepresentationBiases}

In our work, we consider different types of representation biases with respect to three independent aspects:

\noindent\textbf{Concept Representations:} First, we consider representations that encode four different concept categories: \textit{objects} $\phi_o(v_n)$, \textit{activities} $\phi_a(v_n)$  (actions along with tied objects), \textit{verbs} $\phi_\nu(v_n)$ (actions alone, without objects), and \textit{objects+composition+activities} $\phi_d(v_n)$ (objects, their spatial compositions in the video, and activities), see an example in~\cref{fig:method}. 
We aim to assess biases in classification and retrieval benchmarks toward particular concepts. For example, in retrieval, we ideally would like to see that benchmarks are not biased solely toward objects or activities.

\noindent\textbf{Temporal Representations:} Second, we consider the bias of single-frame representations compared to representations that consider multiple frames. Let $\phi_x$ be one of the objects $\phi_o$, activities  $\phi_a$, verbs $\phi_\nu$, or objects+composition+activities  $\phi_d$ representations. Similarly to~\cite{Wang2016TSN}, we consider the following options to encode information from multiple frames: \textit{middle-frame} refers to the case that a video is represented only by a single middle frame of this video, e.g., $\phi_{x,\mathrm{mid}}(v_{n})=\phi_x(f_{n,|v_{n}|//2})$, \textit{max-score frame} refers to the case that a video is represented using a single frame $\phi_{x,\mathrm{max}}(v_{n})=\phi_x(f_{n,i})$ where the frame $f_{n,i}$ has the highest performance score $M(f_{n,i}, \phi_x)$ over all frames,
\textit{average-over-frames}  refers to the case that the video is represented as an average representation of frames ignoring the order of frames $\phi_{x,\mathrm{avg}}(v_{n})=\frac{1}{|v_n|}\sum_{i \leq |v_n|} {\phi_x(f_{n, i})}$, and, finally, \textit{sequence-of-frames}  refers to the case that the video representation encodes concepts and their order in the frames $\phi_{x,\mathrm{seq}}(v_{n})=\phi_x(f_{n,1}, f_{n,2}, .., f_{n,|v_n|})$.

\noindent\textbf{Common Sense Bias vs. Dataset Bias:}
Finally, for any concept-temporal representation $\phi$ (e.g., object middle-frame representation $\phi_{o, mid}$), we can consider two types of representation biases: \textit{common sense bias} or \textit{dataset bias}. Common sense representation bias refers to bias based on ``common sense reasoning'', e.g., a video featuring an object ``piano'' would most likely be classified under the ``playing piano'' activity label. Formally, in~\cref{eq:bias}, this bias $M(D, \phi)$ is measured by taking the maximum performance only over zero-shot models $\gamma_{\phi}$ that have not seen the correlations in the training set of $D$. We distinguish this from \textit{dataset bias}, which refers to biases caused by spurious correlations within the dataset, e.g., where certain objects are unintentionally associated with specific labels, such as ``mirror'' or ``flower bouquet'' frequently appear with the ``apply makeup'' label. In this case, in~\cref{eq:bias}, we measure dataset bias by taking the maximum performance over models $\gamma_{\phi}$ trained or fine-tuned on the training set of $D$.

We note that the concept, temporal, and common sense vs. dataset bias dimensions are independent, allowing us to combine them in any configuration. For example, we can evaluate object middle-frame common sense bias or activity sequence-of-frames dataset bias.

\subsection{Unbiasing through Textual Descriptions (UTD)}
\label{sec:utd}

Since measuring bias for every representation within a family (e.g., a family of object representations) is infeasible, we approximate it by considering a single representative from the family~\cite{Li_2018_RESOUND_ECCV} and propose a way to precisely control what each representation encodes by using textual descriptions.

The key idea of our method is to use textual descriptions as an intermediate video representation. This approach allows us to easily control which concepts in the video are included or excluded in the representation and to combine information into different temporal representations. Namely, to obtain representation $\phi$ that encodes only a specific concept category, we propose to encode visible concepts of this category into textual descriptions, e.g., encode objects with a list of visible objects, e.g., \textit{``table, lamp, ..''}. This approach, compared to cropping or inpainting~\cite{He2016Human, Chung2022HAT}, allows us to isolate specific concepts more precisely, avoiding information leakage from cropped areas or adding artifacts caused by imperfect inpainting. 
Additionally, textual descriptions allow us to use off-the-shelf LLMs with strong common sense reasoning abilities to estimate common sense bias. Since metadata with visible objects, activities, etc., is usually not available, we propose using recent powerful VLMs and LLMs to extract this information.

In the next sections, we first describe our method of obtaining textual description with respect to different concept and temporal categories (\cref{sec:Method_ObtainTextDesc}), then we describe our method of discovering common sense and dataset biases (\cref{sec:Method_RepresentationBiases}), and lastly, we describe our method of unbiasing benchmarks (\cref{sec:unbiasing_datasets}).

\subsubsection{Obtaining Textual Descriptions} 
\label{sec:Method_ObtainTextDesc}

While current VLMs
~\cite{liu2024llavanext,liu2023improvedllava,liu2023llava}, trained to perform complex image reasoning, are able to generate detailed descriptions of video frames, they still face challenges 
when prompted to directly extract specific concepts from a frame, e.g., only visible objects.
Therefore, we decomposed the task of concept extraction into two parts: extraction of detailed frame description using VLMs and then extraction of different concepts from frame description using LLMs (see~\cref{fig:method}).

\noindent\textbf{Obtaining Textual Descriptions Using VLMs.} 
Given train/test videos $v_{n} = \{f_{n,i}\}_{ i \leq |v_n|}$, $n < N$ of the dataset, we feed each frame $f_{n,i}$ of each video $v_{n}$ into the VLM model $d$ to generate a detailed textual description $d_{n,i} = d(f_{n_i})$ (\cref{fig:method}). We found that without complex prompt engineering or few-shot in-context learning strategies~\cite{brown2020language}
VLMs produce long, detailed descriptions. 
While our method is not restricted to a specific VLM, for reproducibility and feasibility, we select an open-sourced strong middle-scale VLM model, LLaVA-1.6-Mistral-7B~\cite{liu2024llavanext} with the prompt: \textit{``Describe the objects' relationships in the photo.'}' Since the resulting textual descriptions include objects, their composition, relationships, and visible activities, we use them as \textit{objects+composition+activities} descriptions.

\noindent\textbf{Extracting Concepts Categories Using LLMs.}
Each of the obtained detailed descriptions $d_{n,i}$ is further fed to the LLM $\zeta$ to extract only \textit{objects} $o_{n,i} = \zeta(d_{n,i}, p_o)$, 
only \textit{activities}  $a_{n,i} = \zeta(d_{n,i}, p_a)$ information, with $p_o$, $p_a$ as respective prompts. 
Then, to obtain information about \textit{verbs}, we task the LLM with a prompt $p_{\nu}$ to delete all object information from activities  $\nu_{n,i} =\zeta(a_{n,i}, p_{\nu})$.
Since such text processing requires understanding what objects and activities are, our prompt includes a few examples following a few-shot in-context learning strategy~\cite{brown2020language}.  We provide all prompts in the supplementary material.

\noindent\textbf{Aggregating Temporal Information.}
We use the obtained textual descriptions $x_{n,i}$ of video $v_{n}$ (where $x_{n,i}$ is one of $o_{n,i}$  $a_{n,i}$, $\nu_{n,i}$, or $d_{n,i}$) in different temporal setups: first, we keep them as single frame descriptions that might be further aggregated when transformed into vector representations (e.g. \textit{average-over-frames}), as we describe in~\cref{sec:Method_RepresentationBiases}, second, we combine them into \textit{sequence-over-frame} $x_n$ setup with the template: \textit{``Frame 1: $x_{n,1}$ Frame 2: $x_{n,2}$ ... Frame $|v_n|$: $x_{n,|v_n|}$''}  (\cref{fig:method}). 
Since \textit{objects+composition+activities} descriptions $d_{n,i}$ when combined with sequence-over-frame templates become too long, we summarize them to 15-word descriptions $d'_{n,i}$ via an LLM before inputting them into the sequence prompt.

\subsubsection{Calculating Representation Biases}
\label{sec:Method_RepresentationBiases}

After obtaining a specific concept-temporal representation $\phi$, we assess common sense and dataset bias for dataset $D$ toward $\phi$ by measuring the performance $M(D, \phi)$, as stated ~\cref{eq:bias}, where maximum is taken over zero-shot models $\gamma_{\phi}$ for \textit{common sense} bias and over fine-tuned models $\gamma_{\phi}$ for \textit{dataset bias} (see ~\cref{sec:RepresentationBiases}).
Since maximizing over all models is infeasible in practice, we design strong models for common sense and dataset-based reasoning by utilizing state-of-the-art LLMs. Our ablation study verifies our design choice (see supplementary material).

\noindent\textbf{Common Sense Bias.} We leverage recent strong LLM-based text embedding models~\cite{wang2023improving, meng2024sfrembedding}. Such models are trained on large datasets to encode text into vector representations that can be used for various tasks, such as information retrieval. We encode our textual descriptions and all class labels/captions using the pre-trained text embedding model $\chi$.  Specifically, we prompt the model with the relevant concept and task. For example, we use prompts like \textit{``Given a list of objects visible on the video frame, retrieve the activity depicted in this video'' } for object description and  \textit{``Given an activity, retrieve a video frame description that may depict this activity''} for class labels in action classification.  We use cosine similarity between obtained embeddings for zero-shot classification/retrieval. Prompts are detailed in the supplement. 
 
 \begin{table*}[]
    
    \centering
    \resizebox{1\linewidth}{!}{
    
    \begin{tabular}{@{}l|ccc ccc | ccc ccc@{}}
    	\toprule
                & \cellcolor{darkpastelpurple!40}UCF & \cellcolor{darkpastelpurple!40}SSv2 & \cellcolor{darkpastelpurple!40}K400 & \cellcolor{darkpastelpurple!40}K600 & \cellcolor{darkpastelpurple!40}K700 & \cellcolor{darkpastelpurple!40}MiT & \cellcolor{babyblueeyes!50}MSR & \cellcolor{babyblueeyes!50}DDM & \cellcolor{babyblueeyes!50}ActN & \cellcolor{babyblueeyes!50}LSMDC & \cellcolor{babyblueeyes!50}YC2 & \cellcolor{babyblueeyes!50}S-MiT \\
                \midrule

obj+comp+act &66.3 & 6.4 & 48.0 & 44.1 & 39.0 & 22.6 & 36.6 & 27.2 & 26.6 & 17.0 & 8.4 & 45.9  \\
objects & \cellcolor{apricot!50}$63.3_{\scaleto{-3.0}{6pt}}$ & \cellcolor{apricot!50}$5.3_{\scaleto{-1.1}{6pt}}$ & \cellcolor{apricot!50}$45.9_{\scaleto{-2.1}{6pt}}$ & \cellcolor{apricot!50}$41.8_{\scaleto{-2.3}{6pt}}$ & \cellcolor{apricot!50}$37.0_{\scaleto{-2.0}{6pt}}$ & \cellcolor{apricot!50}$21.0_{\scaleto{-1.6}{6pt}}$ & \cellcolor{apricot!50}$32.1_{\scaleto{-4.5}{6pt}}$ & \cellcolor{apricot!50}$27.0_{\scaleto{-0.2}{6pt}}$ & \cellcolor{apricot!50}$24.8_{\scaleto{-1.8}{6pt}}$ & \cellcolor{apricot!50}$13.6_{\scaleto{-3.4}{6pt}}$ & \cellcolor{apricot!50}$7.9_{\scaleto{-0.5}{6pt}}$ & \cellcolor{oucrimsonred!60}$29.8_{\scaleto{-16.1}{6pt}}$ \\
activities & \cellcolor{mossgreen!50}$67.4_{\scaleto{+1.1}{6pt}}$ & 6.4 & \cellcolor{apricot!50}$45.2_{\scaleto{-2.8}{6pt}}$ & \cellcolor{apricot!50}$41.4_{\scaleto{-2.7}{6pt}}$ & \cellcolor{apricot!50}$36.7_{\scaleto{-2.3}{6pt}}$ & \cellcolor{apricot!50}$21.0_{\scaleto{-1.6}{6pt}}$ & \cellcolor{red!50}$25.1_{\scaleto{-11.5}{6pt}}$ & \cellcolor{cadmiumorange!50}$21.1_{\scaleto{-6.1}{6pt}}$ & \cellcolor{cadmiumorange!50}$21.4_{\scaleto{-5.2}{6pt}}$ & \cellcolor{apricot!50}$14.7_{\scaleto{-2.3}{6pt}}$ & \cellcolor{apricot!50}$8.1_{\scaleto{-0.3}{6pt}}$ & \cellcolor{apricot!50}$41.1_{\scaleto{-4.8}{6pt}}$ \\
verbs & \cellcolor{oucrimsonred!60}$50.8_{\scaleto{-15.5}{6pt}}$ & \cellcolor{apricot!50}$5.8_{\scaleto{-0.6}{6pt}}$ & \cellcolor{oucrimsonred!60}$24.8_{\scaleto{-23.2}{6pt}}$ & \cellcolor{oucrimsonred!60}$21.4_{\scaleto{-22.7}{6pt}}$ & \cellcolor{oucrimsonred!60}$17.6_{\scaleto{-21.4}{6pt}}$ & \cellcolor{cadmiumorange!50}$16.2_{\scaleto{-6.4}{6pt}}$ & \cellcolor{oucrimsonred!60}$10.5_{\scaleto{-26.1}{6pt}}$ & \cellcolor{oucrimsonred!60}$7.0_{\scaleto{-20.2}{6pt}}$ & \cellcolor{oucrimsonred!60}$7.4_{\scaleto{-19.2}{6pt}}$ & \cellcolor{red!50}$5.5_{\scaleto{-11.5}{6pt}}$ & \cellcolor{cadmiumorange!50}$1.2_{\scaleto{-7.2}{6pt}}$ & \cellcolor{oucrimsonred!60}$13.1_{\scaleto{-32.8}{6pt}}$\\

                \bottomrule
    \end{tabular}
    }
    \vspace{-0.2cm}
    \caption{\small{
    \textbf{Evaluation of concept representation bias.} We report the performance of our model in \textit{sequence-of-frames} temporal setup and \textit{common sense bias} setup and color code with respect to the difference to \textit{objects+composition+activities (obj+comp+act)} concepts. It shows that the overall classification performance drops only slightly when classifying actions based on objects, compared to classification and retrieval on verbs only, with SSv2 as the only dataset with less drop when classifying on verbs only.
    \label{tab:all_biases}
    }}
    \vspace{-0.2cm}
\end{table*} 
 
\noindent\textbf{Dataset Bias.} For action classification datasets, we evaluate dataset bias, as action classification models are often tested in fine-tuned setups and might exploit dataset representation biases learned from training data. For this we a linear logistic regression model trained on our strong text embeddings used in the common sense bias analysis and train model on the official train splits of the datasets. 

\noindent\textbf{Metrics.}  For the performance measure $M(D, \phi)$, we use accuracy for action classification and recall@1 and text-to-video retrieval as the most popular metrics for these tasks.

\subsubsection{Unbiasing Datasets} 
\label{sec:unbiasing_datasets}

In~\cref{sec:benchmarking_datasets}, we find that many video datasets exhibit a significant object representation bias. 
To support robust video model evaluation, we construct \textit{UTD-splits}, object-debiased test splits, by excluding object-biased samples from the original test sets. 
Since our method allows us to measure the performance of the representation $M(v_n, \phi)$ on individual samples $v_n$, we exclude samples that are classified or retrieved correctly ($M(v_n, \phi) = 1$) based on \textit{object sequence-of-frame} representation.
For text-to-video retrieval, we debias with respect to \textit{common-sense bias}, but for action classification, we debias with respect to \textit{dataset bias} since video models are mostly tested in fine-tuned setup.
To minimize the impact of random fluctuations in the process, we use three different prompts for labels/captions in the text embedding models and train three linear models for each classification dataset using bootstrapped training sets (we provide details and analysis in the supplement) and exclude only samples when all nine models agreed on correct classification/retrieval. Note that the percentage of removed samples is determined automatically based on the extent of object representation bias.

Finally, since debiasing may disproportionately remove certain label classes in classification datasets, we additionally construct balanced UTD splits. While maintaining the total number of removed samples as in the original debiasing method, we adjust the number of samples removed from each class based on their average confidence (across nine models) to preserve the original class proportions.

\section{Benchmarking Datasets}
\label{sec:benchmarking_datasets}

\begin{table*}[]
    \renewcommand*{\arraystretch}{1.2}
    \setlength{\tabcolsep}{1pt}
    
    \centering
    \resizebox{1\linewidth}{!}{
    \begin{tabular}{@{}l|ccc |ccc | ccc | ccc | ccc | ccc@{}}
    	\toprule
                & \multicolumn{3}{c}{UCF} & \multicolumn{3}{c}{SSv2} & \multicolumn{3}{c}{K400} & \multicolumn{3}{c}{K600} & \multicolumn{3}{c}{K700} & \multicolumn{3}{c}{MiT} \\
                & obj. & act. & verbs & obj. & act. & verbs& obj. & act. & verbs& obj. & act. & verbs& obj. & act. & verbs& obj. & act. & verbs \\
                \midrule

common sense b. & 63.3 & 67.4 & 50.8 & 5.3 & 6.4 & 5.8 & 45.9 & 45.2 & 24.8 & 41.8 & 41.4 & 21.4 & 37.0 & 36.7 & 17.6 & 21.0 & 21.0 & 16.2 \\
dataset bias & \cellcolor{ForestGreen!50}$80.3\scaleto{+17.0}{6pt}$ & \cellcolor{ForestGreen!50}$83.4\scaleto{+16.0}{6pt}$ & \cellcolor{ForestGreen!50}$73.5\scaleto{+22.7}{6pt}$ & \cellcolor{olivine!50}$13.6\scaleto{+8.3}{6pt}$ & \cellcolor{olivine!50}$16.0\scaleto{+9.6}{6pt}$ & \cellcolor{olivine!50}$12.2\scaleto{+6.4}{6pt}$ & \cellcolor{olivedrab!50}$60.7\scaleto{+14.8}{6pt}$ & \cellcolor{ForestGreen!50}$60.7\scaleto{+15.5}{6pt}$ & \cellcolor{ForestGreen!50}$42.2\scaleto{+17.4}{6pt}$ & \cellcolor{ForestGreen!50}$60.4\scaleto{+18.6}{6pt}$ & \cellcolor{ForestGreen!50}$60.3\scaleto{+18.9}{6pt}$ & \cellcolor{ForestGreen!50}$42.5\scaleto{+21.1}{6pt}$ & \cellcolor{ForestGreen!50}$52.3\scaleto{+15.3}{6pt}$ & \cellcolor{olivedrab!50}$49.7\scaleto{+13.0}{6pt}$ & \cellcolor{ForestGreen!50}$34.9\scaleto{+17.3}{6pt}$ & \cellcolor{olivine!50}$27.8\scaleto{+6.8}{6pt}$ & \cellcolor{olivine!50}$27.3\scaleto{+6.3}{6pt}$ & \cellcolor{olivine!50}$22.9\scaleto{+6.7}{6pt}$ \\

                \bottomrule
    \end{tabular}
     }
      \vspace{-0.2cm}
    \caption{\small{\textbf{Comparison of common sense bias vs dataset bias.} We report the performance of our model in \textit{sequence of frame} temporal setup and color code with respect to difference to \textit{common sense bias}. It shows that the overall performance growth is significant for all concepts, including object concepts, indicating that datasets might contain a significant correlation between labels and visible objects. 
        \textbf{
        \label{tab:dataset_bias}
        }}}
    \vspace{-0.5cm}
    
\end{table*} 
\begin{table}[]
    \centering
    \renewcommand*{\arraystretch}{1.2}
    \setlength{\tabcolsep}{1pt}
    
    \resizebox{1\linewidth}{!}{
    \begin{tabular}{@{}l|ccc ccc | ccc ccc@{}}
    	\toprule
                &\cellcolor{darkpastelpurple!40}UCF & \cellcolor{darkpastelpurple!40}SSv2 & \cellcolor{darkpastelpurple!40}K400 & \cellcolor{darkpastelpurple!40}K600 & \cellcolor{darkpastelpurple!40}K700 & \cellcolor{darkpastelpurple!40}MiT & \cellcolor{babyblueeyes!50}MSR & \cellcolor{babyblueeyes!50}DDM & \cellcolor{babyblueeyes!50}ActN & \cellcolor{babyblueeyes!50}LSMDC & \cellcolor{babyblueeyes!50}YC2 & \cellcolor{babyblueeyes!50}S-MiT\\
                \midrule

 middle-frame &  61.3 & 6.0 & 39.7 & 35.6 & 31.1 & 20.1 & 23.4 & 19.8 & 13.5 & 12.7 & 6.1 & 35.5\\
 max-score frame &  \underline{66.5} & \textbf{7.4} & \textbf{48.0} & \underline{43.3} & \underline{38.7} & 21.4 & 33.4 & \textbf{29.1} & \underline{21.5} & \textbf{18.3} & \textbf{8.9} & \textbf{46.1}  \\
\midrule
avg.-over-frames & \textbf{66.7} & \underline{6.8} & \underline{46.6} & 42.1 & 37.1 & \textbf{23.4} & 31.9 & 26.5 & \textbf{26.6} & 16.4 & \underline{8.6} & 43.8 \\
seq.-of-frames & 66.3 & 6.4 & \textbf{48.0} & \textbf{44.1} & \textbf{39.0} & \underline{22.6} & \textbf{36.6} & \underline{27.2} & \textbf{26.6} & \underline{17.0} & 8.4 & \underline{45.9} \\

                \bottomrule
    \end{tabular}
    }
    \vspace{-0.2cm}
    \caption{\small{\textbf{Evaluation of temporal representation bias.} 
     We report the performance of our model in \textit{objects+composition+activities} concepts setup and \textit{common sense bias} setup.
    \label{tab:temporal}
    }}
    \vspace{-0.2cm}
    
\end{table} 
\begin{table}[]
    \centering
    \renewcommand*{\arraystretch}{1.2}
    \setlength{\tabcolsep}{1pt}
    
    \resizebox{1\linewidth}{!}{
    \begin{tabular}{@{}l|ccc ccc | ccc ccc@{}}
    	\toprule
                &\cellcolor{darkpastelpurple!40}UCF & \cellcolor{darkpastelpurple!40}SSv2 & \cellcolor{darkpastelpurple!40}K400 & \cellcolor{darkpastelpurple!40}K600 & \cellcolor{darkpastelpurple!40}K700 & \cellcolor{darkpastelpurple!40}MiT & \cellcolor{babyblueeyes!50}MSR & \cellcolor{babyblueeyes!50}DDM & \cellcolor{babyblueeyes!50}ActN & \cellcolor{babyblueeyes!50}LSMDC & \cellcolor{babyblueeyes!50}YC2 & \cellcolor{babyblueeyes!50}S-MiT\\
                \midrule

 Fleiss' kappa &  86.1 & 77.8 & 91.2 & 90.7 & 92.5 & 91.2 & 92.0 & 91.7 & 90.9 & 92.5 & 91.3 & 93.2 \\
 \midrule
  \% object-biased samples & 73 &	7 & 55 &55 &48 &24 &28 &23 &20 &11 &6 &36 \\
  \midrule
 \% of full test split &   \multirow{2}{*}{27} & \multirow{2}{*}{93} & \multirow{2}{*}{45} & \multirow{2}{*}{45} & \multirow{2}{*}{52} & \multirow{2}{*}{76} & \multirow{2}{*}{72} & \multirow{2}{*}{77} & \multirow{2}{*}{80} & \multirow{2}{*}{89} & \multirow{2}{*}{94} & \multirow{2}{*}{64} \\
 in UTD-split & \\

                \bottomrule
    \end{tabular}
    }
    \vspace{-0.2cm}
    \caption{\small{\textbf{UTD-splits statistics.} Fleiss' kappa coefficient measures the agreement among 9 models used to create UTD-splits. 
    \label{tab:utd_splits}
    }}
    \vspace{-0.5cm}
    
\end{table}

\begin{table*}[]
    \centering
    \setlength{\tabcolsep}{2pt}
    \begin{tabular}{@{}c c@{}}
        \parbox[t]{0.6\linewidth}{ %
            \vspace{0pt} %
            \centering
    \resizebox{1\linewidth}{!}{
    \begin{tabular}{@{}l|ccc|ccc|ccc@{}}
    	\toprule

                  & \multirow{3}{*}{UCF}  & UCF- & UCF- & \multirow{3}{*}{SSv2}  & SSv2- & SSv2- & \multirow{3}{*}{K400}  & K400- & K400- \\
                 &  & UTD-  & UTD- &  & UTD-  & UTD- &  & UTD-  & UTD- \\
                 &  & split  & s. balanced  &  & split  & s. balanced  &  & split  & s. balanced  \\
                \midrule

videomae-B-K400 & 89.2 & \cellcolor{red!35}$81.3_{\scaleto{-7.9}{6pt}}$ & \cellcolor{cadmiumorange!50}$82.8_{\scaleto{-6.4}{6pt}}$ & 54.2 & \cellcolor{apricot!25}$52.1_{\scaleto{-2.1}{6pt}}$ & \cellcolor{apricot!25}$53.0_{\scaleto{-1.2}{6pt}}$ & 64.3 & \cellcolor{oucrimsonred!70}$45.7_{\scaleto{-18.6}{6pt}}$ & \cellcolor{red!50}$49.5_{\scaleto{-14.8}{6pt}}$ \\
videomae-B-UH & 88.4 & \cellcolor{red!35}$79.2_{\scaleto{-9.2}{6pt}}$ & \cellcolor{red!35}$80.4_{\scaleto{-8.0}{6pt}}$ & 56.2 & \cellcolor{apricot!25}$54.0_{\scaleto{-2.2}{6pt}}$ & \cellcolor{apricot!25}$54.9_{\scaleto{-1.3}{6pt}}$ & 64.3 & \cellcolor{oucrimsonred!70}$45.4_{\scaleto{-18.9}{6pt}}$ & \cellcolor{red!50}$49.6_{\scaleto{-14.7}{6pt}}$ \\
videomae-L-UH & 95.7 & \cellcolor{apricot!50}$90.9_{\scaleto{-4.8}{6pt}}$ & \cellcolor{apricot!50}$91.0_{\scaleto{-4.7}{6pt}}$ & 67.3 & \cellcolor{apricot!25}$65.4_{\scaleto{-1.9}{6pt}}$ & \cellcolor{apricot!25}$66.2_{\scaleto{-1.1}{6pt}}$ & 75.8 & \cellcolor{oucrimsonred!70}$59.3_{\scaleto{-16.5}{6pt}}$ & \cellcolor{red!50}$63.3_{\scaleto{-12.5}{6pt}}$ \\
videomae-H-UH & 95.2 & \cellcolor{apricot!50}$90.9_{\scaleto{-4.3}{6pt}}$ & \cellcolor{apricot!50}$91.0_{\scaleto{-4.2}{6pt}}$ & 67.4 & \cellcolor{apricot!25}$65.5_{\scaleto{-1.9}{6pt}}$ & \cellcolor{apricot!25}$66.3_{\scaleto{-1.1}{6pt}}$ & 76.5 & \cellcolor{oucrimsonred!70}$59.5_{\scaleto{-17.0}{6pt}}$ & \cellcolor{red!50}$63.7_{\scaleto{-12.8}{6pt}}$ \\
\midrule
videomaev2-B-K710-fnK710 & 99.0 & \cellcolor{apricot!25}$98.1_{\scaleto{-0.9}{6pt}}$ & \cellcolor{apricot!25}$97.6_{\scaleto{-1.4}{6pt}}$ & 57.1 & \cellcolor{apricot!25}$54.7_{\scaleto{-2.4}{6pt}}$ & \cellcolor{apricot!25}$55.9_{\scaleto{-1.2}{6pt}}$ & 85.0 & \cellcolor{red!50}$71.9_{\scaleto{-13.1}{6pt}}$ & \cellcolor{red!35}$75.5_{\scaleto{-9.5}{6pt}}$ \\
\midrule
allinone-B-WV2M+CC & 84.5 & \cellcolor{oucrimsonred!80}$63.1_{\scaleto{-21.4}{6pt}}$ & \cellcolor{red!50}$72.9_{\scaleto{-11.6}{6pt}}$ & 26.2 & \cellcolor{apricot!50}$22.5_{\scaleto{-3.7}{6pt}}$ & \cellcolor{apricot!25}$24.7_{\scaleto{-1.5}{6pt}}$ & 66.9 & \cellcolor{oucrimsonred!80}$42.6_{\scaleto{-24.3}{6pt}}$ & \cellcolor{oucrimsonred!70}$50.8_{\scaleto{-16.1}{6pt}}$ \\
allinone-B-WV2M+HT & 81.4 & \cellcolor{oucrimsonred!80}$57.7_{\scaleto{-23.7}{6pt}}$ & \cellcolor{red!50}$68.3_{\scaleto{-13.1}{6pt}}$ & 22.7 & \cellcolor{apricot!50}$19.3_{\scaleto{-3.4}{6pt}}$ & \cellcolor{apricot!25}$21.3_{\scaleto{-1.4}{6pt}}$ & 61.4 & \cellcolor{oucrimsonred!80}$36.9_{\scaleto{-24.5}{6pt}}$ & \cellcolor{oucrimsonred!70}$45.2_{\scaleto{-16.2}{6pt}}$ \\
allinone-B-WV2M+HT+CC+YTT+ & 81.3 & \cellcolor{oucrimsonred!80}$58.5_{\scaleto{-22.8}{6pt}}$ & \cellcolor{oucrimsonred!70}$66.2_{\scaleto{-15.1}{6pt}}$ & 22.0 & \cellcolor{apricot!50}$18.8_{\scaleto{-3.2}{6pt}}$ & \cellcolor{apricot!25}$20.7_{\scaleto{-1.3}{6pt}}$ & 61.2 & \cellcolor{oucrimsonred!80}$36.4_{\scaleto{-24.8}{6pt}}$ & \cellcolor{oucrimsonred!70}$44.6_{\scaleto{-16.6}{6pt}}$ \\
\midrule

umt-B-K710 & 91.3 & \cellcolor{red!50}$76.8_{\scaleto{-14.5}{6pt}}$ & \cellcolor{red!35}$82.4_{\scaleto{-8.9}{6pt}}$ & 50.6 & \cellcolor{apricot!50}$47.9_{\scaleto{-2.7}{6pt}}$ & \cellcolor{apricot!25}$49.1_{\scaleto{-1.5}{6pt}}$ & 77.7 & \cellcolor{oucrimsonred!70}$58.3_{\scaleto{-19.4}{6pt}}$ & \cellcolor{red!50}$63.7_{\scaleto{-14.0}{6pt}}$ \\
umt-B-K710-fnK710 & 99.0 & \cellcolor{apricot!25}$96.8_{\scaleto{-2.2}{6pt}}$ & \cellcolor{apricot!25}$97.1_{\scaleto{-1.9}{6pt}}$ & 49.4 & \cellcolor{apricot!50}$46.6_{\scaleto{-2.8}{6pt}}$ & \cellcolor{apricot!25}$48.1_{\scaleto{-1.3}{6pt}}$ & 85.6 & \cellcolor{red!50}$72.4_{\scaleto{-13.2}{6pt}}$ & \cellcolor{red!35}$76.2_{\scaleto{-9.4}{6pt}}$ \\
umt-L-K710 & 95.7 & \cellcolor{red!35}$87.2_{\scaleto{-8.5}{6pt}}$ & \cellcolor{cadmiumorange!50}$89.2_{\scaleto{-6.5}{6pt}}$ & 59.4 & \cellcolor{apricot!25}$57.0_{\scaleto{-2.4}{6pt}}$ & \cellcolor{apricot!25}$58.1_{\scaleto{-1.3}{6pt}}$ & 84.0 & \cellcolor{oucrimsonred!70}$68.4_{\scaleto{-15.6}{6pt}}$ & \cellcolor{red!50}$73.0_{\scaleto{-11.0}{6pt}}$ \\
umt-L-K710-fnK710 & 98.9 & \cellcolor{apricot!25}$96.7_{\scaleto{-2.2}{6pt}}$ & \cellcolor{apricot!25}$97.1_{\scaleto{-1.8}{6pt}}$ & 57.9 & \cellcolor{apricot!25}$55.5_{\scaleto{-2.4}{6pt}}$ & \cellcolor{apricot!25}$56.7_{\scaleto{-1.2}{6pt}}$ & 89.0 & \cellcolor{red!50}$78.7_{\scaleto{-10.3}{6pt}}$ & \cellcolor{red!35}$81.6_{\scaleto{-7.4}{6pt}}$ \\
\midrule

videomamba-vm-K400 & 90.9 & \cellcolor{oucrimsonred!70}$74.6_{\scaleto{-16.3}{6pt}}$ & \cellcolor{red!35}$81.0_{\scaleto{-9.9}{6pt}}$ & 35.9 & \cellcolor{apricot!50}$32.6_{\scaleto{-3.3}{6pt}}$ & \cellcolor{apricot!25}$34.4_{\scaleto{-1.5}{6pt}}$ & 75.3 & \cellcolor{oucrimsonred!80}$54.4_{\scaleto{-20.9}{6pt}}$ & \cellcolor{red!50}$60.5_{\scaleto{-14.8}{6pt}}$ \\
videomamba-vm-5M & 93.7 & \cellcolor{red!50}$82.0_{\scaleto{-11.7}{6pt}}$ & \cellcolor{cadmiumorange!50}$86.7_{\scaleto{-7.0}{6pt}}$ & 48.5 & \cellcolor{apricot!50}$45.6_{\scaleto{-2.9}{6pt}}$ & \cellcolor{apricot!25}$46.9_{\scaleto{-1.6}{6pt}}$ & 77.5 & \cellcolor{oucrimsonred!70}$58.1_{\scaleto{-19.4}{6pt}}$ & \cellcolor{red!50}$64.0_{\scaleto{-13.5}{6pt}}$ \\
videomamba-vm-17M & 93.2 & \cellcolor{red!50}$80.5_{\scaleto{-12.7}{6pt}}$ & \cellcolor{red!35}$84.3_{\scaleto{-8.9}{6pt}}$ & 47.4 & \cellcolor{apricot!50}$44.7_{\scaleto{-2.7}{6pt}}$ & \cellcolor{apricot!25}$45.9_{\scaleto{-1.5}{6pt}}$ & 77.7 & \cellcolor{oucrimsonred!70}$58.1_{\scaleto{-19.6}{6pt}}$ & \cellcolor{red!50}$63.6_{\scaleto{-14.1}{6pt}}$ \\
videomamba-vm-25M & 94.3 & \cellcolor{red!50}$83.0_{\scaleto{-11.3}{6pt}}$ & \cellcolor{red!35}$86.5_{\scaleto{-7.8}{6pt}}$ & 48.7 & \cellcolor{apricot!50}$45.9_{\scaleto{-2.8}{6pt}}$ & \cellcolor{apricot!25}$47.1_{\scaleto{-1.6}{6pt}}$ & 78.4 & \cellcolor{oucrimsonred!70}$59.1_{\scaleto{-19.3}{6pt}}$ & \cellcolor{red!50}$64.6_{\scaleto{-13.8}{6pt}}$ \\
\midrule

internvid-B-10M-FLT & 94.0 & \cellcolor{red!50}$81.1_{\scaleto{-12.9}{6pt}}$ & \cellcolor{red!35}$84.7_{\scaleto{-9.3}{6pt}}$ & 48.1 & \cellcolor{apricot!50}$45.2_{\scaleto{-2.9}{6pt}}$ & \cellcolor{apricot!25}$46.7_{\scaleto{-1.4}{6pt}}$ & 78.6 & \cellcolor{oucrimsonred!70}$59.3_{\scaleto{-19.3}{6pt}}$ & \cellcolor{red!50}$64.9_{\scaleto{-13.7}{6pt}}$ \\
internvid-B-200M & 94.5 & \cellcolor{red!50}$82.9_{\scaleto{-11.6}{6pt}}$ & \cellcolor{red!35}$85.2_{\scaleto{-9.3}{6pt}}$ & 54.4 & \cellcolor{apricot!50}$51.8_{\scaleto{-2.6}{6pt}}$ & \cellcolor{apricot!25}$52.8_{\scaleto{-1.6}{6pt}}$ & 80.0 & \cellcolor{oucrimsonred!70}$61.7_{\scaleto{-18.3}{6pt}}$ & \cellcolor{red!50}$66.8_{\scaleto{-13.2}{6pt}}$ \\
internvid-L-10M & 93.2 & \cellcolor{red!50}$80.2_{\scaleto{-13.0}{6pt}}$ & \cellcolor{red!35}$85.3_{\scaleto{-7.9}{6pt}}$ & 52.6 & \cellcolor{apricot!50}$50.0_{\scaleto{-2.6}{6pt}}$ & \cellcolor{apricot!25}$51.1_{\scaleto{-1.5}{6pt}}$ & 77.4 & \cellcolor{oucrimsonred!70}$57.4_{\scaleto{-20.0}{6pt}}$ & \cellcolor{red!50}$63.8_{\scaleto{-13.6}{6pt}}$ \\
internvid-L-WebVid10M & 94.1 & \cellcolor{red!50}$82.5_{\scaleto{-11.6}{6pt}}$ & \cellcolor{red!35}$86.3_{\scaleto{-7.8}{6pt}}$ & 54.9 & \cellcolor{apricot!25}$52.4_{\scaleto{-2.5}{6pt}}$ & \cellcolor{apricot!25}$53.4_{\scaleto{-1.5}{6pt}}$ & 78.9 & \cellcolor{oucrimsonred!70}$59.7_{\scaleto{-19.2}{6pt}}$ & \cellcolor{red!50}$65.7_{\scaleto{-13.2}{6pt}}$ \\
internvid-L-10M-DIV & 94.8 & \cellcolor{red!50}$83.2_{\scaleto{-11.6}{6pt}}$ & \cellcolor{red!35}$87.6_{\scaleto{-7.2}{6pt}}$ & 51.7 & \cellcolor{apricot!50}$48.8_{\scaleto{-2.9}{6pt}}$ & \cellcolor{apricot!25}$50.1_{\scaleto{-1.6}{6pt}}$ & 80.3 & \cellcolor{oucrimsonred!70}$62.1_{\scaleto{-18.2}{6pt}}$ & \cellcolor{red!50}$67.4_{\scaleto{-12.9}{6pt}}$ \\
internvid-L-10M-FLT & 95.5 & \cellcolor{red!35}$86.3_{\scaleto{-9.2}{6pt}}$ & \cellcolor{red!35}$88.4_{\scaleto{-7.1}{6pt}}$ & 53.6 & \cellcolor{apricot!50}$50.9_{\scaleto{-2.7}{6pt}}$ & \cellcolor{apricot!25}$52.1_{\scaleto{-1.5}{6pt}}$ & 81.9 & \cellcolor{oucrimsonred!70}$64.7_{\scaleto{-17.2}{6pt}}$ & \cellcolor{red!50}$69.6_{\scaleto{-12.3}{6pt}}$ \\
internvid-L-50M & 95.2 & \cellcolor{red!35}$85.5_{\scaleto{-9.7}{6pt}}$ & \cellcolor{cadmiumorange!50}$88.4_{\scaleto{-6.8}{6pt}}$ & 60.6 & \cellcolor{apricot!25}$58.2_{\scaleto{-2.4}{6pt}}$ & \cellcolor{apricot!25}$59.3_{\scaleto{-1.3}{6pt}}$ & 81.3 & \cellcolor{oucrimsonred!70}$63.7_{\scaleto{-17.6}{6pt}}$ & \cellcolor{red!50}$69.1_{\scaleto{-12.2}{6pt}}$ \\
internvid-L-200M & 96.8 & \cellcolor{cadmiumorange!50}$90.5_{\scaleto{-6.3}{6pt}}$ & \cellcolor{apricot!50}$91.9_{\scaleto{-4.9}{6pt}}$ & 63.1 & \cellcolor{apricot!25}$61.0_{\scaleto{-2.1}{6pt}}$ & \cellcolor{apricot!25}$61.9_{\scaleto{-1.2}{6pt}}$ & 84.6 & \cellcolor{oucrimsonred!70}$69.4_{\scaleto{-15.2}{6pt}}$ & \cellcolor{red!50}$73.9_{\scaleto{-10.7}{6pt}}$ \\

                \bottomrule
    \end{tabular}
    }
    \vspace{-0.2cm}
    \caption{\small{
    \textbf{Benchmarking video models in action classification. }  We report accuracy on full test/val sets and our debiased UTD-splits. The accuracy differences with respect to the full test/val sets are color-coded. 
    \label{tab:sota_ActionClassification}
            }}
        }
        &
        \parbox[t]{0.4\linewidth}{ %
            \vspace{0pt} %
            \centering
            \setlength{\tabcolsep}{2pt}
    \resizebox{1\linewidth}{!}{
    
    \begin{tabular}{@{}l|cc|cc|cc@{}}
    	\toprule
                & \multirow{3}{*}{MSR}  & MSR- & \multirow{3}{*}{DDM}  & DDM- & \multirow{3}{*}{ANet}  & ANet- \\
                 &  & UTD-  &  & UTD- &  & UTD- \\
                 &  & split &  & split &  & split  \\
                \midrule
                
                umt-b-5M & 30.0 & \cellcolor{red!35}$20.5_{\scaleto{-9.5}{6pt}}$ & 30.2 & \cellcolor{red!35}$20.6_{\scaleto{-9.6}{6pt}}$ & 28.6 & \cellcolor{cadmiumorange!50}$22.3_{\scaleto{-6.3}{6pt}}$ \\
umt-b-17M & 35.6 & \cellcolor{red!35}$26.3_{\scaleto{-9.3}{6pt}}$ & 37.7 & \cellcolor{red!50}$27.6_{\scaleto{-10.1}{6pt}}$ & 34.2 & \cellcolor{cadmiumorange!50}$27.3_{\scaleto{-6.9}{6pt}}$ \\
umt-b-25M & 35.3 & \cellcolor{red!50}$24.8_{\scaleto{-10.5}{6pt}}$ & 34.2 & \cellcolor{red!35}$24.6_{\scaleto{-9.6}{6pt}}$ & 25.1 & \cellcolor{cadmiumorange!35}$19.7_{\scaleto{-5.4}{6pt}}$ \\
umt-l-5M & 34.8 & \cellcolor{red!35}$24.9_{\scaleto{-9.9}{6pt}}$ & 33.5 & \cellcolor{red!50}$21.8_{\scaleto{-11.7}{6pt}}$ & 34.8 & \cellcolor{cadmiumorange!50}$28.7_{\scaleto{-6.1}{6pt}}$ \\
umt-l-17M & \textbf{43.6} & \cellcolor{red!50}$31.1_{\scaleto{-12.5}{6pt}}$ & \textbf{46.3} & \cellcolor{red!50}$35.6_{\scaleto{-10.7}{6pt}}$ & \textbf{45.9} & \cellcolor{red!35}$38.7_{\scaleto{-7.2}{6pt}}$ \\
umt-l-25M & 42.3 & \cellcolor{red!50}$30.6_{\scaleto{-11.7}{6pt}}$ & 43.6 & \cellcolor{red!50}$33.5_{\scaleto{-10.1}{6pt}}$ & 36.7 & \cellcolor{cadmiumorange!50}$30.4_{\scaleto{-6.3}{6pt}}$ \\
\midrule

videomamba-vm-5M & 33.3 & \cellcolor{red!50}$23.0_{\scaleto{-10.3}{6pt}}$ & 37.1 & \cellcolor{red!35}$27.1_{\scaleto{-10.0}{6pt}}$ & 37.1 & \cellcolor{cadmiumorange!50}$30.1_{\scaleto{-7.0}{6pt}}$ \\
videomamba-vm-17M & \textbf{34.9} & \cellcolor{red!35}$25.5_{\scaleto{-9.4}{6pt}}$ & 40.6 & \cellcolor{red!50}$28.9_{\scaleto{-11.7}{6pt}}$ & 40.4 & \cellcolor{red!35}$33.0_{\scaleto{-7.4}{6pt}}$ \\
videomamba-vm-25M & \textbf{34.9} & \cellcolor{red!35}$25.5_{\scaleto{-9.4}{6pt}}$ & \textbf{41.4} & \cellcolor{red!50}$30.5_{\scaleto{-10.9}{6pt}}$ & \textbf{41.1} & \cellcolor{red!35}$33.8_{\scaleto{-7.3}{6pt}}$ \\
\midrule

internvid-B-10M-FLT & 37.9 & \cellcolor{red!50}$25.4_{\scaleto{-12.5}{6pt}}$ & 28.6 & \cellcolor{red!50}$17.2_{\scaleto{-11.4}{6pt}}$ & 24.4 & \cellcolor{cadmiumorange!35}$18.8_{\scaleto{-5.6}{6pt}}$ \\
internvid-B-200M & 38.1 & \cellcolor{red!50}$24.7_{\scaleto{-13.4}{6pt}}$ & 30.2 & \cellcolor{red!50}$19.4_{\scaleto{-10.8}{6pt}}$ & 26.2 & \cellcolor{cadmiumorange!50}$20.1_{\scaleto{-6.1}{6pt}}$ \\
internvid-L-10M & 26.7 & \cellcolor{red!35}$18.7_{\scaleto{-8.0}{6pt}}$ & 22.6 & \cellcolor{cadmiumorange!50}$15.6_{\scaleto{-7.0}{6pt}}$ & 21.5 & \cellcolor{cadmiumorange!35}$16.3_{\scaleto{-5.2}{6pt}}$ \\
internvid-L-WV10M & 26.5 & \cellcolor{red!35}$17.6_{\scaleto{-8.9}{6pt}}$ & 22.2 & \cellcolor{red!35}$13.8_{\scaleto{-8.4}{6pt}}$ & 23.1 & \cellcolor{cadmiumorange!35}$17.5_{\scaleto{-5.6}{6pt}}$ \\
internvid-L-10M-DIV & 37.3 & \cellcolor{red!50}$24.2_{\scaleto{-13.1}{6pt}}$ & 26.9 & \cellcolor{red!50}$15.9_{\scaleto{-11.0}{6pt}}$ & 23.2 & \cellcolor{cadmiumorange!35}$17.7_{\scaleto{-5.5}{6pt}}$ \\
internvid-L-10M-FLT & \textbf{38.7} & \cellcolor{red!50}$26.3_{\scaleto{-12.4}{6pt}}$ & 29.2 & \cellcolor{red!50}$18.8_{\scaleto{-10.4}{6pt}}$ & 24.5 & \cellcolor{cadmiumorange!35}$19.0_{\scaleto{-5.5}{6pt}}$ \\
internvid-L-50M & 32.4 & \cellcolor{red!50}$22.0_{\scaleto{-10.4}{6pt}}$ & 26.5 & \cellcolor{red!35}$18.3_{\scaleto{-8.2}{6pt}}$ & 24.4 & \cellcolor{cadmiumorange!50}$18.0_{\scaleto{-6.4}{6pt}}$ \\
internvid-L-200M & 38.2 & \cellcolor{red!50}$24.8_{\scaleto{-13.4}{6pt}}$ & \textbf{30.3} & \cellcolor{red!50}$20.2_{\scaleto{-10.1}{6pt}}$ & \textbf{28.7} & \cellcolor{cadmiumorange!50}$22.1_{\scaleto{-6.6}{6pt}}$ \\

                \bottomrule
    \end{tabular}
    }
    \vspace{-0.2cm}
    \caption{\small{\textbf{Benchmarking video-language models in text-to-video retrieval.} We report accuracy on full test/val sets and our debiased UTD-splits. The accuracy differences with respect to the full test/val sets are color-coded. 
    \label{tab:sota_Text-to-Video}
    }}
        }
    \end{tabular}
    \vspace{-0.5cm}
\end{table*}

In this section, we analyze action classification and text-to-video retrieval datasets on different representation biases using our UTD method.

\subsection{Setup}

\noindent\textbf{Datasets.}
We consider six common \textbf{action classification} datasets: UCF101 (UCF)~\cite{SoomroUCF101}, SomethingSomethingv2 (SSv2)~\cite{Goyal_2017_ICCV}, Kinetics400 (K400)~\cite{Kay2017Kinetics}, -600 (K600)~\cite{Carreira2018Kinetics600}, -700 (K700)~\cite{Carreira2019Kinetics700},  Moments In Time (MiT)~\cite{monfort2019moments} and six \textbf{text-to-video retrieval} datasets: MSRVTT (MSR)~\cite{xu2016msr}, YouCook2 (YC2)~\cite{Das_2013_YouCook}, DiDeMo (DDM)~\cite{anne2017localizing}, LSMDC~\cite{Rohrbach_2015_CVPR}, ActivityNet (ANet)~\cite{Heilbron_2015_CVPR}, and Spoken Moments in Time (S-MiT)~\cite{Monfort_2021_CVPR} .

\noindent\textbf{Implementation Details.}
For all train/test videos, we use $|v_n|=8$ uniformly sampled frames. We use the LLaVA-1.6-7B-mistral model~\cite{liu2024llavanext} as VLM, Mistral-7B-Instruct-v0.2~\cite{jiang2023mistral} as LLM, and SFR-Embedding-Mistral~\cite{meng2024sfrembedding} as text embedding model. The temperature is set to 0. 

\subsection{UTD-descriptions}

Following our UTD pipeline, first, we generate textual descriptions for 12 considered datasets, resulting in \textit{UTD-descriptions} dataset (see qualitative examples in the supplement).  
We conducted a user study to validate the quality of object, activity, and verb predictions. Participants
were asked to verify which objects from a provided list were visible in a given image. The study used 100 random frames from 100 videos of the UCF dataset (which has noisy, low-resolution frames that might be an additional
challenge to the VLM). 
Seven participants validated 761 annotated objects in total, where 87.6\% (667 out of 761) were identified as visible, with only 94 not selected as visible. To better
understand the VLM's errors, we manually classified these 94 objects into five categories: 1) attribute error (e.g., the object is
``right hand'' instead of ``left hand''), 2) misclassification (the object is present but incorrectly identified as a different object, e.g., a
``snowboard'' instead of ``snow slide''), 3) hallucination, 4) human annotation mistake – the object is visible, and 5) other. We found
that only 45 objects were classified as hallucinations, 13 as misclassifications, 20 as ``object visible'' 8 as attribute errors, and 8 as
other (see examples in the supplement). This results in a hallucination rate of only 6\%, which we find acceptable for our analysis.

\subsection{Concept Representation Bias}

Further, based on generated descriptions, we evaluate representation bias with respect to four concept categories, namely objects, activities, verbs, and objects+composition+activities, as described in~\cref{sec:RepresentationBiases}. For this, we consider \textit{common sense bias} in \textit{sequence-of-frames} temporal setup in~\cref{tab:all_biases} (all concept-temporal combination results can be found in the supplement).
It shows that using only object information results in only a slight drop in performance compared to objects+composition+activities, indicating that most datasets can, to a certain extent, be addressed with object information, thus showing an object representation bias. Interestingly, classification based on only objects results in 63.3pp accuracy for UCF101 and 45.9pp for Kinetics400. 
This may be due to the fact that some action classes are tied to the objects (e.g. ``playing piano’’), making visible objects equally or more reliable indicators of a class than activities.
This becomes clear when we consider the verbs-based representation. 
Here, the performance drops significantly for most action classification datasets, except for SSv2 and MiT. %
Also, in the case of text-to-video retrieval, it shows that objects are better indicators than activities for some datasets (MSRVTT, DiDeMo, ActivityNet), and activities are better indicators for others (YouCook2, LSMDC, S-MiT). %

\subsection{Common Sense vs. Dataset Bias}

We further consider common sense bias compared to dataset bias for the action classification datasets in~\cref{tab:dataset_bias}. 
Overall, across all benchmarks and concepts, dataset bias is significantly higher than common sense bias, and the results indicate that considered benchmarks also contain a dataset-specific object bias in the training set.
The performance on UCF using only objects increases from 63.3pp to 80.3pp, and even on SSv2, which was developed to minimize object representation bias, using only correlation of seen objects in the train set to the classes, we were able to predict 13.6pp of videos correctly. 
This significant increase based on available object information only shows that some actions might be connected to unrelated objects that hint at a certain class without being directly associated with it.   %

\subsection{Temporal Representation Bias}

Finally, we consider representation bias with respect to the temporal aspect in~\cref{{tab:temporal}}.
Namely, we consider which temporal representations, single frame vs. multiple frames, best-score frame vs. middle frame, as well as average-over-frames vs. sequence-of-frames embeddings, perform best.
For all datasets except ActivityNet, max-score-frame performance is comparable ($< 3.5\%$ difference) to multiple frames performance (average of frames/sequence over frames). Moreover, for SSv2, Kinetics400,  DiDeMo, LSMDC, YouCook2, and S-MiT, we found that max-score-frame is a better indicator of class/captions than all frames. %

\subsection{UTD-splits}

Since our analysis reveals significant object bias in benchmarks,
we generate object-debiased UTD-splits for considered benchmarks, as described in~\cref{sec:unbiasing_datasets}. Namely, we remove object-biased samples where nine models that use different prompts in the text embedding model or trained on the bootstrapped train set uniformly agree that the sample can be predicted correctly based solely on objects. 
\cref{tab:utd_splits} summarizes the percentage of object-biased samples identified in the datasets, the level of model agreement (measured by Fleiss' kappa coefficient, which is notably high), and the number of samples retained from the original datasets.

\section{Benchmarking Video Models}\label{sec:benchmarking}

In this section, we benchmark state-of-the-art video encoders on considered datasets and their UTD-splits.

\subsection{Setup}

\noindent\textbf{Video Models.} We consider state-of-the-art video encoders with varying model sizes and pre-training configurations with open-sourced weights. For \textbf{action classification}, we benchmark 24 different variants of VideoMAE~\cite{tong2022videomae}, VideoMAE2~\cite{wang2023videomae}, All-in-one~\cite{wang2023all}, Unmasked Teacher (UMT)~\cite{li2023unmasked}, VideoMamba~\cite{li2024videomamba}, and InternVid~\cite{wang2023internvid} families. 
For \textbf{text-to-video retrieval}, we consider 17 video-language models of UMT~\cite{li2023unmasked}, VideoMamba~\cite{li2024videomamba}, and InternVid~\cite{wang2023internvid} families. In total, we analyze 30 models and list all considered variants in the supplement. 

\noindent\textbf{Implementation Details.}
To evaluate action classification performance in a standard aligned setup, we follow the VideoGLUE~\cite{yuan2023videoglue} benchmark of using a frozen video backbone and lightweight pooing head as a test of the strength of video model representation. 
To adapt each video backbone for a specific dataset, we create a model that 1) uses frozen video backbone to extract spatial-temporal features, and then 2) uses a single attention pooling layer~\cite{yuan2023videoglue} to aggregate spatial-temporal features into a single vector representation and 3) a single linear projection layer to classify on the required number of classes. 
We train the model on a full train split of the corresponding dataset. 
For video model evaluation in text-to-video retrieval, we reported their zero-shot performance on our debiased splits. We follow the corresponding model recipes in using re-ranking and how many examples are reranked. We report the accuracy for classification and the recall@1 for text-to-video retrieval.
Additional details are provided in the supplement.

\subsection{Results}

We evaluate 
the performance of 
video models for action classification on the UCF, SSv2, and K400 datasets and their UTD-splits in ~\cref{tab:sota_ActionClassification} and for text-to-video retrieval on the MSRVTT, DiDeMo, and ActivityNet datasets and their UTD-splits in~\cref{tab:sota_Text-to-Video}. Performance for other datasets is reported in the supplement. 
Across all models, performance drops notably on debiased splits (except for SSv2), suggesting that current models strongly rely on object representation bias.
Evaluation on UTD debiased splits shows a higher range of scores of models, revealing that some models (e.g. all-in-one models) were more heavily relying on object bias rather than utilizing action/motion cues for video understanding. Further, we can observe that the best performance flips for some models for the debiased case compared to the full test set, e.g., for videomae-B-K400 compared to videomamba-vm-K400 on the UCF dataset.

Additionally, by comparing the performance decline between the original datasets and the UTD-splits, we can assess models' sensitivity to object bias.
Overall, we see in the case of \textit{action classification}, \cref{tab:sota_ActionClassification}, that VideoMAE models are most robust to object bias, whereas Allinone models are most impacted by object bias. We also note that performance drops correlate between the UTD-split and UTD-split-balanced versions.
We can further observe that larger backbones/architectures, and to a certain extent, more training data, lead to a reduced bias compared to smaller models, e.g. InternVid and VideoMAE.  
This pattern does not hold for \textit{text-to-video retrieval}, shown in \cref{tab:sota_Text-to-Video}. Here, larger backbones and backbones trained with more data turn out to have higher object bias than smaller models or models trained with less data (see InternVid).
We attribute this behavior, to the fact that retrieval models leverage information from both the vision as well as the text backbone. Thus, if larger text backbones also learn more object properties or a larger vocabulary of objects, this might further encourage object-centering learning in the visual stream as well.

\section{Conclusion}

In this paper, we presented a new way to examine representation biases in video classification and retrieval datasets by leveraging recent, powerful VLMs and LLMs. Our investigation confirms that most video datasets are heavily focused on object recognition. Based on our bias analysis, we propose a method to reduce object bias in these benchmarks, enabling more robust video understanding evaluation. With that, we are able to benchmark state-of-the-art video models and show an analysis of the object bias in these models. We hope that our findings, together with the presented benchmark, offer valuable insights for the future development of more robust video understanding benchmarks and models.

\section{Acknowledgments}
Nina Shvetsova is supported in part by German Federal Ministry of Education and Research (BMBF) project STCL - 01IS22067. Nina Shvetsova also acknowledges travel support from ELISE (GA no 951847). Prof. Hilde Kuehne is supported in part by the ERC Starting Grant GraViLa 101117556.

{
    \small
    \bibliographystyle{ieeenat_fullname}
    \bibliography{egbib}
}

\clearpage

\addcontentsline{toc}{section}{Appendix} 
\appendix
\noindent{\Large\bf Supplementary Material}\\[1em]

\numberwithin{table}{section}
\numberwithin{figure}{section}

In the supplementary material, we provide additional experimental results, implementation details, and qualitative examples. Furthermore, we discuss the UTD dataset's license, limitations, and broader impact, and provide a datasheet~\cite{gebru2021datasheets} for the UTD dataset.
Specifically, we first present additional experimental results, including extended results on benchmarking video models and common-sense bias, ablation studies, as well as analysis of class distribution in UTD-splits in~\cref{sec:add_results}. Next, we provide further implementation details of our UTD method in~\cref{sec:add_imp_details}. We then demonstrate qualitative examples from our UTD-descriptions and UTD-splits datasets, along with samples from the user study, in~\cref{sec:qualitative}. 
Finally, we discuss the UTD dataset's license in~\cref{sec:license}, the limitations of our work and its broader impact in~\cref{sec:limitations}, and provide a datasheet for the UTD dataset in~\cref{sec:datasheet}.

\section{Additional Results}
\label{sec:add_results}

\subsection{Benchmarking Video Models}

\begin{table*}
    \centering
    \setlength{\tabcolsep}{2pt}

    \resizebox{1\linewidth}{!}{
    
    \begin{tabular}{@{}l|ccc|ccc|ccc|ccc|ccc|ccc@{}}
    	\toprule
                  & \multirow{3}{*}{UCF}  & UCF- & UCF- & \multirow{3}{*}{SSv2}  & SSv2- & SSv2- & \multirow{3}{*}{K400}  & K400- & K400- & \multirow{3}{*}{K600}  & K600- & K600- & \multirow{3}{*}{K700}  & K700- & K700- & \multirow{3}{*}{MiT}  & MiT- & MiT- \\
                 &  & UTD-  & UTD- &  & UTD-  & UTD- &  & UTD-  & UTD- &  & UTD-  & UTD- &  & UTD-  & UTD- &  & UTD-  & UTD- \\
                 &  & split  & s. balanced  &  & split  & s. balanced  &  & split  & s. balanced  &  & split  & s. balanced  &  & split  & s. balanced  &  & split  & s. balanced \\
                \midrule

videomae-L-UH  & 95.7 & \cellcolor{apricot!50}$90.9_{\scaleto{-4.8}{6pt}}$ & \cellcolor{apricot!50}$91.0_{\scaleto{-4.7}{6pt}}$ & 67.3 & \cellcolor{apricot!25}$65.4_{\scaleto{-1.9}{6pt}}$ & \cellcolor{apricot!25}$66.2_{\scaleto{-1.1}{6pt}}$ & 75.8 & \cellcolor{oucrimsonred!70}$59.3_{\scaleto{-16.5}{6pt}}$ & \cellcolor{red!50}$63.3_{\scaleto{-12.5}{6pt}}$ & 76.9 & \cellcolor{oucrimsonred!70}$61.1_{\scaleto{-15.8}{6pt}}$ & \cellcolor{red!50}$65.8_{\scaleto{-11.1}{6pt}}$ & 65.6 & \cellcolor{oucrimsonred!70}$47.6_{\scaleto{-18.0}{6pt}}$ & \cellcolor{red!50}$52.7_{\scaleto{-12.9}{6pt}}$ & 38.1 & \cellcolor{red!50}$26.4_{\scaleto{-11.7}{6pt}}$ & \cellcolor{red!35}$30.8_{\scaleto{-7.3}{6pt}}$ \\
videomaev2-B-K710-fnK710 & 99.0 & \cellcolor{apricot!25}$98.1_{\scaleto{-0.9}{6pt}}$ & \cellcolor{apricot!25}$97.6_{\scaleto{-1.4}{6pt}}$ & 57.1 & \cellcolor{apricot!25}$54.7_{\scaleto{-2.4}{6pt}}$ & \cellcolor{apricot!25}$55.9_{\scaleto{-1.2}{6pt}}$ & 85.0 & \cellcolor{red!50}$71.9_{\scaleto{-13.1}{6pt}}$ & \cellcolor{red!35}$75.5_{\scaleto{-9.5}{6pt}}$ & 85.6 & \cellcolor{red!50}$73.2_{\scaleto{-12.4}{6pt}}$ & \cellcolor{red!35}$77.2_{\scaleto{-8.4}{6pt}}$ & 75.9 & \cellcolor{oucrimsonred!70}$60.0_{\scaleto{-15.9}{6pt}}$ & \cellcolor{red!50}$65.3_{\scaleto{-10.6}{6pt}}$ & 40.9 & \cellcolor{red!50}$28.9_{\scaleto{-12.0}{6pt}}$ & \cellcolor{cadmiumorange!50}$33.9_{\scaleto{-7.0}{6pt}}$ \\
\midrule
allinone-B-WV2M  & 84.5 & \cellcolor{oucrimsonred!80}$63.1_{\scaleto{-21.4}{6pt}}$ & \cellcolor{red!50}$72.9_{\scaleto{-11.6}{6pt}}$ & 26.2 & \cellcolor{apricot!50}$22.5_{\scaleto{-3.7}{6pt}}$ & \cellcolor{apricot!25}$24.7_{\scaleto{-1.5}{6pt}}$ & 66.9 & \cellcolor{oucrimsonred!80}$42.6_{\scaleto{-24.3}{6pt}}$ & \cellcolor{oucrimsonred!70}$50.8_{\scaleto{-16.1}{6pt}}$ & 68.0 & \cellcolor{oucrimsonred!80}$45.0_{\scaleto{-23.0}{6pt}}$ & \cellcolor{red!50}$53.3_{\scaleto{-14.7}{6pt}}$ & 55.2 & \cellcolor{oucrimsonred!80}$32.1_{\scaleto{-23.1}{6pt}}$ & \cellcolor{oucrimsonred!70}$40.0_{\scaleto{-15.2}{6pt}}$ & 29.9 & \cellcolor{red!50}$16.9_{\scaleto{-13.0}{6pt}}$ & \cellcolor{red!35}$22.0_{\scaleto{-7.9}{6pt}}$ \\
\midrule
umt-B-fnK710 & 99.0 & \cellcolor{apricot!25}$96.8_{\scaleto{-2.2}{6pt}}$ & \cellcolor{apricot!25}$97.1_{\scaleto{-1.9}{6pt}}$ & 49.4 & \cellcolor{apricot!50}$46.6_{\scaleto{-2.8}{6pt}}$ & \cellcolor{apricot!25}$48.1_{\scaleto{-1.3}{6pt}}$ & 85.6 & \cellcolor{red!50}$72.4_{\scaleto{-13.2}{6pt}}$ & \cellcolor{red!35}$76.2_{\scaleto{-9.4}{6pt}}$ & 86.2 & \cellcolor{red!50}$73.8_{\scaleto{-12.4}{6pt}}$ & \cellcolor{red!35}$78.1_{\scaleto{-8.1}{6pt}}$ & 76.9 & \cellcolor{oucrimsonred!70}$61.5_{\scaleto{-15.4}{6pt}}$ & \cellcolor{red!50}$66.8_{\scaleto{-10.1}{6pt}}$ & 39.7 & \cellcolor{red!50}$27.1_{\scaleto{-12.6}{6pt}}$ & \cellcolor{red!35}$32.1_{\scaleto{-7.6}{6pt}}$ \\
umt-L-fnK710 & 98.9 & \cellcolor{apricot!25}$96.7_{\scaleto{-2.2}{6pt}}$ & \cellcolor{apricot!25}$97.1_{\scaleto{-1.8}{6pt}}$ & 57.9 & \cellcolor{apricot!25}$55.5_{\scaleto{-2.4}{6pt}}$ & \cellcolor{apricot!25}$56.7_{\scaleto{-1.2}{6pt}}$ & 89.0 & \cellcolor{red!50}$78.7_{\scaleto{-10.3}{6pt}}$ & \cellcolor{red!35}$81.6_{\scaleto{-7.4}{6pt}}$ & 89.0 & \cellcolor{red!50}$78.9_{\scaleto{-10.1}{6pt}}$ & \cellcolor{cadmiumorange!50}$82.3_{\scaleto{-6.7}{6pt}}$ & 82.1 & \cellcolor{red!50}$69.3_{\scaleto{-12.8}{6pt}}$ & \cellcolor{red!35}$74.0_{\scaleto{-8.1}{6pt}}$ & 44.5 & \cellcolor{red!50}$32.3_{\scaleto{-12.2}{6pt}}$ & \cellcolor{red!35}$37.2_{\scaleto{-7.3}{6pt}}$ \\
\midrule
videomamba-vm-25M & 94.3 & \cellcolor{red!50}$83.0_{\scaleto{-11.3}{6pt}}$ & \cellcolor{red!35}$86.5_{\scaleto{-7.8}{6pt}}$ & 48.7 & \cellcolor{apricot!50}$45.9_{\scaleto{-2.8}{6pt}}$ & \cellcolor{apricot!25}$47.1_{\scaleto{-1.6}{6pt}}$ & 78.4 & \cellcolor{oucrimsonred!70}$59.1_{\scaleto{-19.3}{6pt}}$ & \cellcolor{red!50}$64.6_{\scaleto{-13.8}{6pt}}$ & 78.5 & \cellcolor{oucrimsonred!70}$60.2_{\scaleto{-18.3}{6pt}}$ & \cellcolor{red!50}$66.6_{\scaleto{-11.9}{6pt}}$ & 68.1 & \cellcolor{oucrimsonred!80}$47.5_{\scaleto{-20.6}{6pt}}$ & \cellcolor{red!50}$54.5_{\scaleto{-13.6}{6pt}}$ & 37.9 & \cellcolor{red!50}$24.5_{\scaleto{-13.4}{6pt}}$ & \cellcolor{red!35}$29.7_{\scaleto{-8.2}{6pt}}$ \\
\midrule
internvid-B-10M-FLT & 94.0 & \cellcolor{red!50}$81.1_{\scaleto{-12.9}{6pt}}$ & \cellcolor{red!35}$84.7_{\scaleto{-9.3}{6pt}}$ & 48.1 & \cellcolor{apricot!50}$45.2_{\scaleto{-2.9}{6pt}}$ & \cellcolor{apricot!25}$46.7_{\scaleto{-1.4}{6pt}}$ & 78.6 & \cellcolor{oucrimsonred!70}$59.3_{\scaleto{-19.3}{6pt}}$ & \cellcolor{red!50}$64.9_{\scaleto{-13.7}{6pt}}$ & 78.7 & \cellcolor{oucrimsonred!70}$60.1_{\scaleto{-18.6}{6pt}}$ & \cellcolor{red!50}$66.6_{\scaleto{-12.1}{6pt}}$ & 68.5 & \cellcolor{oucrimsonred!80}$47.8_{\scaleto{-20.7}{6pt}}$ & \cellcolor{red!50}$55.0_{\scaleto{-13.5}{6pt}}$ & 39.3 & \cellcolor{red!50}$26.0_{\scaleto{-13.3}{6pt}}$ & \cellcolor{red!35}$31.1_{\scaleto{-8.2}{6pt}}$ \\
internvid-B-200M & 94.5 & \cellcolor{red!50}$82.9_{\scaleto{-11.6}{6pt}}$ & \cellcolor{red!35}$85.2_{\scaleto{-9.3}{6pt}}$ & 54.4 & \cellcolor{apricot!50}$51.8_{\scaleto{-2.6}{6pt}}$ & \cellcolor{apricot!25}$52.8_{\scaleto{-1.6}{6pt}}$ & 80.0 & \cellcolor{oucrimsonred!70}$61.7_{\scaleto{-18.3}{6pt}}$ & \cellcolor{red!50}$66.8_{\scaleto{-13.2}{6pt}}$ & 79.9 & \cellcolor{oucrimsonred!70}$62.0_{\scaleto{-17.9}{6pt}}$ & \cellcolor{red!50}$68.5_{\scaleto{-11.4}{6pt}}$ & 70.2 & \cellcolor{oucrimsonred!70}$50.3_{\scaleto{-19.9}{6pt}}$ & \cellcolor{red!50}$57.6_{\scaleto{-12.6}{6pt}}$ & 39.9 & \cellcolor{red!50}$26.6_{\scaleto{-13.3}{6pt}}$ & \cellcolor{red!35}$31.9_{\scaleto{-8.0}{6pt}}$ \\
internvid-L-10M-FLT & 95.5 & \cellcolor{red!35}$86.3_{\scaleto{-9.2}{6pt}}$ & \cellcolor{red!35}$88.4_{\scaleto{-7.1}{6pt}}$ & 53.6 & \cellcolor{apricot!50}$50.9_{\scaleto{-2.7}{6pt}}$ & \cellcolor{apricot!25}$52.1_{\scaleto{-1.5}{6pt}}$ & 81.9 & \cellcolor{oucrimsonred!70}$64.7_{\scaleto{-17.2}{6pt}}$ & \cellcolor{red!50}$69.6_{\scaleto{-12.3}{6pt}}$ & 81.5 & \cellcolor{oucrimsonred!70}$64.8_{\scaleto{-16.7}{6pt}}$ & \cellcolor{red!50}$70.8_{\scaleto{-10.7}{6pt}}$ & 72.4 & \cellcolor{oucrimsonred!70}$53.3_{\scaleto{-19.1}{6pt}}$ & \cellcolor{red!50}$60.3_{\scaleto{-12.1}{6pt}}$ & 41.9 & \cellcolor{red!50}$28.9_{\scaleto{-13.0}{6pt}}$ & \cellcolor{red!35}$33.8_{\scaleto{-8.1}{6pt}}$ \\
internvid-L-200M & 96.8 & \cellcolor{cadmiumorange!50}$90.5_{\scaleto{-6.3}{6pt}}$ & \cellcolor{apricot!50}$91.9_{\scaleto{-4.9}{6pt}}$ & 63.1 & \cellcolor{apricot!25}$61.0_{\scaleto{-2.1}{6pt}}$ & \cellcolor{apricot!25}$61.9_{\scaleto{-1.2}{6pt}}$ & 84.6 & \cellcolor{oucrimsonred!70}$69.4_{\scaleto{-15.2}{6pt}}$ & \cellcolor{red!50}$73.9_{\scaleto{-10.7}{6pt}}$ & 84.3 & \cellcolor{red!50}$69.3_{\scaleto{-15.0}{6pt}}$ & \cellcolor{red!35}$75.1_{\scaleto{-9.2}{6pt}}$ & 76.4 & \cellcolor{oucrimsonred!70}$59.4_{\scaleto{-17.0}{6pt}}$ & \cellcolor{red!50}$66.0_{\scaleto{-10.4}{6pt}}$ & 44.3 & \cellcolor{red!50}$31.7_{\scaleto{-12.6}{6pt}}$ & \cellcolor{red!35}$36.9_{\scaleto{-7.4}{6pt}}$ \\

 \bottomrule
    \end{tabular}
    }
    \caption{\small{\textbf{Benchmarking video models in action classification on all six considered classification datasets.}  We report accuracy on full test/val sets and our debiased UTD-splits. The accuracy differences with respect to the full test/val sets are color-coded. 
    \label{tab:sota_ActionClassification_add}
    }}
    
\end{table*} 

\begin{table*}
    \centering

    \resizebox{0.9\linewidth}{!}{
    
    \begin{tabular}{@{}l|cc|cc|cc|cc|cc|cc@{}}
    	\toprule
                & \multirow{3}{*}{MSR}  & MSR- & \multirow{3}{*}{DDM}  & DDM- & \multirow{3}{*}{ANet}  & ANet-& \multirow{3}{*}{LSMDC}  & LSMDC- & \multirow{3}{*}{YC2}  & YC2- & \multirow{3}{*}{S-MiT}  & S-MiT- \\
                 &  & UTD-  &  & UTD- &  & UTD- &  & UTD-  &  & UTD- &  & UTD- \\
                 &  & split &  & split &  & split &  & split &  & split &  & split \\
                \midrule

umt-b-5M & 30.0 & \cellcolor{red!35}$20.5_{\scaleto{-9.5}{6pt}}$ & 30.2 & \cellcolor{red!35}$20.6_{\scaleto{-9.6}{6pt}}$ & 28.6 & \cellcolor{cadmiumorange!50}$22.3_{\scaleto{-6.3}{6pt}}$ & 14.1 & \cellcolor{apricot!50}$9.2_{\scaleto{-4.9}{6pt}}$ & 6.1 & \cellcolor{apricot!25}$5.2_{\scaleto{-0.9}{6pt}}$ & 47.9 & \cellcolor{red!50}$34.0_{\scaleto{-13.9}{6pt}}$ \\
umt-b-17M & 35.6 & \cellcolor{red!35}$26.3_{\scaleto{-9.3}{6pt}}$ & 37.7 & \cellcolor{red!50}$27.6_{\scaleto{-10.1}{6pt}}$ & 34.2 & \cellcolor{cadmiumorange!50}$27.3_{\scaleto{-6.9}{6pt}}$ & 16.6 & \cellcolor{cadmiumorange!35}$11.1_{\scaleto{-5.5}{6pt}}$ & 8.4 & \cellcolor{apricot!25}$6.9_{\scaleto{-1.5}{6pt}}$ & 53.5 & \cellcolor{red!50}$39.5_{\scaleto{-14.0}{6pt}}$ \\
umt-b-25M & 35.3 & \cellcolor{red!50}$24.8_{\scaleto{-10.5}{6pt}}$ & 34.2 & \cellcolor{red!35}$24.6_{\scaleto{-9.6}{6pt}}$ & 25.1 & \cellcolor{cadmiumorange!35}$19.7_{\scaleto{-5.4}{6pt}}$ & 13.1 & \cellcolor{apricot!50}$9.0_{\scaleto{-4.1}{6pt}}$ & 10.3 & \cellcolor{apricot!25}$8.7_{\scaleto{-1.6}{6pt}}$ & 53.9 & \cellcolor{red!50}$40.2_{\scaleto{-13.7}{6pt}}$ \\
umt-l-5M & 34.8 & \cellcolor{red!35}$24.9_{\scaleto{-9.9}{6pt}}$ & 33.5 & \cellcolor{red!50}$21.8_{\scaleto{-11.7}{6pt}}$ & 34.8 & \cellcolor{cadmiumorange!50}$28.7_{\scaleto{-6.1}{6pt}}$ & 21.5 & \cellcolor{apricot!50}$16.5_{\scaleto{-5.0}{6pt}}$ & 7.1 & \cellcolor{apricot!25}$5.8_{\scaleto{-1.3}{6pt}}$ & 51.9 & \cellcolor{red!50}$37.9_{\scaleto{-14.0}{6pt}}$ \\
umt-l-17M & 43.6 & \cellcolor{red!50}$31.1_{\scaleto{-12.5}{6pt}}$ & 46.3 & \cellcolor{red!50}$35.6_{\scaleto{-10.7}{6pt}}$ & 45.9 & \cellcolor{red!35}$38.7_{\scaleto{-7.2}{6pt}}$ & 21.6 & \cellcolor{apricot!50}$16.7_{\scaleto{-4.9}{6pt}}$ & 14.4 & \cellcolor{apricot!25}$11.9_{\scaleto{-2.5}{6pt}}$ & 60.7 & \cellcolor{red!50}$46.9_{\scaleto{-13.8}{6pt}}$ \\
umt-l-25M & 42.3 & \cellcolor{red!50}$30.6_{\scaleto{-11.7}{6pt}}$ & 43.6 & \cellcolor{red!50}$33.5_{\scaleto{-10.1}{6pt}}$ & 36.7 & \cellcolor{cadmiumorange!50}$30.4_{\scaleto{-6.3}{6pt}}$ & 19.4 & \cellcolor{cadmiumorange!35}$14.2_{\scaleto{-5.2}{6pt}}$ & 15.1 & \cellcolor{apricot!25}$12.8_{\scaleto{-2.3}{6pt}}$ & 60.8 & \cellcolor{red!50}$47.9_{\scaleto{-12.9}{6pt}}$ \\
\midrule
videomamba-vm-5M & 33.3 & \cellcolor{red!50}$23.0_{\scaleto{-10.3}{6pt}}$ & 37.1 & \cellcolor{red!35}$27.1_{\scaleto{-10.0}{6pt}}$ & 37.1 & \cellcolor{cadmiumorange!50}$30.1_{\scaleto{-7.0}{6pt}}$ & 17.6 & \cellcolor{apricot!50}$12.7_{\scaleto{-4.9}{6pt}}$ & 6.5 & \cellcolor{apricot!25}$5.6_{\scaleto{-0.9}{6pt}}$ & 47.6 & \cellcolor{red!50}$34.0_{\scaleto{-13.6}{6pt}}$ \\
videomamba-vm-17M & 34.9 & \cellcolor{red!35}$25.5_{\scaleto{-9.4}{6pt}}$ & 40.6 & \cellcolor{red!50}$28.9_{\scaleto{-11.7}{6pt}}$ & 40.4 & \cellcolor{red!35}$33.0_{\scaleto{-7.4}{6pt}}$ & 20.1 & \cellcolor{apricot!50}$15.5_{\scaleto{-4.6}{6pt}}$ & 7.7 & \cellcolor{apricot!25}$6.6_{\scaleto{-1.1}{6pt}}$ & 51.6 & \cellcolor{red!50}$38.3_{\scaleto{-13.3}{6pt}}$ \\
videomamba-vm-25M & 34.9 & \cellcolor{red!35}$25.5_{\scaleto{-9.4}{6pt}}$ & 41.4 & \cellcolor{red!50}$30.5_{\scaleto{-10.9}{6pt}}$ & 41.1 & \cellcolor{red!35}$33.8_{\scaleto{-7.3}{6pt}}$ & 20.4 & \cellcolor{apricot!50}$15.4_{\scaleto{-5.0}{6pt}}$ & 9.3 & \cellcolor{apricot!25}$7.9_{\scaleto{-1.4}{6pt}}$ & 53.2 & \cellcolor{red!50}$39.7_{\scaleto{-13.5}{6pt}}$ \\
\midrule
internvid-B-10M-FLT & 37.9 & \cellcolor{red!50}$25.4_{\scaleto{-12.5}{6pt}}$ & 28.6 & \cellcolor{red!50}$17.2_{\scaleto{-11.4}{6pt}}$ & 24.4 & \cellcolor{cadmiumorange!35}$18.8_{\scaleto{-5.6}{6pt}}$ & 17.0 & \cellcolor{cadmiumorange!50}$10.7_{\scaleto{-6.3}{6pt}}$ & 8.1 & \cellcolor{apricot!25}$5.9_{\scaleto{-2.2}{6pt}}$ & 48.9 & \cellcolor{red!50}$34.6_{\scaleto{-14.3}{6pt}}$ \\
internvid-B-200M & 38.1 & \cellcolor{red!50}$24.7_{\scaleto{-13.4}{6pt}}$ & 30.2 & \cellcolor{red!50}$19.4_{\scaleto{-10.8}{6pt}}$ & 26.2 & \cellcolor{cadmiumorange!50}$20.1_{\scaleto{-6.1}{6pt}}$ & 18.3 & \cellcolor{cadmiumorange!50}$11.9_{\scaleto{-6.4}{6pt}}$ & 8.6 & \cellcolor{apricot!25}$6.4_{\scaleto{-2.2}{6pt}}$ & 49.8 & \cellcolor{red!50}$35.8_{\scaleto{-14.0}{6pt}}$ \\
internvid-L-10M & 26.7 & \cellcolor{red!35}$18.7_{\scaleto{-8.0}{6pt}}$ & 22.6 & \cellcolor{cadmiumorange!50}$15.6_{\scaleto{-7.0}{6pt}}$ & 21.5 & \cellcolor{cadmiumorange!35}$16.3_{\scaleto{-5.2}{6pt}}$ & 11.4 & \cellcolor{apricot!50}$7.0_{\scaleto{-4.4}{6pt}}$ & 6.8 & \cellcolor{apricot!25}$5.5_{\scaleto{-1.3}{6pt}}$ & 35.9 & \cellcolor{red!50}$23.6_{\scaleto{-12.3}{6pt}}$ \\
internvid-L-WV10M & 26.5 & \cellcolor{red!35}$17.6_{\scaleto{-8.9}{6pt}}$ & 22.2 & \cellcolor{red!35}$13.8_{\scaleto{-8.4}{6pt}}$ & 23.1 & \cellcolor{cadmiumorange!35}$17.5_{\scaleto{-5.6}{6pt}}$ & 12.3 & \cellcolor{apricot!50}$7.6_{\scaleto{-4.7}{6pt}}$ & 6.7 & \cellcolor{apricot!25}$5.6_{\scaleto{-1.1}{6pt}}$ & 38.8 & \cellcolor{red!50}$25.9_{\scaleto{-12.9}{6pt}}$ \\
internvid-L-10M-DIV & 37.3 & \cellcolor{red!50}$24.2_{\scaleto{-13.1}{6pt}}$ & 26.9 & \cellcolor{red!50}$15.9_{\scaleto{-11.0}{6pt}}$ & 23.2 & \cellcolor{cadmiumorange!35}$17.7_{\scaleto{-5.5}{6pt}}$ & 15.7 & \cellcolor{cadmiumorange!35}$10.3_{\scaleto{-5.4}{6pt}}$ & 8.6 & \cellcolor{apricot!25}$6.7_{\scaleto{-1.9}{6pt}}$ & 48.9 & \cellcolor{red!50}$34.4_{\scaleto{-14.5}{6pt}}$ \\
internvid-L-10M-FLT & 38.7 & \cellcolor{red!50}$26.3_{\scaleto{-12.4}{6pt}}$ & 29.2 & \cellcolor{red!50}$18.8_{\scaleto{-10.4}{6pt}}$ & 24.5 & \cellcolor{cadmiumorange!35}$19.0_{\scaleto{-5.5}{6pt}}$ & 19.5 & \cellcolor{cadmiumorange!35}$13.5_{\scaleto{-6.0}{6pt}}$ & 9.4 & \cellcolor{apricot!25}$7.5_{\scaleto{-1.9}{6pt}}$ & 50.5 & \cellcolor{red!50}$36.2_{\scaleto{-14.3}{6pt}}$ \\
internvid-L-50M & 32.4 & \cellcolor{red!50}$22.0_{\scaleto{-10.4}{6pt}}$ & 26.5 & \cellcolor{red!35}$18.3_{\scaleto{-8.2}{6pt}}$ & 24.4 & \cellcolor{cadmiumorange!50}$18.0_{\scaleto{-6.4}{6pt}}$ & 17.8 & \cellcolor{cadmiumorange!50}$11.7_{\scaleto{-6.1}{6pt}}$ & 8.0 & \cellcolor{apricot!25}$6.4_{\scaleto{-1.6}{6pt}}$ & 45.7 & \cellcolor{red!50}$31.5_{\scaleto{-14.2}{6pt}}$ \\
internvid-L-200M & 38.2 & \cellcolor{red!50}$24.8_{\scaleto{-13.4}{6pt}}$ & 30.3 & \cellcolor{red!50}$20.2_{\scaleto{-10.1}{6pt}}$ & 28.7 & \cellcolor{cadmiumorange!50}$22.1_{\scaleto{-6.6}{6pt}}$ & 20.1 & \cellcolor{cadmiumorange!50}$13.7_{\scaleto{-6.4}{6pt}}$ & 11.0 & \cellcolor{apricot!25}$8.8_{\scaleto{-2.2}{6pt}}$ & 53.7 & \cellcolor{red!50}$39.0_{\scaleto{-14.7}{6pt}}$ \\

                \bottomrule
    \end{tabular}
    }
    \caption{\small{\textbf{Benchmarking video-language models in text-to-video retrieval on all six considered retrieval datasets.} We report accuracy on full test/val sets and our debiased UTD-splits. The accuracy differences with respect to the full test/val sets are color-coded. 
    \label{tab:sota_Text-to-Video_add}
    }}
    
\end{table*}

In this section, we present additional benchmarking results for state-of-the-art video models on our object-debiased UTD-splits. Specifically, we extend the analysis presented in Tables 5 and 6 of the main paper by including three considered classification datasets, namely Kinetics 600, Kinetics 700, and MiT, and three considered retrieval datasets, namely LSMDC, YouCook2, and Spoken-MiT. 
In~\cref{tab:sota_ActionClassification_add}, we provide the performance of selected video models on all classification datasets, evaluated both on the full test/val sets and on our debiased UTD-splits. The models were chosen based on their strong performance in Table 5 of the main paper. And in~\cref{tab:sota_Text-to-Video_add}, we present the evaluation results for video models across all considered retrieval datasets.

\subsection{Common Sense Bias}

\begin{table*}[]
    \renewcommand*{\arraystretch}{1.2}
    \setlength{\tabcolsep}{2pt}

    \resizebox{1\linewidth}{!}{
    
    \begin{tabular}{@{}l|cccc|cccc|cccc@{}}
    	\toprule
                 \multicolumn{13}{c}{\textbf{\cellcolor{darkpastelpurple!40}Action Classification Datasets}} \\
                 \midrule
                & \multicolumn{4}{c}{\cellcolor{darkpastelpurple!40}UCF} & \multicolumn{4}{c}{\cellcolor{darkpastelpurple!40}SSv2} & \multicolumn{4}{c}{\cellcolor{darkpastelpurple!40}K400} \\
                & seq.-of-f. & avg.-over-f. & max-score-f. & middle f. & seq.-of-f. & avg.-over-f. & max-score-f. & middle f.  & seq.-of-f. & avg.-over-f. & max-score-f. & middle f.  \\
                \midrule
                
obj+comp+act & 66.3 & \cellcolor{ForestGreen!50}$66.7_{\scaleto{+0.4}{6pt}}$ & \cellcolor{ForestGreen!50}$66.5_{\scaleto{+0.2}{6pt}}$ & \cellcolor{apricot!50}$61.3_{\scaleto{-5.0}{6pt}}$ & 6.4 & \cellcolor{ForestGreen!50}$6.8_{\scaleto{+0.4}{6pt}}$ & \cellcolor{ForestGreen!50}$7.4_{\scaleto{+1.0}{6pt}}$ & \cellcolor{apricot!50}$6.0_{\scaleto{-0.4}{6pt}}$ & 48.0 & \cellcolor{apricot!50}$46.6_{\scaleto{-1.4}{6pt}}$ & 48.0 & \cellcolor{cadmiumorange!50}$39.7_{\scaleto{-8.3}{6pt}}$ \\
objects & \cellcolor{apricot!50}$63.3_{\scaleto{-3.0}{6pt}}$ & \cellcolor{apricot!50}$61.9_{\scaleto{-4.4}{6pt}}$ & \cellcolor{apricot!50}$62.5_{\scaleto{-3.8}{6pt}}$ & \cellcolor{cadmiumorange!50}$57.4_{\scaleto{-8.9}{6pt}}$ & \cellcolor{apricot!50}$5.3_{\scaleto{-1.1}{6pt}}$ & \cellcolor{apricot!50}$4.6_{\scaleto{-1.8}{6pt}}$ & \cellcolor{apricot!50}$5.1_{\scaleto{-1.3}{6pt}}$ & \cellcolor{apricot!50}$4.0_{\scaleto{-2.4}{6pt}}$ & \cellcolor{apricot!50}$45.9_{\scaleto{-2.1}{6pt}}$ & \cellcolor{cadmiumorange!50}$41.0_{\scaleto{-7.0}{6pt}}$ & \cellcolor{cadmiumorange!50}$42.3_{\scaleto{-5.7}{6pt}}$ & \cellcolor{red!50}$35.0_{\scaleto{-13.0}{6pt}}$ \\
activities & \cellcolor{ForestGreen!50}$67.4_{\scaleto{+1.1}{6pt}}$ & \cellcolor{ForestGreen!50}$67.4_{\scaleto{+1.1}{6pt}}$ & \cellcolor{apricot!50}$65.7_{\scaleto{-0.6}{6pt}}$ & \cellcolor{cadmiumorange!50}$59.0_{\scaleto{-7.3}{6pt}}$ & 6.4 & \cellcolor{apricot!50}$6.0_{\scaleto{-0.4}{6pt}}$ & \cellcolor{apricot!50}$6.0_{\scaleto{-0.4}{6pt}}$ & \cellcolor{apricot!50}$4.9_{\scaleto{-1.5}{6pt}}$ & \cellcolor{apricot!50}$45.2_{\scaleto{-2.8}{6pt}}$ & \cellcolor{cadmiumorange!50}$42.1_{\scaleto{-5.9}{6pt}}$ & \cellcolor{cadmiumorange!50}$42.8_{\scaleto{-5.2}{6pt}}$ & \cellcolor{red!50}$31.3_{\scaleto{-16.7}{6pt}}$ \\
verbs & \cellcolor{red!50}$50.8_{\scaleto{-15.5}{6pt}}$ & \cellcolor{oucrimsonred!62}$41.5_{\scaleto{-24.8}{6pt}}$ & \cellcolor{oucrimsonred!62}$43.7_{\scaleto{-22.6}{6pt}}$ & \cellcolor{oucrimsonred!62}$31.1_{\scaleto{-35.2}{6pt}}$ & \cellcolor{apricot!50}$5.8_{\scaleto{-0.6}{6pt}}$ & \cellcolor{apricot!50}$4.4_{\scaleto{-2.0}{6pt}}$ & \cellcolor{apricot!50}$4.1_{\scaleto{-2.3}{6pt}}$ & \cellcolor{apricot!50}$3.1_{\scaleto{-3.3}{6pt}}$ & \cellcolor{oucrimsonred!62}$24.8_{\scaleto{-23.2}{6pt}}$ & \cellcolor{oucrimsonred!62}$16.1_{\scaleto{-31.9}{6pt}}$ & \cellcolor{oucrimsonred!62}$20.1_{\scaleto{-27.9}{6pt}}$ & \cellcolor{oucrimsonred!62}$10.9_{\scaleto{-37.1}{6pt}}$ \\
                \midrule  

                & \multicolumn{4}{c}{\cellcolor{darkpastelpurple!40}K600} & \multicolumn{4}{c}{\cellcolor{darkpastelpurple!40}K700} & \multicolumn{4}{c}{\cellcolor{darkpastelpurple!40}MiT} \\
                & seq.-of-f. & avg.-over-f. & max-score-f. & middle f.  & seq.-of-f. & avg.-over-f. & max-score-f. & middle f.  & seq.-of-f. & avg.-over-f. & max-score-f. & middle f. \\
                \midrule

obj+comp+act & 44.1 & \cellcolor{apricot!50}$42.1_{\scaleto{-2.0}{6pt}}$ & \cellcolor{apricot!50}$43.3_{\scaleto{-0.8}{6pt}}$ & \cellcolor{cadmiumorange!50}$35.6_{\scaleto{-8.5}{6pt}}$ & 39.0 & \cellcolor{apricot!50}$37.1_{\scaleto{-1.9}{6pt}}$ & \cellcolor{apricot!50}$38.7_{\scaleto{-0.3}{6pt}}$ & \cellcolor{cadmiumorange!50}$31.1_{\scaleto{-7.9}{6pt}}$ & 22.6 & \cellcolor{ForestGreen!50}$23.4_{\scaleto{+0.8}{6pt}}$ & \cellcolor{apricot!50}$21.4_{\scaleto{-1.2}{6pt}}$ & \cellcolor{apricot!50}$20.1_{\scaleto{-2.5}{6pt}}$ \\
objects & \cellcolor{apricot!50}$41.8_{\scaleto{-2.3}{6pt}}$ & \cellcolor{cadmiumorange!50}$36.5_{\scaleto{-7.6}{6pt}}$ & \cellcolor{cadmiumorange!50}$37.5_{\scaleto{-6.6}{6pt}}$ & \cellcolor{red!50}$31.0_{\scaleto{-13.1}{6pt}}$ & \cellcolor{apricot!50}$37.0_{\scaleto{-2.0}{6pt}}$ & \cellcolor{cadmiumorange!50}$32.2_{\scaleto{-6.8}{6pt}}$ & \cellcolor{cadmiumorange!50}$33.3_{\scaleto{-5.7}{6pt}}$ & \cellcolor{red!50}$26.7_{\scaleto{-12.3}{6pt}}$ & \cellcolor{apricot!50}$21.0_{\scaleto{-1.6}{6pt}}$ & \cellcolor{apricot!50}$19.4_{\scaleto{-3.2}{6pt}}$ & \cellcolor{apricot!50}$20.1_{\scaleto{-2.5}{6pt}}$ & \cellcolor{apricot!50}$17.6_{\scaleto{-5.0}{6pt}}$ \\
activities & \cellcolor{apricot!50}$41.4_{\scaleto{-2.7}{6pt}}$ & \cellcolor{cadmiumorange!50}$38.1_{\scaleto{-6.0}{6pt}}$ & \cellcolor{cadmiumorange!50}$38.5_{\scaleto{-5.6}{6pt}}$ & \cellcolor{red!50}$28.1_{\scaleto{-16.0}{6pt}}$ & \cellcolor{apricot!50}$36.7_{\scaleto{-2.3}{6pt}}$ & \cellcolor{cadmiumorange!50}$33.0_{\scaleto{-6.0}{6pt}}$ & \cellcolor{cadmiumorange!50}$33.8_{\scaleto{-5.2}{6pt}}$ & \cellcolor{red!50}$24.0_{\scaleto{-15.0}{6pt}}$ & \cellcolor{apricot!50}$21.0_{\scaleto{-1.6}{6pt}}$ & \cellcolor{apricot!50}$20.0_{\scaleto{-2.6}{6pt}}$ & \cellcolor{apricot!50}$20.1_{\scaleto{-2.5}{6pt}}$ & \cellcolor{cadmiumorange!50}$15.6_{\scaleto{-7.0}{6pt}}$ \\
verbs & \cellcolor{oucrimsonred!62}$21.4_{\scaleto{-22.7}{6pt}}$ & \cellcolor{oucrimsonred!62}$13.8_{\scaleto{-30.3}{6pt}}$ & \cellcolor{oucrimsonred!62}$16.7_{\scaleto{-27.4}{6pt}}$ & \cellcolor{oucrimsonred!62}$8.9_{\scaleto{-35.2}{6pt}}$ & \cellcolor{oucrimsonred!62}$17.6_{\scaleto{-21.4}{6pt}}$ & \cellcolor{oucrimsonred!62}$11.0_{\scaleto{-28.0}{6pt}}$ & \cellcolor{oucrimsonred!62}$13.6_{\scaleto{-25.4}{6pt}}$ & \cellcolor{oucrimsonred!62}$7.0_{\scaleto{-32.0}{6pt}}$ & \cellcolor{cadmiumorange!50}$16.2_{\scaleto{-6.4}{6pt}}$ & \cellcolor{red!50}$12.2_{\scaleto{-10.4}{6pt}}$ & \cellcolor{cadmiumorange!50}$14.1_{\scaleto{-8.5}{6pt}}$ & \cellcolor{red!50}$9.2_{\scaleto{-13.4}{6pt}}$ \\
                \midrule
                 \multicolumn{13}{c}{\textbf{\cellcolor{babyblueeyes!50}Text-to-Video Retrieval Datasets}} \\
                \midrule  

                &  \multicolumn{4}{c}{\cellcolor{babyblueeyes!50}MSR}  & \multicolumn{4}{c}{\cellcolor{babyblueeyes!50}DDM} & \multicolumn{4}{c}{\cellcolor{babyblueeyes!50}ActN} \\
                & seq.-of-f. & avg.-over-f. & max-score-f. & middle f.  & seq.-of-f. & avg.-over-f. & max-score-f. & middle f.  & seq.-of-f. & avg.-over-f. & max-score-f. & middle f. \\
                \midrule

obj+comp+act & 36.6 & \cellcolor{apricot!50}$31.9_{\scaleto{-4.7}{6pt}}$ & \cellcolor{apricot!50}$33.4_{\scaleto{-3.2}{6pt}}$ & \cellcolor{red!50}$23.4_{\scaleto{-13.2}{6pt}}$ & 27.2 & \cellcolor{apricot!50}$26.5_{\scaleto{-0.7}{6pt}}$ & \cellcolor{ForestGreen!50}$29.1_{\scaleto{+1.9}{6pt}}$ & \cellcolor{cadmiumorange!50}$19.8_{\scaleto{-7.4}{6pt}}$ & 26.6 & 26.6 & \cellcolor{cadmiumorange!50}$21.5_{\scaleto{-5.1}{6pt}}$ & \cellcolor{red!50}$13.5_{\scaleto{-13.1}{6pt}}$ \\
objects & \cellcolor{apricot!50}$32.1_{\scaleto{-4.5}{6pt}}$ & \cellcolor{cadmiumorange!50}$28.7_{\scaleto{-7.9}{6pt}}$ & \cellcolor{cadmiumorange!50}$29.7_{\scaleto{-6.9}{6pt}}$ & \cellcolor{red!50}$17.8_{\scaleto{-18.8}{6pt}}$ & \cellcolor{apricot!50}$27.0_{\scaleto{-0.2}{6pt}}$ & \cellcolor{apricot!50}$26.5_{\scaleto{-0.7}{6pt}}$ & \cellcolor{apricot!50}$25.5_{\scaleto{-1.7}{6pt}}$ & \cellcolor{red!50}$16.8_{\scaleto{-10.4}{6pt}}$ & \cellcolor{apricot!50}$24.8_{\scaleto{-1.8}{6pt}}$ & \cellcolor{apricot!50}$22.7_{\scaleto{-3.9}{6pt}}$ & \cellcolor{cadmiumorange!50}$17.8_{\scaleto{-8.8}{6pt}}$ & \cellcolor{red!50}$11.5_{\scaleto{-15.1}{6pt}}$ \\
activities & \cellcolor{red!50}$25.1_{\scaleto{-11.5}{6pt}}$ & \cellcolor{red!50}$22.4_{\scaleto{-14.2}{6pt}}$ & \cellcolor{red!50}$18.6_{\scaleto{-18.0}{6pt}}$ & \cellcolor{oucrimsonred!62}$11.6_{\scaleto{-25.0}{6pt}}$ & \cellcolor{cadmiumorange!50}$21.1_{\scaleto{-6.1}{6pt}}$ & \cellcolor{cadmiumorange!50}$21.0_{\scaleto{-6.2}{6pt}}$ & \cellcolor{cadmiumorange!50}$18.4_{\scaleto{-8.8}{6pt}}$ & \cellcolor{red!50}$11.3_{\scaleto{-15.9}{6pt}}$ & \cellcolor{cadmiumorange!50}$21.4_{\scaleto{-5.2}{6pt}}$ & \cellcolor{cadmiumorange!50}$17.8_{\scaleto{-8.8}{6pt}}$ & \cellcolor{red!50}$13.2_{\scaleto{-13.4}{6pt}}$ & \cellcolor{red!50}$8.4_{\scaleto{-18.2}{6pt}}$ \\
verbs & \cellcolor{oucrimsonred!62}$10.5_{\scaleto{-26.1}{6pt}}$ & \cellcolor{oucrimsonred!62}$8.7_{\scaleto{-27.9}{6pt}}$ & \cellcolor{oucrimsonred!62}$7.8_{\scaleto{-28.8}{6pt}}$ & \cellcolor{oucrimsonred!62}$4.2_{\scaleto{-32.4}{6pt}}$ & \cellcolor{oucrimsonred!62}$7.0_{\scaleto{-20.2}{6pt}}$ & \cellcolor{oucrimsonred!62}$6.0_{\scaleto{-21.2}{6pt}}$ & \cellcolor{oucrimsonred!62}$5.5_{\scaleto{-21.7}{6pt}}$ & \cellcolor{oucrimsonred!62}$3.5_{\scaleto{-23.7}{6pt}}$ & \cellcolor{red!50}$7.4_{\scaleto{-19.2}{6pt}}$ & \cellcolor{oucrimsonred!62}$5.1_{\scaleto{-21.5}{6pt}}$ & \cellcolor{oucrimsonred!62}$3.9_{\scaleto{-22.7}{6pt}}$ & \cellcolor{oucrimsonred!62}$2.5_{\scaleto{-24.1}{6pt}}$ \\

            \midrule  

                & \multicolumn{4}{c}{\cellcolor{babyblueeyes!50}LSMDC} & \multicolumn{4}{c}{\cellcolor{babyblueeyes!50}YC2}  & \multicolumn{4}{c}{\cellcolor{babyblueeyes!50}S-MiT}   \\
                & seq.-of-f. & avg.-over-f. & max-score-f. & middle f. & seq.-of-f. & avg.-over-f. & max-score-f. & middle f.   & seq.-of-f. & avg.-over-f. & max-score-f. & middle f. \\
                \midrule

obj+comp+act & 17.0 & \cellcolor{apricot!50}$16.4_{\scaleto{-0.6}{6pt}}$ & \cellcolor{ForestGreen!50}$18.3_{\scaleto{+1.3}{6pt}}$ & \cellcolor{apricot!50}$12.7_{\scaleto{-4.3}{6pt}}$ & 8.4 & \cellcolor{ForestGreen!50}$8.6_{\scaleto{+0.2}{6pt}}$ & \cellcolor{ForestGreen!50}$8.9_{\scaleto{+0.5}{6pt}}$ & \cellcolor{apricot!50}$6.1_{\scaleto{-2.3}{6pt}}$ & 45.9 & \cellcolor{apricot!50}$43.8_{\scaleto{-2.1}{6pt}}$ & \cellcolor{ForestGreen!50}$46.1_{\scaleto{+0.2}{6pt}}$ & \cellcolor{red!50}$35.5_{\scaleto{-10.4}{6pt}}$ \\
objects & \cellcolor{apricot!50}$13.6_{\scaleto{-3.4}{6pt}}$ & \cellcolor{cadmiumorange!50}$11.7_{\scaleto{-5.3}{6pt}}$ & \cellcolor{apricot!50}$12.3_{\scaleto{-4.7}{6pt}}$ & \cellcolor{cadmiumorange!50}$9.2_{\scaleto{-7.8}{6pt}}$ & \cellcolor{apricot!50}$7.9_{\scaleto{-0.5}{6pt}}$ & \cellcolor{apricot!50}$6.2_{\scaleto{-2.2}{6pt}}$ & \cellcolor{apricot!50}$7.1_{\scaleto{-1.3}{6pt}}$ & \cellcolor{apricot!50}$4.9_{\scaleto{-3.5}{6pt}}$ & \cellcolor{red!50}$29.8_{\scaleto{-16.1}{6pt}}$ & \cellcolor{red!50}$27.5_{\scaleto{-18.4}{6pt}}$ & \cellcolor{red!50}$26.4_{\scaleto{-19.5}{6pt}}$ & \cellcolor{oucrimsonred!62}$18.2_{\scaleto{-27.7}{6pt}}$ \\
activities & \cellcolor{apricot!50}$14.7_{\scaleto{-2.3}{6pt}}$ & \cellcolor{apricot!50}$13.6_{\scaleto{-3.4}{6pt}}$ & \cellcolor{cadmiumorange!50}$11.4_{\scaleto{-5.6}{6pt}}$ & \cellcolor{cadmiumorange!50}$7.8_{\scaleto{-9.2}{6pt}}$ & \cellcolor{apricot!50}$8.1_{\scaleto{-0.3}{6pt}}$ & \cellcolor{apricot!50}$7.3_{\scaleto{-1.1}{6pt}}$ & \cellcolor{apricot!50}$6.3_{\scaleto{-2.1}{6pt}}$ & \cellcolor{apricot!50}$4.1_{\scaleto{-4.3}{6pt}}$ & \cellcolor{apricot!50}$41.1_{\scaleto{-4.8}{6pt}}$ & \cellcolor{cadmiumorange!50}$38.4_{\scaleto{-7.5}{6pt}}$ & \cellcolor{cadmiumorange!50}$37.8_{\scaleto{-8.1}{6pt}}$ & \cellcolor{red!50}$29.3_{\scaleto{-16.6}{6pt}}$ \\
verbs & \cellcolor{red!50}$5.5_{\scaleto{-11.5}{6pt}}$ & \cellcolor{red!50}$6.0_{\scaleto{-11.0}{6pt}}$ & \cellcolor{red!50}$4.5_{\scaleto{-12.5}{6pt}}$ & \cellcolor{red!50}$3.0_{\scaleto{-14.0}{6pt}}$ & \cellcolor{cadmiumorange!50}$1.2_{\scaleto{-7.2}{6pt}}$ & \cellcolor{cadmiumorange!50}$1.1_{\scaleto{-7.3}{6pt}}$ & \cellcolor{cadmiumorange!50}$1.0_{\scaleto{-7.4}{6pt}}$ & \cellcolor{cadmiumorange!50}$0.6_{\scaleto{-7.8}{6pt}}$ & \cellcolor{oucrimsonred!62}$13.1_{\scaleto{-32.8}{6pt}}$ & \cellcolor{oucrimsonred!62}$9.5_{\scaleto{-36.4}{6pt}}$ & \cellcolor{oucrimsonred!62}$8.8_{\scaleto{-37.1}{6pt}}$ & \cellcolor{oucrimsonred!62}$4.8_{\scaleto{-41.1}{6pt}}$ \\
                \bottomrule
    \end{tabular}
    }
    \caption{\small{
    \textbf{Evaluation of common sense bias with respect to all 16 conceptual-temporal combinations on all 12 considered datasets.} We color code with respect to the difference to objects+composition+activities (obj+comp+act) concepts in sequence-of-frames (seq.-of-f.) temporal setup.
    \label{tab:full_common_sense}
    }}
    
\end{table*} 

In~\cref{tab:full_common_sense}, we present additional common sense bias results for all 16 conceptual-temporal combinations across the 12 datasets considered. The observed effects align with results discussed in the main paper. Specifically, the overall classification performance drops only slightly when predictions are based solely on objects compared to the objects+composition+activities setup across most datasets.

\subsection{Ablation Study}

\begin{table}
    
    \centering
    \resizebox{1\linewidth}{!}{
    
    \begin{tabular}{@{}l|cc|cc@{}}
    	\toprule
                 \multirow{2}{*}{Text Embedding Model} & \multicolumn{2}{c}{\cellcolor{darkpastelpurple!40}Action Classification} & \multicolumn{2}{c}{\cellcolor{babyblueeyes!50}Text-to-Video Retrieval}  \\
                & \cellcolor{darkpastelpurple!40}UCF & \cellcolor{darkpastelpurple!40}SSv2 & \cellcolor{babyblueeyes!50}MSR & \cellcolor{babyblueeyes!50}DDM \\
                \midrule

CLIP~\cite{radford2021learning} text encoder & 51.1 & 1.8 & 6.1 & 7.0\\
Long-CLIP~\cite{zhang2025long} text encoder & 48.5 & 2.8 & 24.0 & 19.0 \\
E5-Mistral-7B-Instruct~\cite{wang2023improving} & 65.6 & 6.2 & 35.1 & 26.7 \\
\textbf{SFR-Embedding-Mistral}~\cite{meng2024sfrembedding}  & \textbf{66.3} & \textbf{6.4} & \textbf{36.6}  & \textbf{27.2} \\
                \bottomrule
    \end{tabular}
    }
    \caption{\small{
    \textbf{Ablation on text embedding model.}  Evaluation is performed in objects+composition+activities concepts in sequence-of-frames temporal setup on the full test/validation splits of the respective datasets. We report accuracy for classification and recall@1 for retrieval. Selected option is bolded.
    \label{tab:text_model_comp}
    }}
    
\end{table} 

\begin{table}
    
    \centering
    \resizebox{1\linewidth}{!}{
    
    \begin{tabular}{@{}l|cc|cc@{}}
    	\toprule
                 \multirow{2}{*}{Vision-Language Model} & \multicolumn{2}{c}{\cellcolor{darkpastelpurple!40}Action Classification} & \multicolumn{2}{c}{\cellcolor{babyblueeyes!50}Text-to-Video Retrieval}  \\
                & \cellcolor{darkpastelpurple!40}UCF & \cellcolor{darkpastelpurple!40}SSv2 & \cellcolor{babyblueeyes!50}MSR & \cellcolor{babyblueeyes!50}DDM \\
                \midrule

LLaVA-v1.5-7B~\cite{liu2024improved} & 61.8 & 5.9 & 35.5 & \textbf{28.7} \\
\textbf{LLaVA-1.6-Mistral-7B}~\cite{liu2024llavanext}  & \textbf{66.3} & \textbf{6.4} & \textbf{36.6}  & 27.2 \\
                \bottomrule
    \end{tabular}
    }
    \caption{\small{
    \textbf{Ablation on vision-language model.}  Evaluation is performed in objects+composition+activities concepts in sequence-of-frames temporal setup on the full test/validation splits of the respective datasets. We report accuracy for classification and recall@1 for retrieval. Selected option is bolded.
    \label{tab:vlm_comp}
    }}
    
\end{table} 

\begin{table*}
    
    \centering
    \resizebox{1\linewidth}{!}{
    
    \begin{tabular}{@{}l|cccccc|cccccc@{}}
    	\toprule
                 & \multicolumn{6}{c}{\cellcolor{darkpastelpurple!40}Action Classification} & \multicolumn{6}{c}{\cellcolor{babyblueeyes!50}Text-to-Video Retrieval}  \\
                & \cellcolor{darkpastelpurple!40}UCF & \cellcolor{darkpastelpurple!40}SSv2 & \cellcolor{darkpastelpurple!40}K400 & \cellcolor{darkpastelpurple!40}K600 & \cellcolor{darkpastelpurple!40}K700 & \cellcolor{darkpastelpurple!40}MiT & \cellcolor{babyblueeyes!50}MSR & \cellcolor{babyblueeyes!50}DDM & \cellcolor{babyblueeyes!50}ANet & \cellcolor{babyblueeyes!50}LSMDC & \cellcolor{babyblueeyes!50}YC2 & \cellcolor{babyblueeyes!50}S-MiT \\
                \midrule
CLIP ViT-B/32        & \underline{67.7} & 1.8 & 46.7 & 41.3 & 34.8 & 18.6 & 31.4 & 26.3 & 20.4 & 14.3 & 4.9 & 34.7 \\
CLIP ViT-L/14        & \textbf{75.9} & \underline{2.8} & \textbf{58.7} & \textbf{54.7} & \textbf{48.6 }& \textbf{23.7} & \underline{36.3} & \textbf{29.6} & \underline{26.2} & \textbf{19.9} & \underline{8.1} & \underline{40.4} \\
Ours (obj+comp+act, sequence-of-frames) & 66.3 & \textbf{6.4} & \underline{48.0} & \underline{44.1} & \underline{39.0} & \underline{22.6} & \textbf{36.6} & \underline{27.2} & \textbf{26.6} & \underline{17.0} & \textbf{8.4} & \textbf{45.9} \\
                \bottomrule
    \end{tabular}
    }
    \caption{\small{
    \textbf{Comparison in zero-shot action classification and text-to-video retrieval.} Note, that our model uses only textual descriptions of video. Evaluation is performed on full test/val splits of the respective datasets. We report accuracy for classification and Recall@1 for retrieval. 
    \label{tab:clip_comp}
    }}
    
\end{table*}

As discussed in the main paper, to estimate representation bias, we design a strong model that performs action classification and text-to-video retrieval based solely on the textual descriptions of videos.  Throughout our pipeline, we utilize state-of-the-art models, namely LLaVA-1.6-Mistral-7B~\cite{liu2024llavanext} as the VLM, Mistral-7B-Instruct-v0.2~\cite{jiang2023mistral} as the LLM, and SFR-Embedding-Mistral~\cite{meng2024sfrembedding} for text embedding model.  In this section, we provide additional analysis of our model. 

\noindent\textbf{Text Embedding Model.} First, in~\cref{tab:text_model_comp}, we compare four strong models for text encoding ($\chi$ in Figure 3 of the main paper). Namely, we consider large versions of the CLIP~\cite{radford2021learning} text encoder and LongCLIP~\cite{zhang2025long} text encoder, which extends CLIP to better handle long-text inputs. Additionally, we examine two LLM-based text embedding models: E5-Mistral-7B-Instruct~\cite{wang2023improving} and its fine-tuned version, SFR-Embedding-Mistral~\cite{meng2024sfrembedding}, trained on more data.
As shown in~\cref{tab:text_model_comp}, the text embedding models E5-Mistral-7B-Instruct and SFR-Embedding-Mistral, both pretrained on large text datasets to effectively encode text for tasks such as information retrieval, outperform the CLIP-based text embedding models. In our pipeline, we employ the best-performing SFR-Embedding-Mistral model.

\noindent\textbf{Vision-Language Model.} Next, in~\ref{tab:vlm_comp}, we ablate two VLM models for extracting detailed textual descriptions. Specifically, we evaluate LLaVA-v1.5-7B~\cite{liu2024improved} and LLaVA-1.6-Mistral-7B~\cite{liu2024llavanext}, finding that the latter model achieves the best performance.

\noindent\textbf{Comparison to CLIP.} 
Finally, we evaluate how well our model, which relies solely on textual descriptions, performs for zero-shot video classification and retrieval compared to the CLIP baseline~\cite{radford2021learning} The results in \cref{tab:clip_comp} demonstrate that our model generally outperforms CLIP ViT-B/32 and performs almost on par with the CLIP ViT-L/14 backbone. This highlights the feasibility of performing action classification and video retrieval based purely on textual descriptions.
We further observe that caption-based retrieval performs better compared to caption-based action classification. We attribute this to the fact that captions capture more specific aspects of individual videos, which are also reflected in the generated captions.

\subsection{Class Distribution in UTD-Splits}
\begin{figure*}[]
\centering

\begin{subfigure}[t]{0.9\linewidth}
\includegraphics[width=1\textwidth]{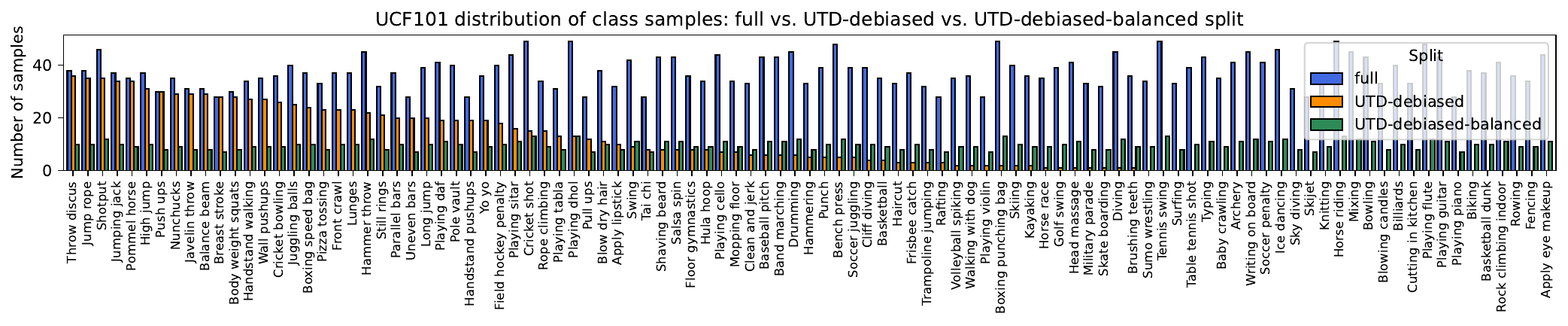}
\end{subfigure}

\begin{subfigure}[t]{0.9\linewidth}
\includegraphics[width=1\textwidth]{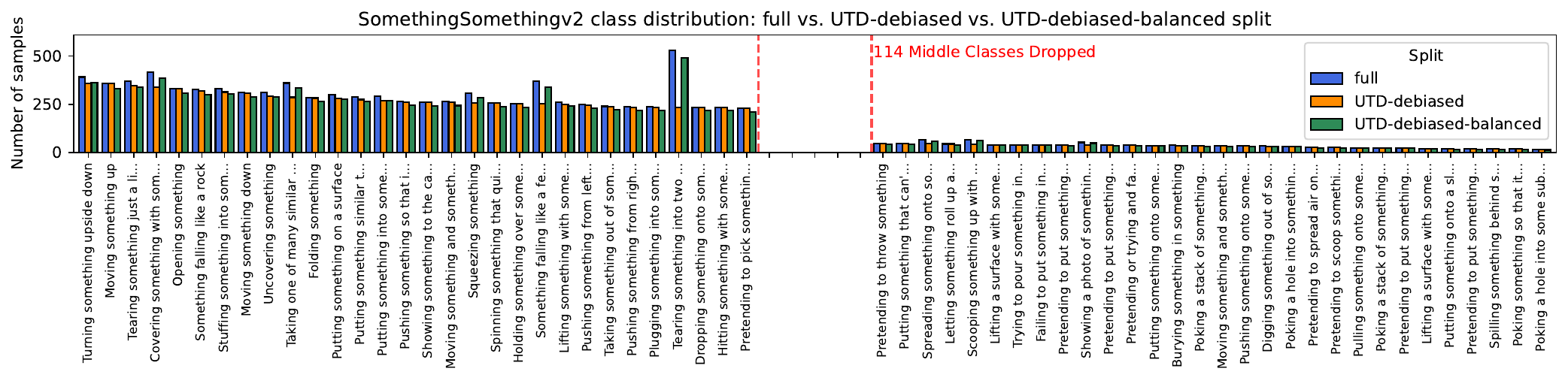}
\end{subfigure}%

\begin{subfigure}[t]{0.9\linewidth}
\includegraphics[width=1\textwidth]{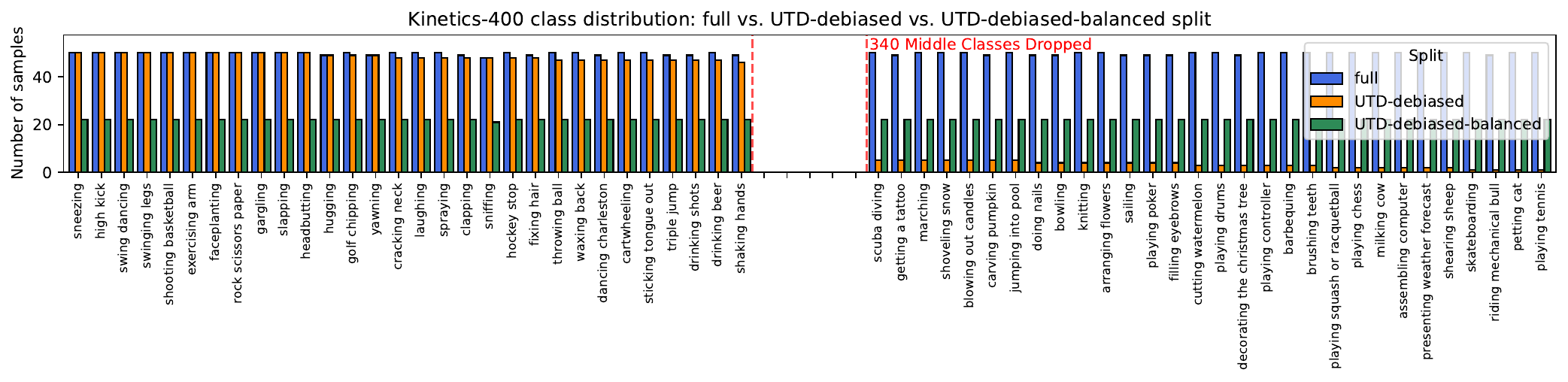}
\end{subfigure}%

\begin{subfigure}[t]{0.9\linewidth}
\includegraphics[width=1\textwidth]{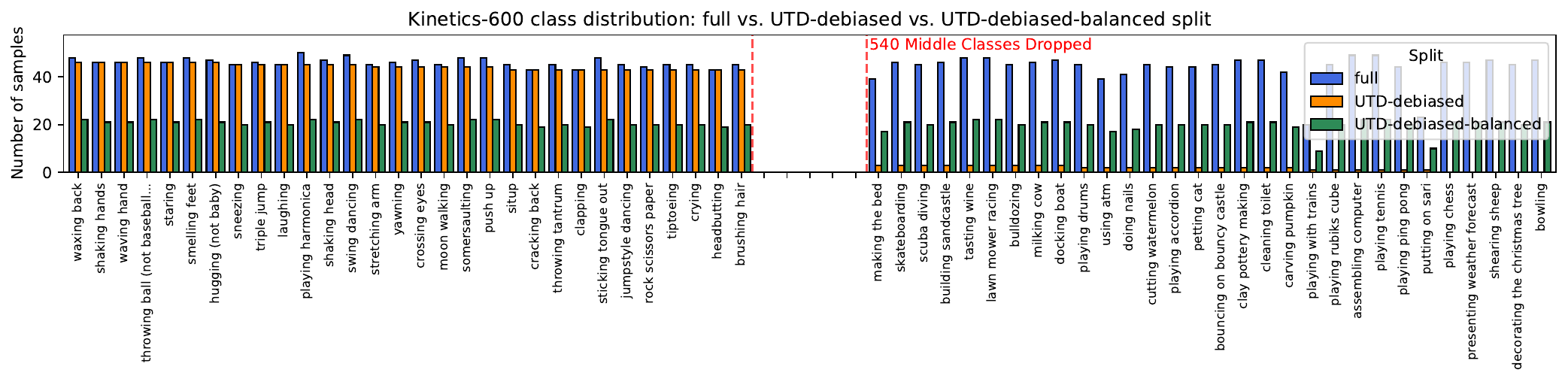}
\end{subfigure}%

\begin{subfigure}[t]{0.9\linewidth}
\includegraphics[width=1\textwidth]{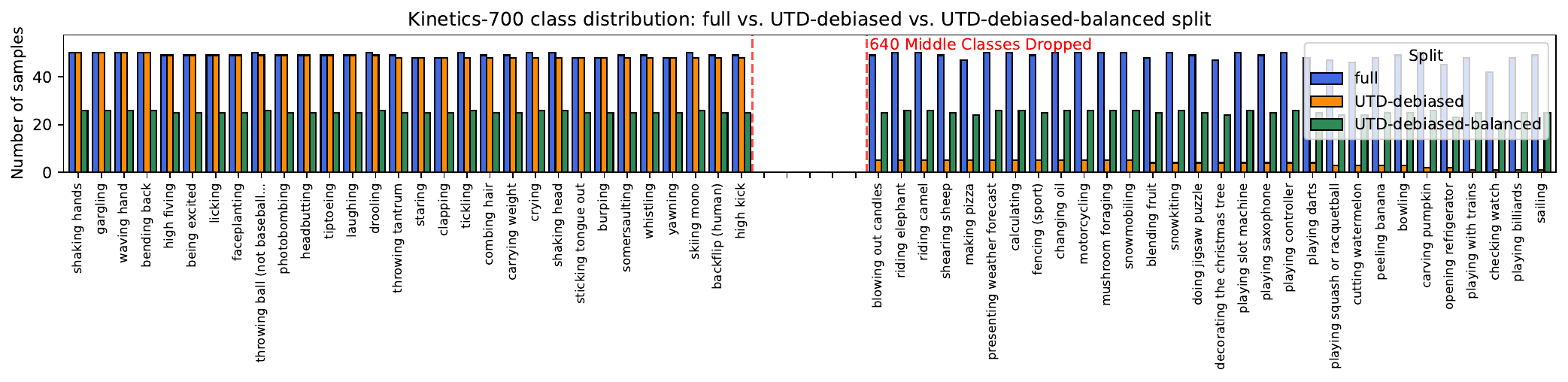}
\end{subfigure}%

\begin{subfigure}[t]{0.9\linewidth}
\includegraphics[width=1\textwidth]{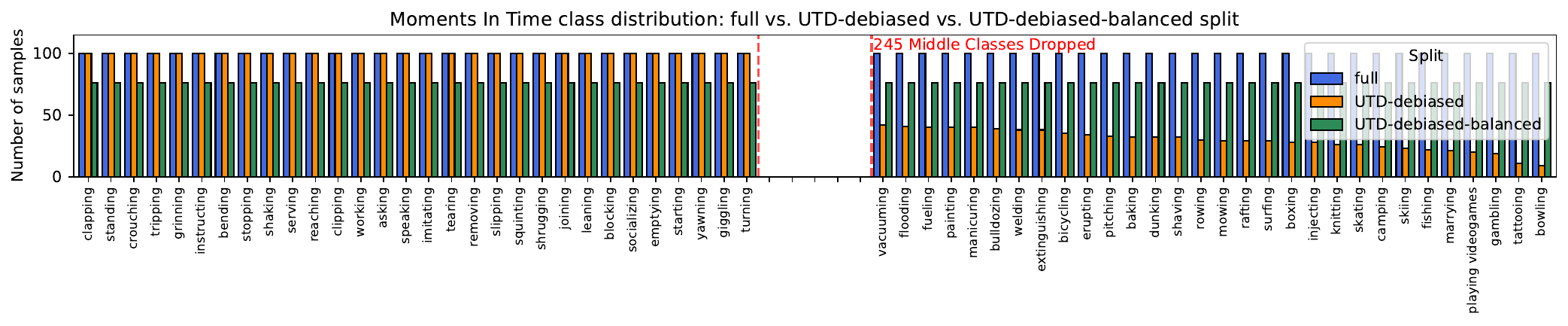}
\end{subfigure}%

\caption{
Comparison of class distribution in full test/val split vs. UTD-debiased split vs. UTD-debiased-balanced split for six considered classification datasets. \label{fig:UTD_class}} 
\end{figure*}

In \cref{fig:UTD_class}, we show the class distribution in our UTD- and UTD-balanced test/validation splits, in comparison with the original class distribution in the full test/validation splits. We observe that the results align well with our expectations, for example, the class ``Apply Eye Makeup'' in UCF-101 is significantly reduced in the UTD-split due to a strong object bias.

\section{Additional Implementation Details}
\label{sec:add_imp_details}

In this section, we provide further details about our UTD method.

\subsection{Obtaining Textual Descriptions}

We used a few-shot in-context learning strategy~\cite{brown2020language} when prompting LLM to extract objects, activities, and verbs from objects+composition+activities descriptions. Namely, we used 3-shots for objects and activities and 5-shots for verbs. We also did simple postprocessing of LLM output. Since the LLM is prompted to output an enumerated list (see prompts in~\cref{tab:prompts_concepts}), we delete numeration and delete any output in brackets.

We list the prompts used in our work to obtain textual descriptions of frames for different concept categories, namely objects+composition+activities ($d_{n,i}$), objects ($o_{n,i}$), activities ($a_{n,i}$), and verbs ($\nu_{n,i}$), in~\cref{tab:prompts_concepts}

\subsection{Getting Text Embeddings}

Following~\cite{meng2024sfrembedding}, we used the prompt template ``\lstinline{Instruct: <instruction>\nQuery: <input text>}'' to obtain text embeddings for various input text descriptions using the SFR-Embedding-Mistral model~\cite{meng2024sfrembedding}. In~\cref{tab:prompt_classification}, we provide the instructions used for action classification in different settings, and in~\cref{tab:prompt_retrieval}, we present instructions used for text-to-video retrieval.

\subsection{Unbiasing Datasets}

As stated in the main paper, we debias the test/val sets of the considered datasets by excluding samples that are classified or retrieved correctly ($M(v_n, \phi) = 1$) based on \textit{object sequence-of-frame} representation.

For text-to-video retrieval, we debias datasets with respect to \textit{common-sense bias}.
To minimize the impact of random fluctuations in the process, we prompt the text embedding models with three different prompts for query captions and three different prompts for object textual descriptions for videos, generating three embeddings for each query caption and for each video. Prompts are reported in~\cref{tab:instructions_unbiasing}. Consequently, we perform zero-shot text-to-video retrieval inference using all nine combinations of query embeddings and video embeddings, resulting in nine predictions. We exclude a text query from the test/val set only if all nine predictions agree on the correct Top-1 retrieval for the corresponding video.

For action classification, we debias datasets with respect to \textit{dataset bias}. Specifically, we generate three different embeddings for the object textual descriptions of each video using three different instructions for the text embedding model (\cref{tab:instructions_unbiasing}), resulting in three sets of video embeddings. Using each set of embeddings, we further train three linear classifiers on bootstrapped training sets (we sample training sets using sampling replacement to match the original training set size). A sample is excluded from the test/val set only if all nine models (across the three embedding sets and three linear models) agree on the correct Top-1 classification. The percentage of removed samples is determined automatically based on the extent of object representation bias in the dataset. 

Since debiasing may disproportionately remove certain label classes in classification datasets, we additionally construct balanced UTD splits. While maintaining the total number of removed samples as in the original debiasing method, we adjust the number of samples removed from each class based on their average confidence (across nine models) to preserve the original class proportions.

\noindent\textbf{Reliability of filtering biased samples.}
To summarize, we ensure robust and reliable filtering of biased samples by leveraging state-of-the-art VLMs and LLMs, applying prompt engineering, and employing an in-context learning strategy to extract specific concepts (e.g., objects) while minimizing leakage from unrelated concepts. We also conduct a user study to validate the reliability of the extracted concepts. To further mitigate false positives, we aggregate predictions across nine different prompt/model combinations.

\subsection{Benchmarking Video Models}

\begin{table*}[]
\centering
\resizebox{1\linewidth}{!}{
\begin{tabular}{l l l l}
\toprule
\textbf{Model} & \textbf{Backbone} & \textbf{Pretraining Datasets} & \textbf{Finetuning Datasets} \\
\midrule
VideoMAE-B-K400~\cite{tong2022videomae} & ViT-B/16 & Kinetics-400 (w/o labels)~\cite{Kay2017Kinetics} & - \\
VideoMAE-B-UH~\cite{tong2022videomae} & ViT-B/16 & UnlabeledHybrid~\cite{wang2023videomae}: K700~\cite{Carreira2019Kinetics700} + WebVid2M~\cite{bain2021frozen} + SS~\cite{Goyal_2017_ICCV} + AVA~\cite{gu2018ava} + Instagram (collected) & - \\
VideoMAE-L-UH~\cite{tong2022videomae} & ViT-L/14 & UnlabeledHybrid~\cite{wang2023videomae}: K700~\cite{Carreira2019Kinetics700} + WebVid2M~\cite{bain2021frozen} + SS~\cite{Goyal_2017_ICCV} + AVA~\cite{gu2018ava} + Instagram (collected) & - \\
VideoMAE-H-UH~\cite{tong2022videomae} & ViT-H/16 & UnlabeledHybrid~\cite{wang2023videomae}: K700~\cite{Carreira2019Kinetics700} + WebVid2M~\cite{bain2021frozen} + SS~\cite{Goyal_2017_ICCV} + AVA~\cite{gu2018ava} + Instagram (collected) & - \\
\midrule
VideoMAEv2-B-K710-fnK710~\cite{wang2023videomae} & ViT-B/16 & Kinetics-710~\cite{li2022uniformerv2} (Kinetics-400~\cite{Kay2017Kinetics} + Kinetics-600~\cite{Carreira2018Kinetics600} + Kinetics-700~\cite{Carreira2019Kinetics700}) (w/o labels) & Kinetics-710~\cite{li2022uniformerv2} \\
\midrule
AllInOne-B-WV2M+CC~\cite{wang2023all} & ViT-B/16 & WebVid2M~\cite{bain2021frozen} + CC3M~\cite{sharma2018conceptual} & - \\
AllInOne-B-WV2M+HT~\cite{wang2023all} & ViT-B/16 & WebVid2M~\cite{bain2021frozen} + HowTo100M~\cite{miech2019howto100m} & - \\
AllInOne-B-WV2M+HT+CC+YTT+~\cite{wang2023all} & ViT-B/16 & WebVid2M~\cite{bain2021frozen} + HowTo100M~\cite{miech2019howto100m} + YTT~\cite{zellers2021merlot} + CC3M~\cite{sharma2018conceptual} + CC12M~\cite{changpinyo2021conceptual} + COCO~\cite{lin2014microsoft} + VG~\cite{krishna2017visual} + SBU~\cite{ordonez2011im2text} & - \\
\midrule
UMT-B-K710~\cite{li2023unmasked} & UMT-B/16 & Kinetics-710~\cite{li2022uniformerv2} (w/o labels) & - \\
UMT-B-fnK710~\cite{li2023unmasked} & UMT-B/16 & Kinetics-710~\cite{li2022uniformerv2} (w/o labels) & Kinetics-710~\cite{li2022uniformerv2} \\
UMT-L-K710~\cite{li2023unmasked} & UMT-L/16 & Kinetics-710~\cite{li2022uniformerv2} (w/o labels) & - \\
UMT-L-fnK710~\cite{li2023unmasked} & UMT-L/16 & Kinetics-710~\cite{li2022uniformerv2} (w/o labels) & Kinetics-710~\cite{li2022uniformerv2} \\
UMT-B-5M~\cite{li2023unmasked} & UMT-B/16 & Kinetics-710~\cite{li2022uniformerv2} (w/o labels) + WebVid2M~\cite{bain2021frozen} + CC3M~\cite{sharma2018conceptual} & - \\
UMT-B-17M~\cite{li2023unmasked} & UMT-B/16 & Kinetics-710~\cite{li2022uniformerv2} (w/o labels) + WebVid2M~\cite{bain2021frozen} + CC3M~\cite{sharma2018conceptual} + CC12M~\cite{changpinyo2021conceptual} + COCO~\cite{lin2014microsoft} + VG~\cite{krishna2017visual} + SBU~\cite{ordonez2011im2text} & - \\
UMT-B-25M~\cite{li2023unmasked} & UMT-B/16 & Kinetics-710~\cite{li2022uniformerv2} (w/o labels) + WebVid10M~\cite{bain2021frozen} + CC3M~\cite{sharma2018conceptual} + CC12M~\cite{changpinyo2021conceptual} + COCO~\cite{lin2014microsoft} + VG~\cite{krishna2017visual} + SBU~\cite{ordonez2011im2text} & - \\
UMT-L-5M~\cite{li2023unmasked} & UMT-L/16 & Kinetics-710~\cite{li2022uniformerv2} (w/o labels) + WebVid2M~\cite{bain2021frozen} + CC3M~\cite{sharma2018conceptual} & - \\
UMT-L-17M~\cite{li2023unmasked} & UMT-L/16 & Kinetics-710~\cite{li2022uniformerv2} (w/o labels) + WebVid2M~\cite{bain2021frozen} + CC3M~\cite{sharma2018conceptual} + CC12M~\cite{changpinyo2021conceptual} + COCO~\cite{lin2014microsoft} + VG~\cite{krishna2017visual} + SBU~\cite{ordonez2011im2text} & - \\
UMT-L-25M~\cite{li2023unmasked} & UMT-L/16 & Kinetics-710~\cite{li2022uniformerv2} (w/o labels) + WebVid10M~\cite{bain2021frozen} + CC3M~\cite{sharma2018conceptual} + CC12M~\cite{changpinyo2021conceptual} + COCO~\cite{lin2014microsoft} + VG~\cite{krishna2017visual} + SBU~\cite{ordonez2011im2text} & - \\
\midrule
VideoMamba-VM-K400~\cite{li2024videomamba} & VideoMamba-M & Kinetics-400~\cite{Kay2017Kinetics} (w/o labels) & - \\
VideoMamba-VM-5M~\cite{li2024videomamba} & VideoMamba-M & Kinetics-400~\cite{Kay2017Kinetics} (w/o labels) + WebVid2M~\cite{bain2021frozen} + CC3M~\cite{sharma2018conceptual} & - \\
VideoMamba-VM-17M~\cite{li2024videomamba} & VideoMamba-M & Kinetics-400~\cite{Kay2017Kinetics} (w/o labels) + WebVid2M~\cite{bain2021frozen} + CC3M~\cite{sharma2018conceptual} + CC12M~\cite{changpinyo2021conceptual} + COCO~\cite{lin2014microsoft} + VG~\cite{krishna2017visual} + SBU~\cite{ordonez2011im2text} & - \\
VideoMamba-VM-25M~\cite{li2024videomamba} & VideoMamba-M & Kinetics-400~\cite{Kay2017Kinetics} (w/o labels) + WebVid10M~\cite{bain2021frozen} + CC3M~\cite{sharma2018conceptual} + CC12M~\cite{changpinyo2021conceptual} + COCO~\cite{lin2014microsoft} + VG~\cite{krishna2017visual} + SBU~\cite{ordonez2011im2text} & - \\
\midrule
InternVid-B-10M-FLT~\cite{wang2023internvid} & ViCLIP-B/16 & InternVid-10M-FLT~\cite{wang2023internvid} & - \\
InternVid-B-200M~\cite{wang2023internvid} & ViCLIP-B/16 & InternVid-200M~\cite{wang2023internvid} & - \\
InternVid-L-10M~\cite{wang2023internvid} & ViCLIP-L/14 & InternVid-10M~\cite{wang2023internvid} & - \\
InternVid-L-WebVid10M~\cite{wang2023internvid} & ViCLIP-L/14 & WebVid10M~\cite{bain2021frozen} & - \\
InternVid-L-10M-DIV~\cite{wang2023internvid} & ViCLIP-L/14 & InternVid-10M-DIV~\cite{wang2023internvid} & - \\
InternVid-L-10M-FLT~\cite{wang2023internvid} & ViCLIP-L/14 & InternVid-10M-FLT~\cite{wang2023internvid} & - \\
InternVid-L-50M~\cite{wang2023internvid} & ViCLIP-L/14 & InternVid-50M~\cite{wang2023internvid} & - \\
InternVid-L-200M~\cite{wang2023internvid} & ViCLIP-L/14 & InternVid-200M~\cite{wang2023internvid} & - \\
\bottomrule
\end{tabular}
}
\caption{Overview of the 30 models analyzed in this paper, including their backbones, pretraining datasets, and supervised finetuning datasets.\label{tab:methods-details}}
\end{table*}

Table~\ref{tab:methods-details} summarizes all 30 video models evaluated in this work, detailing their architectural backbones, pretraining datasets, and supervised finetuning setups.

Further, we provide additional implementation details on our evaluation setup of video backbones in action classification and text-to-video retrieval. To evaluate a video backbone in action classification, following the VideoGLUE benchmark~\cite{yuan2023videoglue}, we train a classification model with the corresponding frozen video backbone and single-layer pooler head~\cite{yuan2023videoglue} with one classification linear layer as described in the main paper.
For training and evaluation, we use 8 uniformly sample frames as inputs, however, for video backbones that create 3D tokens with a stride over frames, such as VideoMAE~\cite{tong2022videomae,wang2023videomae}, we respectively scale the number of input frames, namely, we use 16 frames for VideoMAE and VideoMAEv2, to ensure that models use the same number of tokens for the same model size. We train a model for 50 epochs with an AdamW optimizer~\cite{loshchilov2017decoupled} and a learning rate of 0.001. We use a cosine weight decay scheduler with five epochs warmup. We follow the augmentation pipeline of the VideoGLUE benchmark~\cite{yuan2023videoglue}. We do not use multi-crop evaluation to simplify and standardize the evaluation setting for all datasets. 

For text-to-video retrieval, we evaluate the zero-shot capabilities of the text-video models and use the respective backbones without fine-tuning. Same as in action classification, we use 8 uniformly sampled frames as inputs. We also follow the corresponding model recipes in using re-ranking, namely, we rerank 128 videos with highest similarities based on dual encoder output by a joint encoder for Unmasked Teacher (UMT)~\cite{li2023unmasked} and VideoMamba~\cite{li2024videomamba} models.

\section{Qualitative Results}
\label{sec:qualitative}

\subsection{UTD-descriptions}

\begin{figure*}[]
\begin{center}
\includegraphics[width=0.8\linewidth]{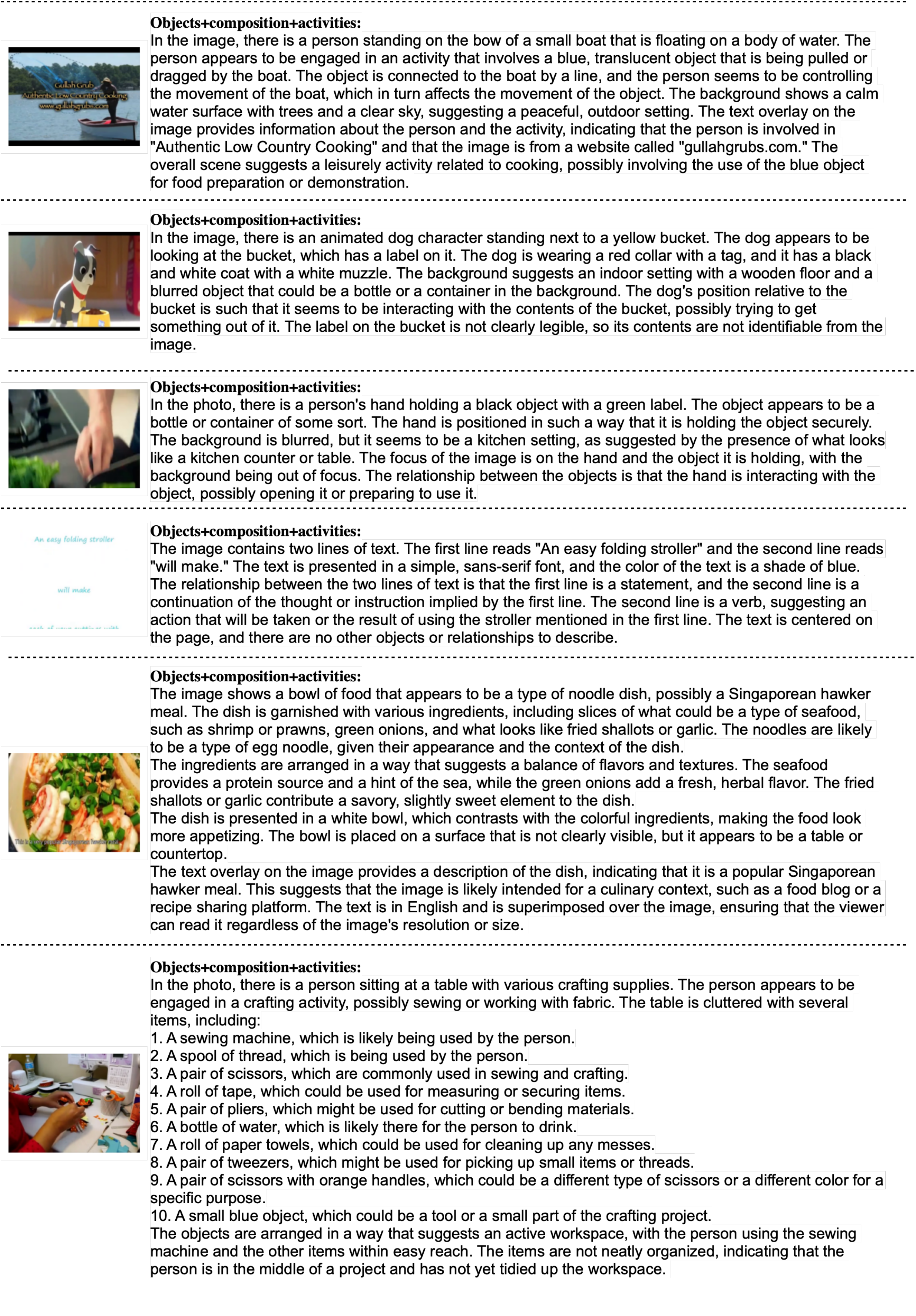}
\end{center}
\vspace{-2mm}
\caption{ \small{
{Qualitative examples of objects+composition+activities textual descriptions for random videos in MSRVTT dataset.}}
\label{fig:qualitative_1}
}
\vspace{-6mm}
\end{figure*}

\begin{figure*}[]
\begin{center}

\begin{subfigure}[t]{0.75\linewidth}
\includegraphics[width=1\linewidth]{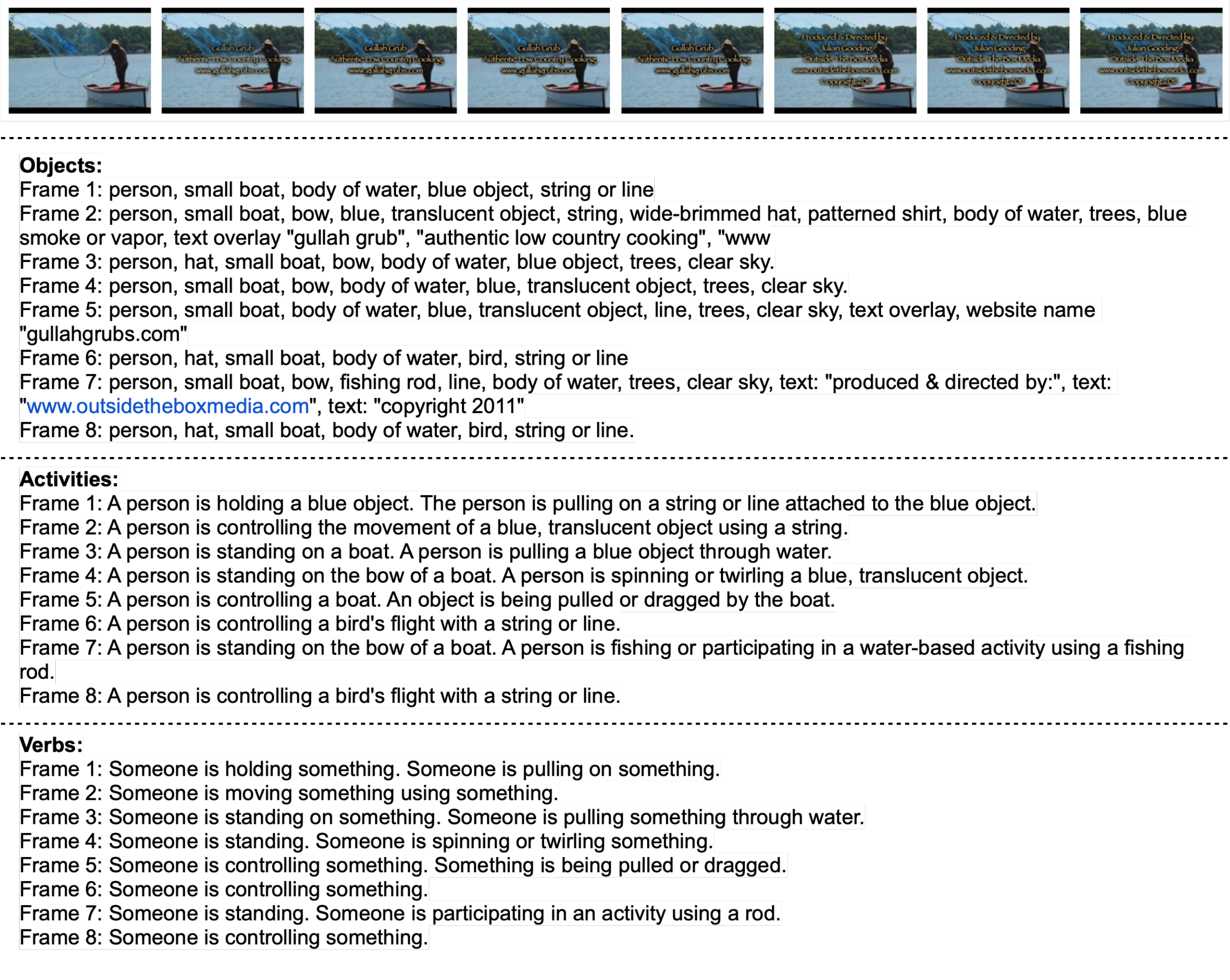}
\end{subfigure}%

\begin{subfigure}[t]{0.75\linewidth}
\includegraphics[width=1\linewidth]{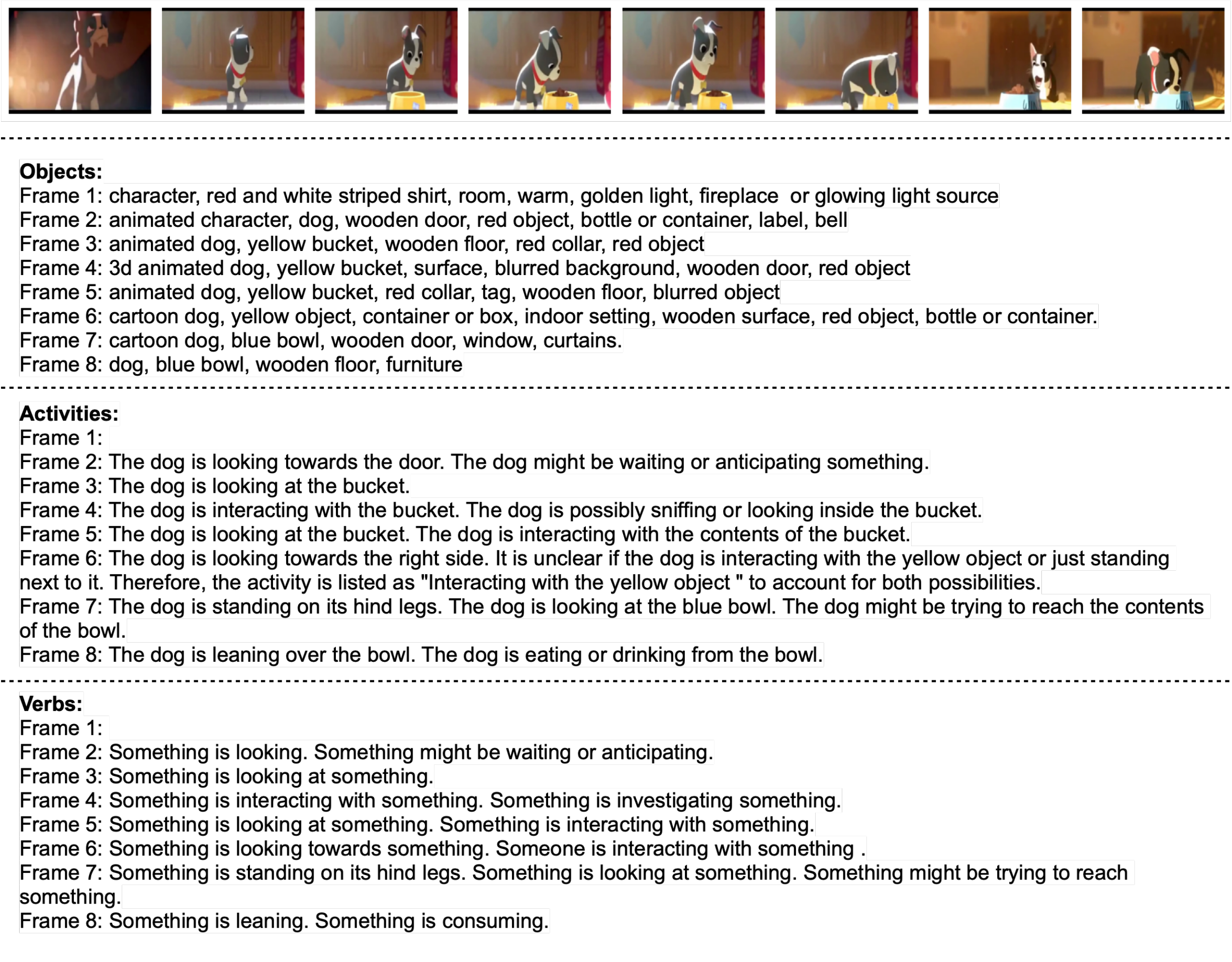}
\end{subfigure}%

\end{center}
\caption{ \small{
\label{fig:qualitative_2} {Qualitative examples of objects, activities, and verbs textual descriptions for random videos in MSRVTT dataset.}
}}
\end{figure*}

\begin{figure*}[]
\begin{center}

\begin{subfigure}[t]{0.75\linewidth}
\includegraphics[width=1\linewidth]{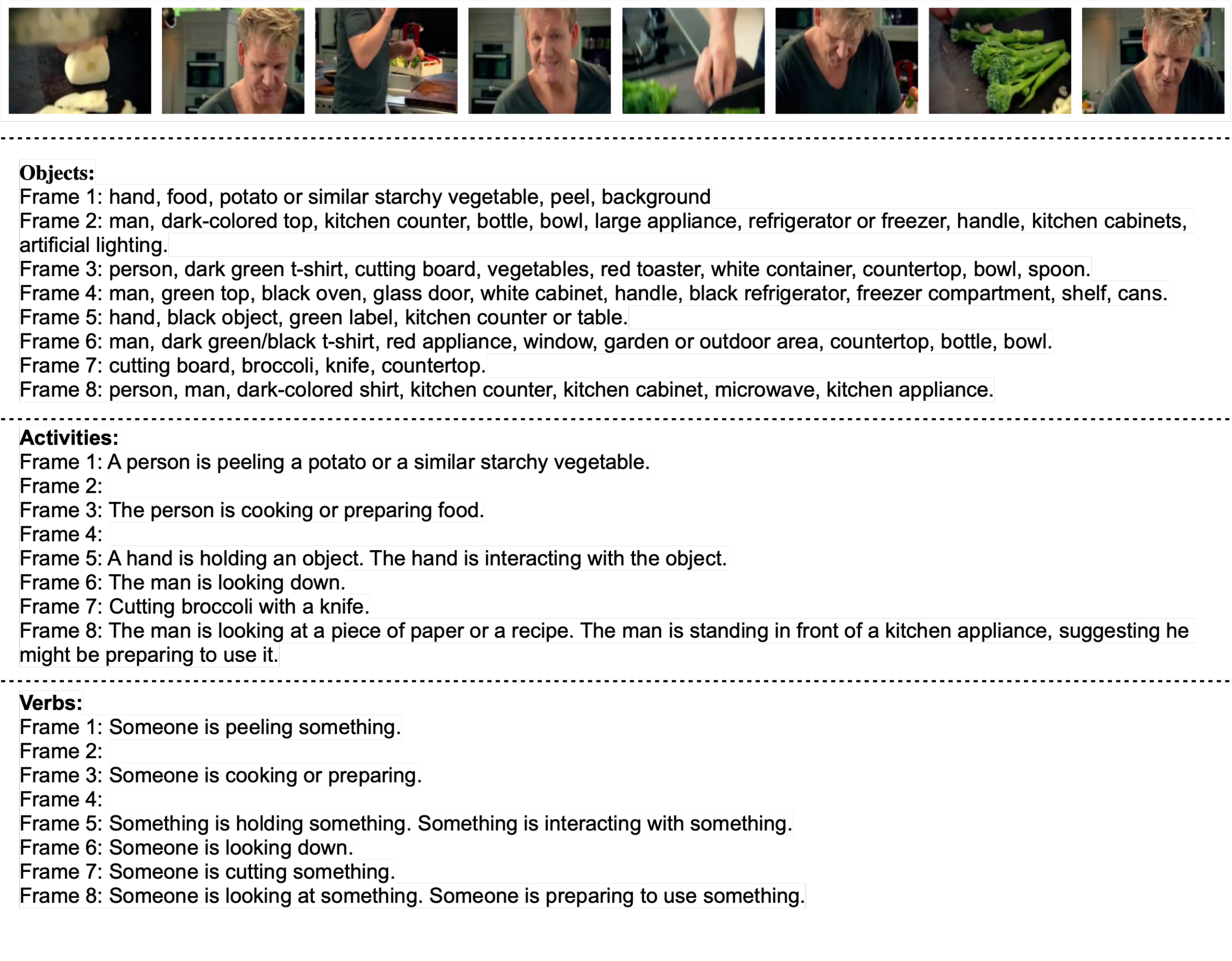}
\end{subfigure}%

\begin{subfigure}[t]{0.75\linewidth}
\includegraphics[width=1\linewidth]{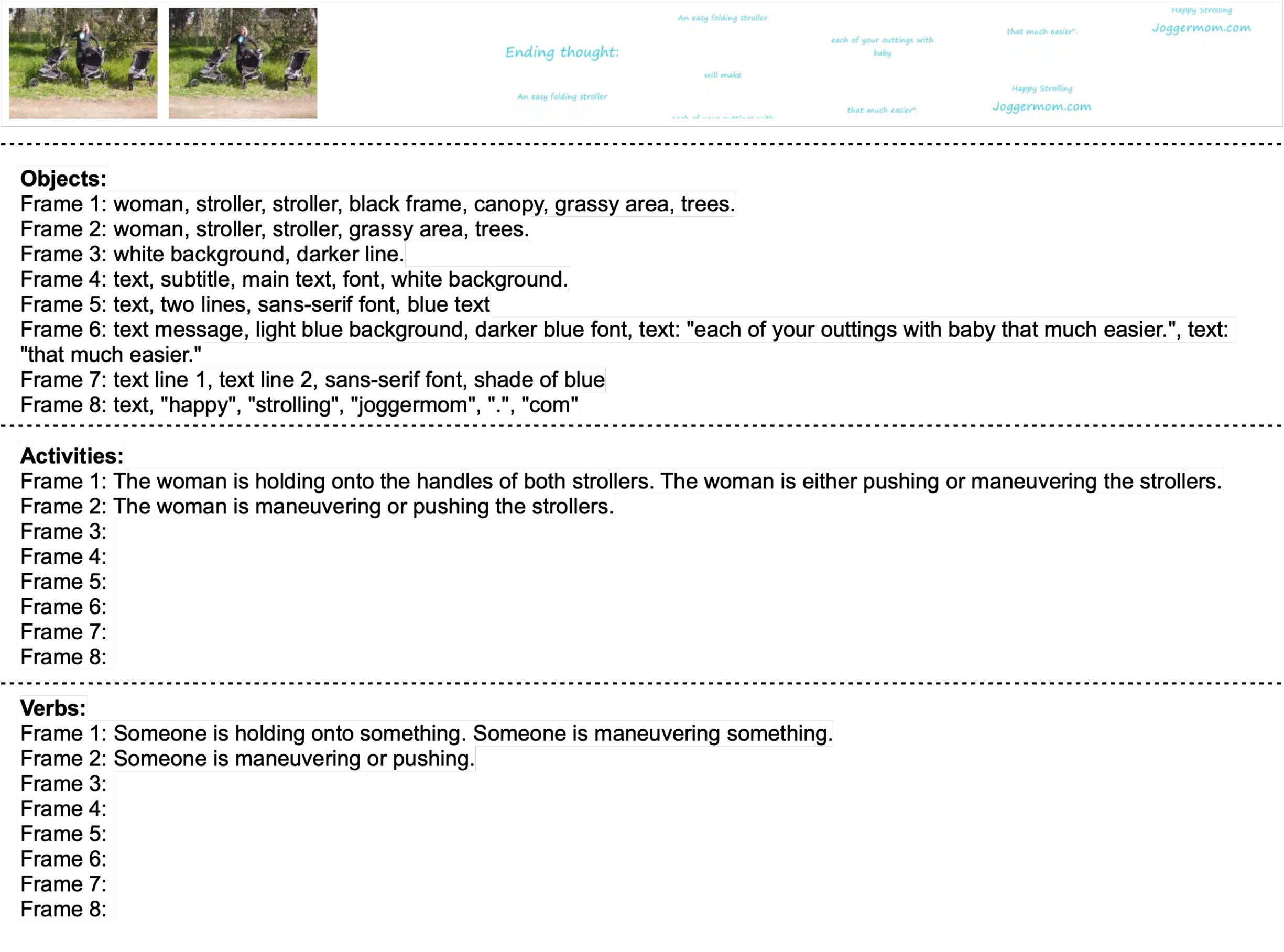}
\end{subfigure}%

\end{center}
\caption{ \small{
\label{fig:qualitative_3} {Qualitative examples of objects, activities, and verbs textual descriptions for random videos in MSRVTT dataset.}
}}
\end{figure*}

\begin{figure*}[]
\begin{center}

\begin{subfigure}[t]{0.75\linewidth}
\includegraphics[width=1\linewidth]{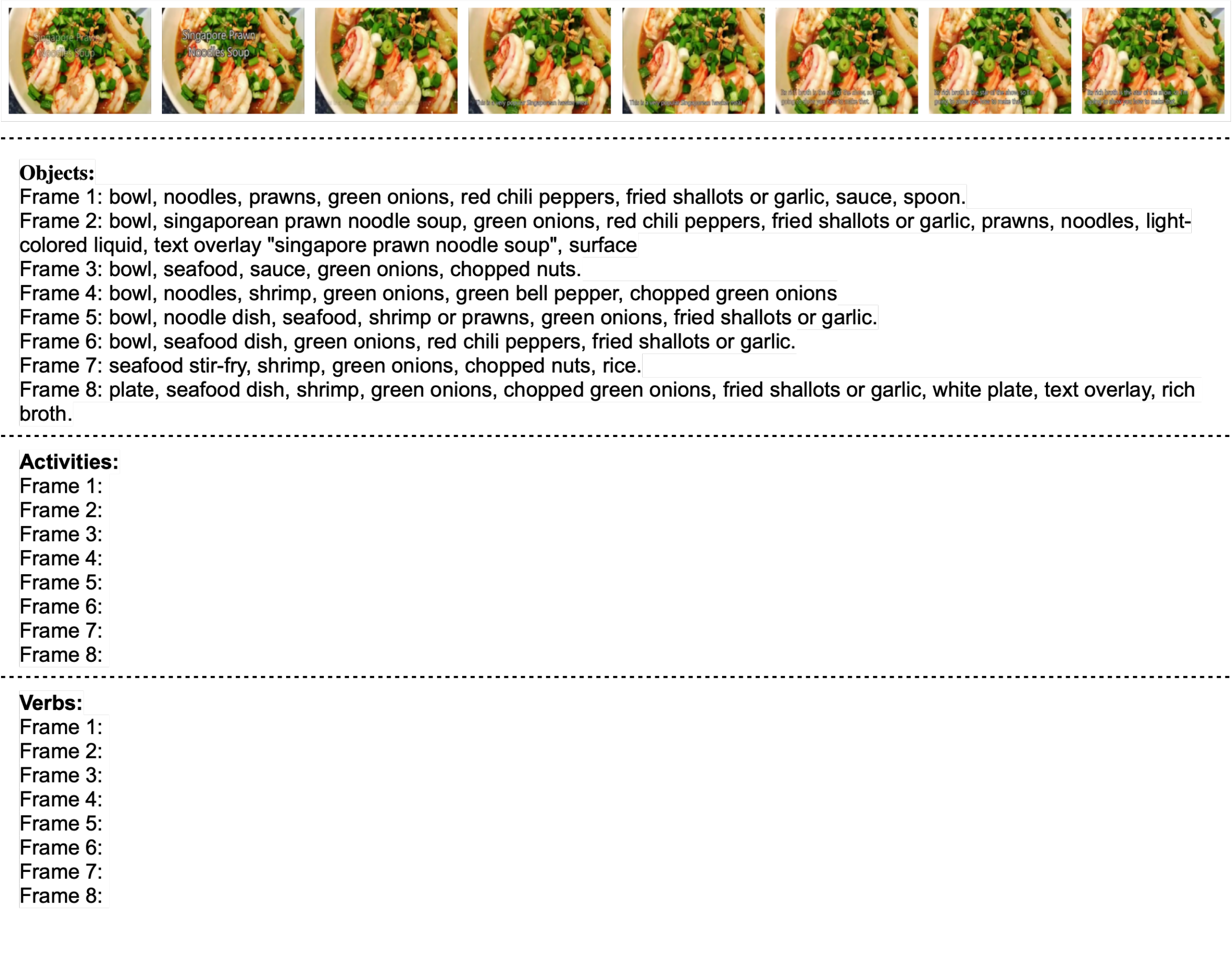}
\end{subfigure}%
\vspace{-6mm}

\begin{subfigure}[t]{0.75\linewidth}
\includegraphics[width=1\linewidth]{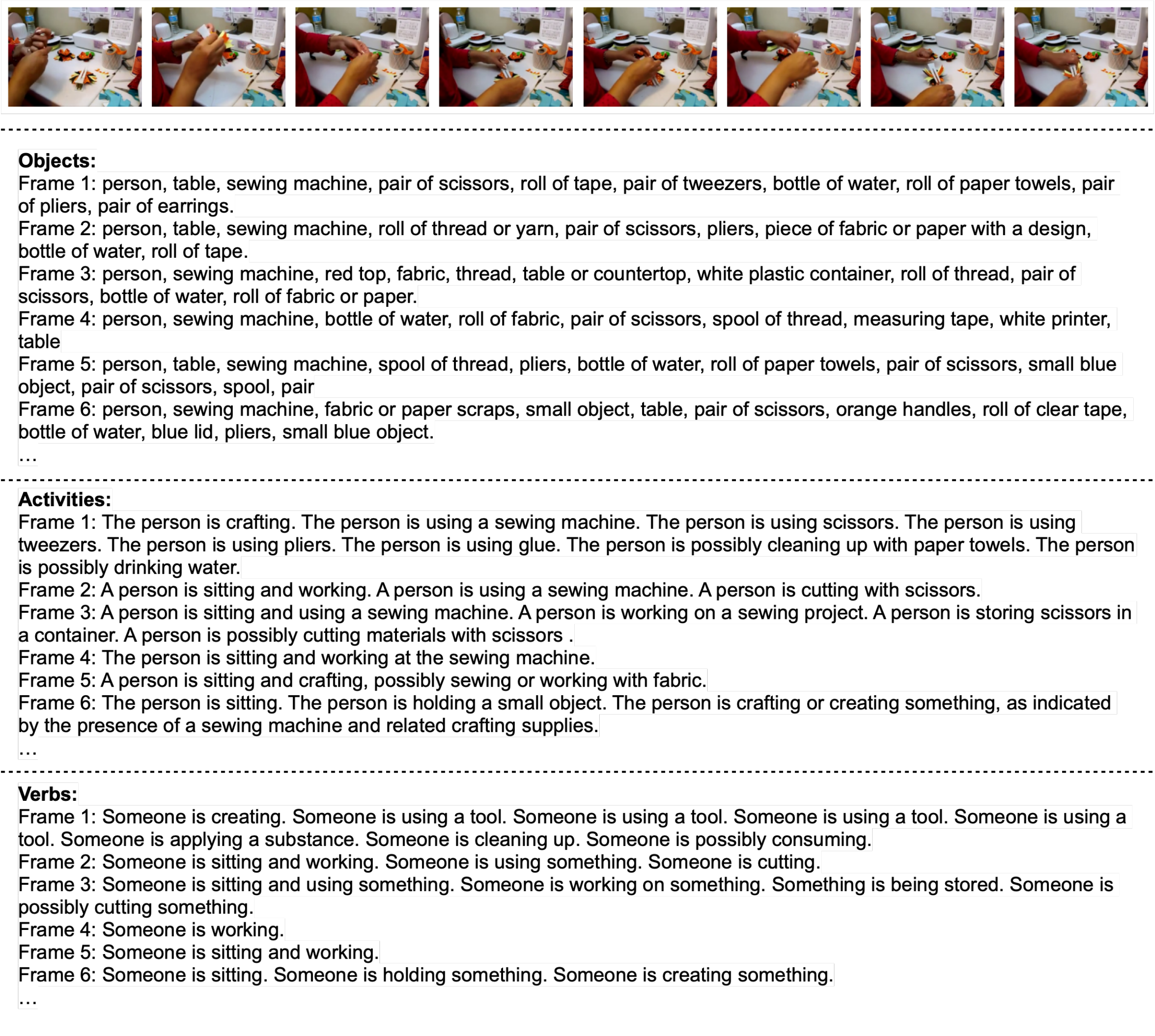}
\end{subfigure}%
\vspace{-6mm}

\end{center}
\caption{ \small{
\label{fig:qualitative_4} {Qualitative examples of objects, activities, and verbs textual descriptions for random videos in MSRVTT dataset.}
}}
\end{figure*}

In \cref{fig:qualitative_1,fig:qualitative_2,fig:qualitative_3,fig:qualitative_4}, we present qualitative results of textual descriptions of our \textit{UTD-descriptions} dataset for different concept categories using random videos from the MSRVTT dataset. We observed that the VLM provides detailed descriptions of objects+composition+activities (\cref{fig:qualitative_1}). Furthermore, the LLM successfully parses these descriptions into objects, activities, and verbs (\cref{fig:qualitative_2,fig:qualitative_3,fig:qualitative_4}).

\subsection{UTD-splits}

\begin{figure*}[]
\begin{center}

\begin{subfigure}[t]{1\linewidth}
\begin{subfigure}[t]{0.245\linewidth}
\includegraphics[width=1\linewidth]{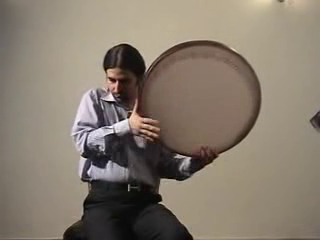}
\vspace{-4mm}\caption*{Playing Daf}
\end{subfigure}%
\hfill
\begin{subfigure}[t]{0.245\linewidth}
\includegraphics[width=1\linewidth]{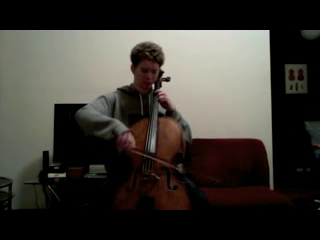}
\vspace{-4mm}\caption*{Playing Cello}
\end{subfigure}%
\hfill
\begin{subfigure}[t]{0.245\linewidth}
\includegraphics[width=1\linewidth]{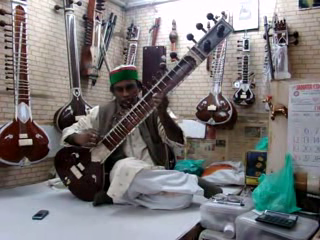}
\vspace{-4mm}\caption*{Playing Sitar}
\end{subfigure}%
\hfill
\begin{subfigure}[t]{0.245\linewidth}
\includegraphics[width=1\linewidth]{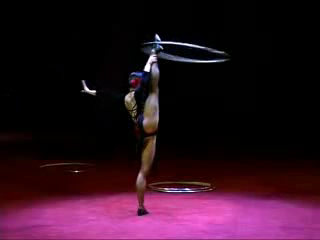}
\vspace{-4mm}\caption*{Hula Hoop}
\end{subfigure}%

\begin{subfigure}[t]{0.245\linewidth}
\includegraphics[width=1\linewidth]{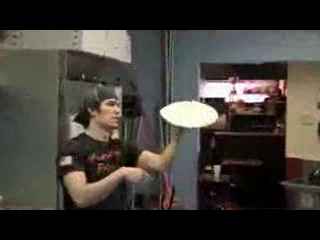}
\vspace{-4mm}\caption*{Pizza Tossing}
\end{subfigure}%
\hfill
\begin{subfigure}[t]{0.245\linewidth}
\includegraphics[width=1\linewidth]{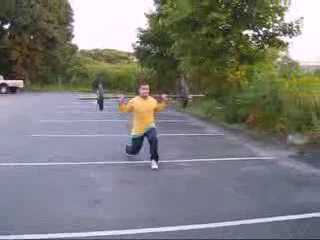}
\vspace{-4mm}\caption*{Lunges}
\end{subfigure}%
\hfill
\begin{subfigure}[t]{0.245\linewidth}
\includegraphics[width=1\linewidth]{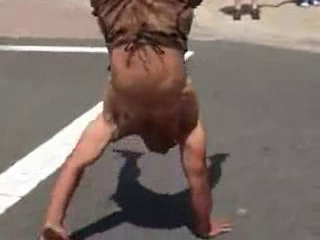}
\vspace{-4mm}\caption*{Handstand Walking}
\end{subfigure}%
\hfill
\begin{subfigure}[t]{0.245\linewidth}
\includegraphics[width=1\linewidth]{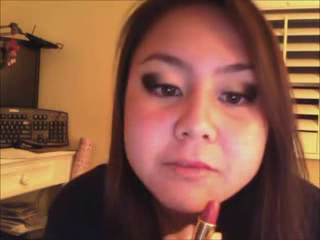}
\vspace{-4mm}\caption*{Apply Lipstick}
\end{subfigure}%

\caption{Examples from the UCF101 test split (that are not included in the UCF101-UTD split)}
\end{subfigure}%
\vspace{5mm}

\begin{subfigure}[t]{1\linewidth}
\begin{subfigure}[t]{0.245\linewidth}
\includegraphics[width=1\linewidth]{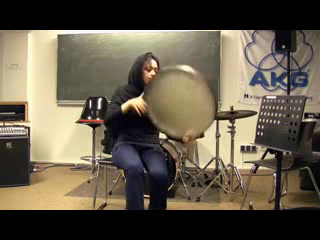}
\vspace{-4mm}\caption*{Playing Daf}
\end{subfigure}%
\hfill
\begin{subfigure}[t]{0.245\linewidth}
\includegraphics[width=1\linewidth]{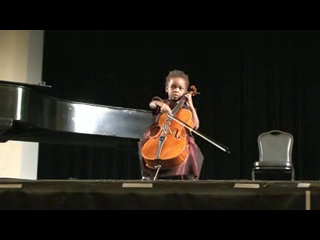}
\vspace{-4mm}\caption*{Playing Cello}
\end{subfigure}%
\hfill
\begin{subfigure}[t]{0.245\linewidth}
\includegraphics[width=1\linewidth]{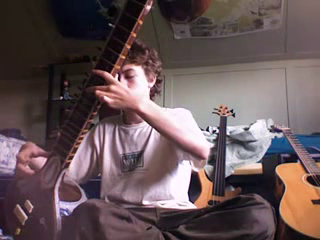}
\vspace{-4mm}\caption*{Playing Sitar}
\end{subfigure}%
\hfill
\begin{subfigure}[t]{0.245\linewidth}
\includegraphics[width=1\linewidth]{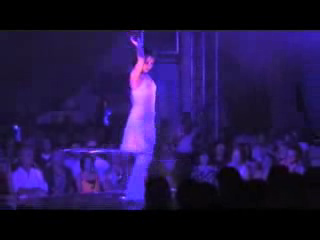}
\vspace{-4mm}\caption*{Hula Hoop}
\end{subfigure}%

\begin{subfigure}[t]{0.245\linewidth}
\includegraphics[width=1\linewidth]{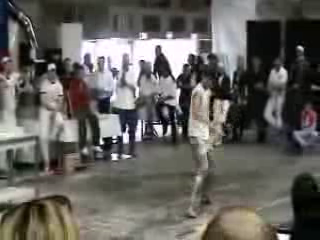}
\vspace{-4mm}\caption*{Pizza Tossing}
\end{subfigure}%
\hfill
\begin{subfigure}[t]{0.245\linewidth}
\includegraphics[width=1\linewidth]{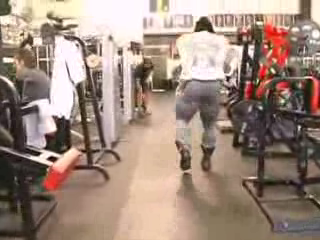}
\vspace{-4mm}\caption*{Lunges}
\end{subfigure}%
\hfill
\begin{subfigure}[t]{0.245\linewidth}
\includegraphics[width=1\linewidth]{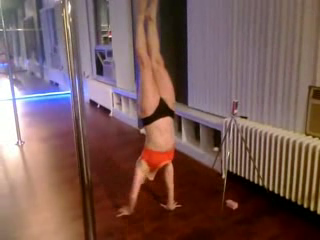}
\vspace{-4mm}\caption*{Handstand Walking}
\end{subfigure}%
\hfill
\begin{subfigure}[t]{0.245\linewidth}
\includegraphics[width=1\linewidth]{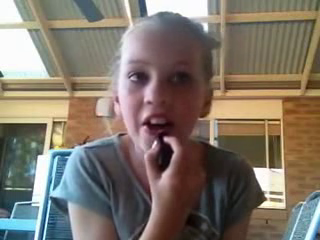}
\vspace{-4mm}\caption*{Apply Lipstick}
\end{subfigure}%

\caption{Examples from our UCF101-UTD-split}
\end{subfigure}%
\vspace{2mm}

\end{center}
\caption{ \small{Video examples with their class label from full UCF101 test set and our object-debiased UTD-split. We observe that samples from our object-debiased UTD-split require a level of video understanding beyond simple object recognition.  For instance, in the case of playing musical instruments, e.g., Playing Daf or Playing Cello, the videos often include other musical instruments in the background, e.g., a piano or drums in the case of Playing Daf, alongside the primary instrument. Similarly, in the Pizza Tossing class, the pizza in the UTD-split example is hardly visible, and a video requires analysis beyond this single frame for correct class prediction.  \label{fig:UTD_splits}
}}
\end{figure*}

In~\cref{fig:UTD_splits}, we present qualitative examples from the full test set and our object-debiased UTD-split on the UCF101 dataset.
We observe that samples from our object-debiased UTD-split demand a deeper level of video understanding beyond simple object recognition. 
For instance, in examples of classes involving playing musical instruments, such as ``Playing Daf'', ``Playing Cello'', or ``Playing Sitar'', the videos often include additional instruments in the background, such as a piano and drums in the case of ``Playing Daf'' example, alongside the primary instrument. Similarly, in the ``Pizza Tossing'' example, the pizza in the UTD-split example is barely visible, and a video requires analysis beyond this single frame for correct class prediction.

\subsection{User Study}

\begin{figure*}[]
\begin{center}

\begin{subfigure}[t]{0.8\linewidth}
\begin{subfigure}[t]{0.31\linewidth}
\includegraphics[width=1\linewidth]{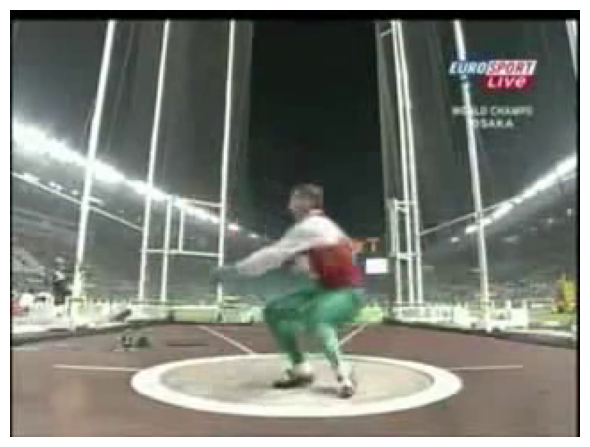}
\vspace{-6mm}\caption*{sleeveless top}
\end{subfigure}%
\hfill
\begin{subfigure}[t]{0.31\linewidth}
\includegraphics[width=1\linewidth]{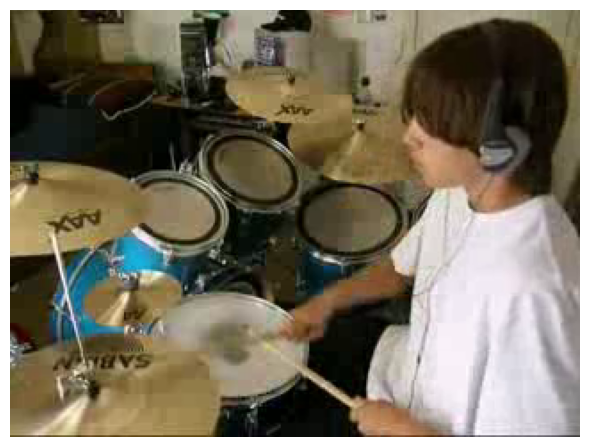}
\vspace{-6mm}\caption*{snare drum}
\end{subfigure}%
\hfill
\begin{subfigure}[t]{0.31\linewidth}
\includegraphics[width=1\linewidth]{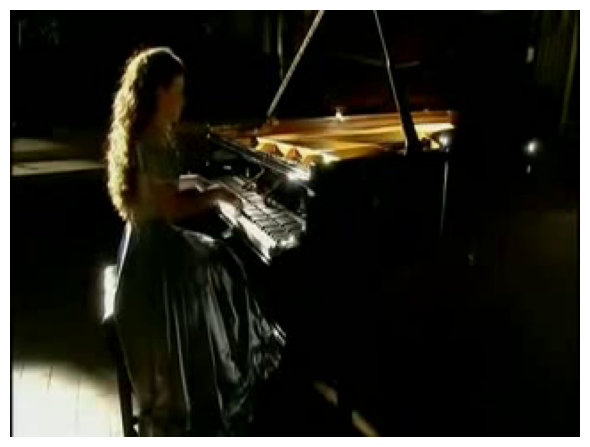}
\vspace{-6mm}\caption*{upright piano}
\end{subfigure}%
\vspace{-1mm}
\caption{Attribute error}
\end{subfigure}%
\vspace{2mm}

\begin{subfigure}[t]{0.8\linewidth}
\begin{subfigure}[t]{0.31\linewidth}
\includegraphics[width=1\linewidth]{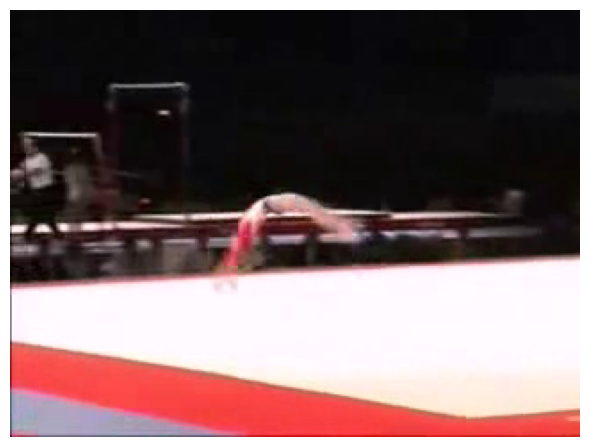}
\vspace{-6mm}
\caption*{platform}
\end{subfigure}%
\hfill
\begin{subfigure}[t]{0.31\linewidth}
\includegraphics[width=1\linewidth]{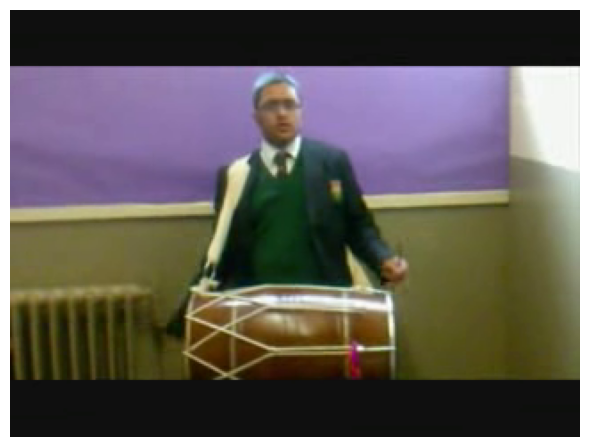}
\vspace{-6mm}
\caption*{whiteboard or chalkboard}
\end{subfigure}%
\hfill
\begin{subfigure}[t]{0.31\linewidth}
\includegraphics[width=1\linewidth]{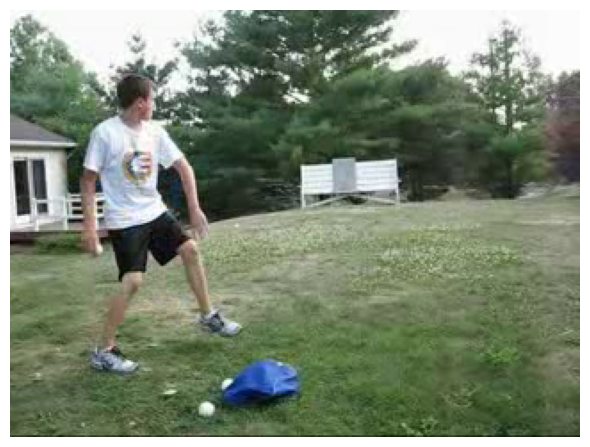}
\vspace{-6mm}
\caption*{blue sports ball}
\end{subfigure}%
\vspace{-1mm}
\caption{Misclassification}
\end{subfigure}%
\vspace{2mm}

\begin{subfigure}[t]{0.8\linewidth}
\begin{subfigure}[t]{0.31\linewidth}
\includegraphics[width=1\linewidth]{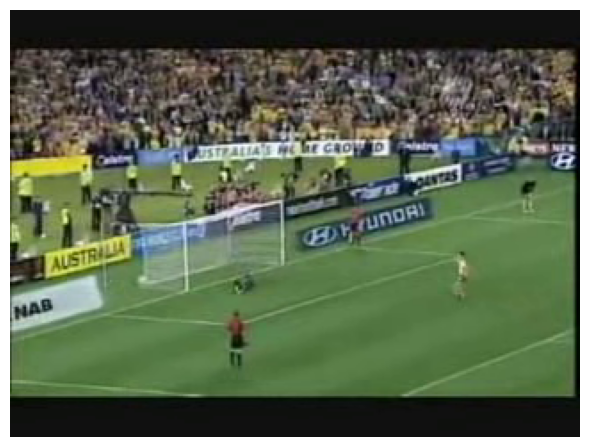}
\vspace{-6mm}
\caption*{rows of seats}
\end{subfigure}%
\hfill
\begin{subfigure}[t]{0.31\linewidth}
\includegraphics[width=1\linewidth]{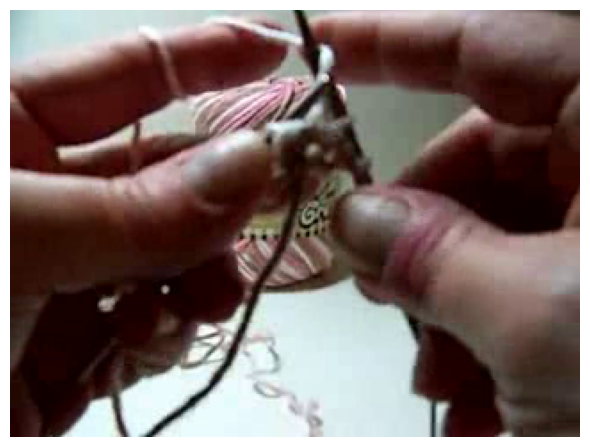}
\vspace{-6mm}
\caption*{metallic base}
\end{subfigure}%
\hfill
\begin{subfigure}[t]{0.31\linewidth}
\includegraphics[width=1\linewidth]{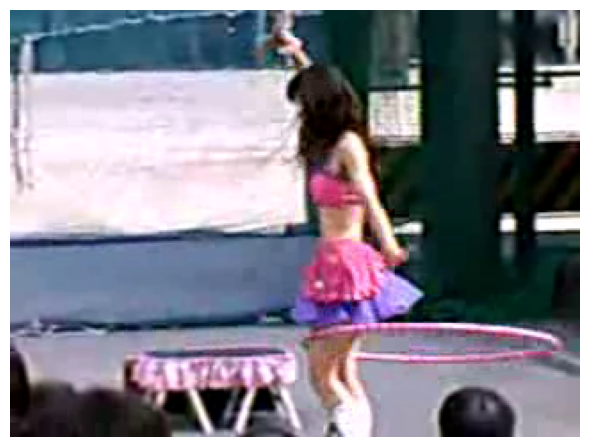}
\vspace{-6mm}
\caption*{fence}
\end{subfigure}%
\vspace{-1mm}
\caption{Hallucination}
\end{subfigure}%
\vspace{2mm}

\begin{subfigure}[t]{0.8\linewidth}
\begin{subfigure}[t]{0.31\linewidth}
\includegraphics[width=1\linewidth]{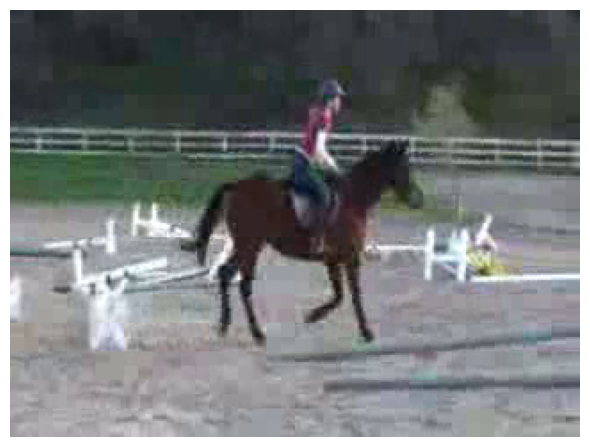}
\vspace{-6mm}
\caption*{white rails}
\end{subfigure}%
\hfill
\begin{subfigure}[t]{0.31\linewidth}
\includegraphics[width=1\linewidth]{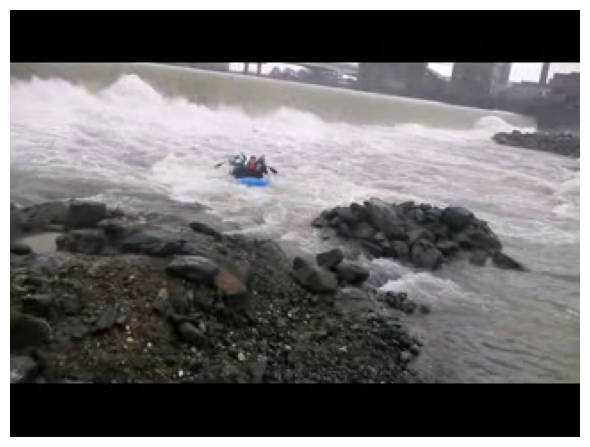}
\vspace{-6mm}
\caption*{white-capped waves}
\end{subfigure}%
\hfill
\begin{subfigure}[t]{0.31\linewidth}
\includegraphics[width=1\linewidth]{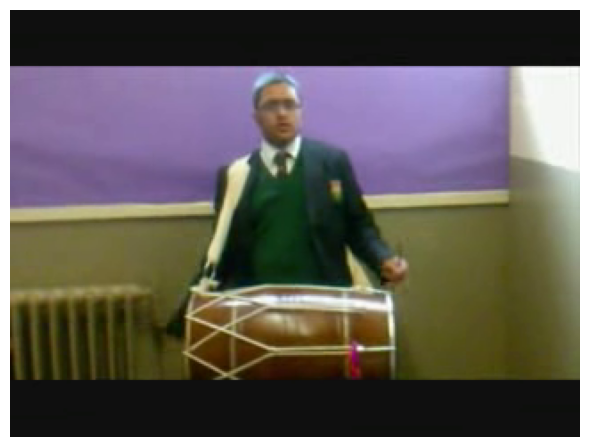}
\vspace{-6mm}
\caption*{red strap}
\end{subfigure}%
\vspace{-1mm}
\caption{Human annotation mistake – the object is visible}
\end{subfigure}%
\vspace{2mm}

\begin{subfigure}[t]{0.8\linewidth}
\begin{subfigure}[t]{0.31\linewidth}
\includegraphics[width=1\linewidth]{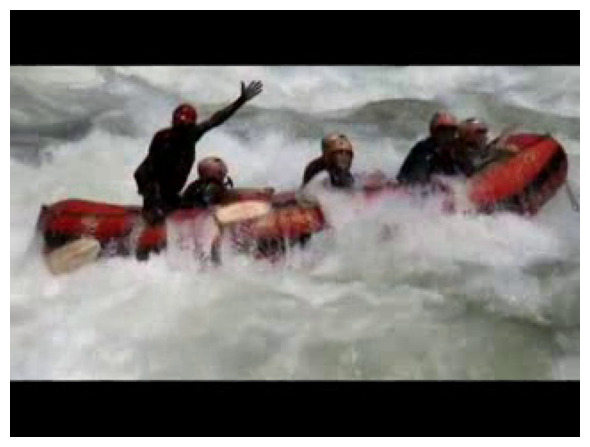}
\vspace{-6mm}
\caption*{body of water}
\end{subfigure}%
\hfill
\begin{subfigure}[t]{0.31\linewidth}
\includegraphics[width=1\linewidth]{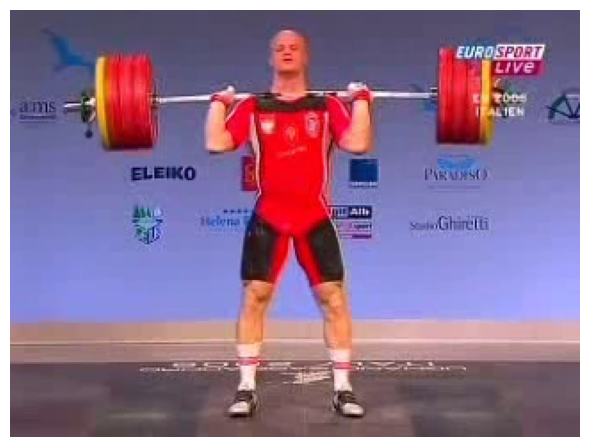}
\vspace{-6mm}
\caption*{air}
\end{subfigure}%
\hfill
\begin{subfigure}[t]{0.31\linewidth}
\includegraphics[width=1\linewidth]{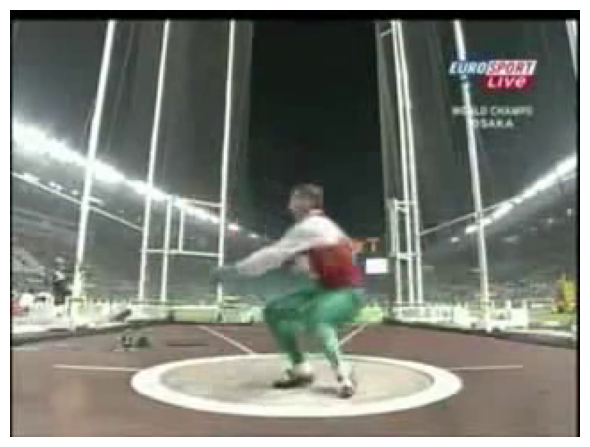}
\vspace{-6mm}
\caption*{large}
\end{subfigure}%
\vspace{-1mm}
\caption{Other}
\end{subfigure}%
\vspace{2mm}

\end{center}
\caption{ \small{Examples of manual classification of objects predicted by the VLM for the image, but not selected as visible in the image during the user study. We consider five categories: attribute error, misclassification, hallucination, human annotation mistake (the object is visible), and other. \label{fig:user_study}
}}
\end{figure*}

As stated in the main paper, our user study shows
that 87.6\% (667 out of 761) of the VLM-recognized objects are identified as visible, with only 94 objects not selected as visible. To better
understand the VLM’s errors, we manually classifiy these 94 objects into five categories: 1) attribute error (e.g., the object is ``right hand'' instead of ``left hand''), 2) misclassification (the object is present but incorrectly identified as a different object, e.g., a ``snowboard'' instead of ``snow slide''), 3) hallucination, 4) human annotation mistake – the object is visible, and 5) other. In~\cref{fig:user_study} we provide examples from these five categories.

\section{License Information}
\label{sec:license}

Our \textit{UTD-descriptions} are generated by LLaVA-1.6-7B-mistral model~\cite{liu2023llava,liu2024llavanext} as well as Mistral-7B-Instruct-v0.2~\cite{jiang2023mistral}. The models were used according to their licenses. LLaVA-1.6-7B-mistral model complies with the base LLM's model license\footnote{\url{https://github.com/haotian-liu/LLaVA/blob/main/docs/MODEL_ZOO.md}}, which is Mistral-7B. Mistral-7B model follows Apache 2.0 license\footnote{\url{https://mistral.ai/news/announcing-mistral-7b/}}. 
We release our \textit{UTD dataset}, comprising \textit{UTD-descriptions} and \textit{UTD-splits}, under the CC BY 4.0 license. However, specific components of the underlying dataset may be governed by stricter licensing conditions from the corresponding video datasets.

\section{Limitations and Broader Impact}
\label{sec:limitations}

\textbf{Limitations.} Due to the high human annotation cost of detailed frame text descriptions for videos, our method of evaluating and debiasing datasets relies on the textual description of video frames generated by VLMs. Therefore, these textual descriptions may potentially misrepresent the video due to the potential limitation of VLM models. Namely, such textual description might be prone to hallucinations, namely describing things that are not present in the frame, having social biases, such as guessing a person's occupation based on how the person looks or implying some information that is not visible.  Then, we further extract different concept categories from these textual descriptions using LLM. These steps might be prone to leaking information about other concepts, such as adding activity information to the objects list. Therefore, the quality of our textual descriptions with respect to different concepts also limited the filtering abilities of LLMs. In our user study evaluating the quality of generated descriptions of objects, we found the hallucination rate to be approximately 6\%. Our biased discovering method is also limited by the performance of pretrained text embedding models, and using stronger text embedding models might lead to discovering even stronger biases in the datasets, such as more videos being classified correctly based on only objects using common sense. 

\noindent\textbf{Potential Negative Societal Impact.} In our work, we perform bias analysis of the existing video classification and text-to-video retrieval datasets, 12 datasets in total, as well as provide debiased evaluation splits -- \textit{UTD-splits}. Since our annotations are based on existing datasets, our data distribution reflects the social biases inherent in those sources.
We also provide \textit{UTD-descriptions} for these 12 datasets -- textual descriptions of objects, activities, verbs, and objects+composition+activities categories. As discussed in the limitations, these descriptions might be prone to social biases potentially present in the current VLM models. Therefore, such descriptions might potentially propagate these social biases.
To create the dataset, we utilize VLM and LLM models, which contribute to increased energy consumption and carbon emissions as a negative externality.

\clearpage
\onecolumn
\section{$UTD$ Dataset}
\label{sec:datasheet}

We release the $UTD$ dataset, which consists of two parts: 
\begin{enumerate}
    \item \textbf{$UTD$-descriptions.} This includes frame annotations for four conceptual categories visible in video frames: \textit{objects, activities, verbs}, and \textit{objects+composition+activities}. $UTD$-descriptions are provided for 8 uniformly sampled frames from the training and test/val sets of 12 action recognition and text-to-video retrieval datasets. 
The annotations \textit{for objects+composition+activities} are generated using the LLaVA-1.6-7B-mistral VLM prompted to describe visible object relationships in a frame. 
From these descriptions, \textit{objects, activities}, and \textit{verbs (activities without associated objects)}
are derived using the Mistral-7B-Instruct-v0.2 model.

    \item \textbf{$UTD$-splits.} This includes object-debiased test/val splits, 
which are subsets of the original test/val splits with object-biased items removed. These debiased splits are provided for the 12 considered activity recognition and text-to-video retrieval datasets. For the 6 activity recognition datasets,
we additionally provide debiased-balanced splits, where the most object-biased samples are removed
while preserving the original class distribution to ensure fair evaluation across categories.
\end{enumerate}

The download instructions, documentation, and usage guidance may be found 
on our project webpage: \url{https://utd-project.github.io/}
Below, we provide the datasheet for our $UTD$ dataset, license information, and statement of responsibility.

\subsection*{$UTD$ Dataset Datasheet}

\dssectionheader{Motivation}

\dsquestionex{For what purpose was the dataset created?}{Was there a specific task in mind? Was there a specific gap that needed to be filled? Please provide a description.}

The $UTD$ Dataset is a benchmark designed to assess the performance of video backbones. It consists of debiased evaluation subsets, specifically video IDs, of 12 popular action classification and text-to-video retrieval datasets ($UTD$-splits), namely UCF101~\cite{SoomroUCF101}, SomethingSomethingv2~\cite{Goyal_2017_ICCV}, Kinetics-400~\cite{Kay2017Kinetics}, -600~\cite{Carreira2018Kinetics600}, and -700~\cite{Carreira2019Kinetics700}, and Moments In Time~\cite{monfort2019moments}, MSRVTT~\cite{xu2016msr}, DiDeMo~\cite{anne2017localizing}, ActivityNet~\cite{Heilbron_2015_CVPR}, LSMDC~\cite{Rohrbach_2015_CVPR}, YouCook2~\cite{Das_2013_YouCook}, and Spoken Moments In Time~\cite{Monfort_2021_CVPR}.  The goal is to evaluate the robustness of video models to object representation bias and to provide a challenging benchmark for evaluating video models with reduced object bias in the evaluation set.
While previous work has focused on assessing and mitigating various representation biases in video benchmarks, debiased solutions have rarely been adopted for benchmarking. This is due to several reasons, such as the additional training and/or testing overhead required or the necessity to address out-of-domain problems. Our work introduces a novel method for evaluating and debiasing existing datasets via their textual descriptions. This approach allows us to identify and remove samples with object representation bias from the evaluation sets.
Additionally, the dataset includes $UTD$-descriptions, which are textual descriptions of four conceptual categories visible in video frames: objects, activities, verbs, and objects+composition+activities. These annotations cover the 12 corresponding datasets and aim further to advance the measurement of representation biases in the field.

\dsquestion{Who created this dataset (e.g., which team, research group) and on behalf of which entity (e.g., company, institution, organization)?}

The dataset was created by a research group affiliated with the Goethe University Frankfurt, Tuebingen AI Center/University of Tuebingen, University of Oxford, MPI for Informatics, and MIT-IBM Watson AI Lab. 

\dsquestionex{Who funded the creation of the dataset?} {If there is an associated grant, please provide the name of the grantor and the grant name and number.} 

Individual researchers within the research group have been funded by the German Federal Ministry of Education and Research (BMBF) project STCL - 01IS22067, the ERC Starting Grant GraViLa 101117556, and supported by travel grants from ELISE (GA no 951847).

\dssectionheader{Composition}

\dsquestionex{What do the instances that comprise the dataset
represent (e.g., documents, photos, people, countries)?} {Are there
multiple types of instances (e.g., movies, users, and ratings;
people and interactions between them; nodes and edges)? Please
provide a description.} 

Our dataset builds upon existing datasets. It contains only textual annotation or meta-annotation. $UTD$-splits contain lists of video IDs.  $UTD$-descriptions contain texts.  

\dsquestionex{How many instances are there in total (of each type, if appropriate)?}{} 

$UTD$-splits contain debiased splits for 12 different action classifications and text-to-video retrieval datasets. $UTD$-descriptions contain textual annotation for train/test videos of corresponding datasets, describing $\sim$1.9M videos in total.

\dsquestionex{Does the dataset contain all possible instances, or is it a sample (not necessarily random) of instances from a larger set?}{If the dataset is a sample, then what is the larger set? Is the sample representative of the larger set (e.g., geographic coverage)? If so, please describe how this representativeness was validated/verified. If it is not representative of the larger set, please describe why not (e.g., to cover a more diverse range of instances because instances were withheld or unavailable).} 

Our dataset builds upon existing datasets. It provides video IDs of the samples that are more representative of the corresponding tasks and samples that cannot be easily solved with simple techniques.
The main purpose of this dataset is to filter out these non-representative, easy samples from existing datasets.

\dsquestionex{What data does each instance consist of?} {``Raw'' data
  (e.g., unprocessed text or images) or features? In either case,
  please provide a description.} 
  
  An instance of $UTD$-splits is the name of a video dataset along with a list of video IDs in the object-debiased subsets and, for six activity recognition datasets, a list of video IDs in the object-debiased-balanced subsets.
 
  An instance of $UTD$-descriptions is a video ID, for which we provide annotations for four conceptual categories: objects, activities, verbs, and objects+composition+activities. These annotations are provided for 8 uniformly sampled frames for video corresponding to the video ID.
  
\dsquestionex{Is there a label or target associated with each
    instance?} {If so, please provide a description.} 

    Not applicable.  
    We provide annotations for existing datasets that already have established labels.
    These type of labels varies across datasets and tasks.
    
\dsquestionex{Is any information missing from individual instances?}
  {If so, please provide a description, explaining why this information
  is missing (e.g., because it was unavailable). This does not include
  intentionally removed information, but might include, e.g., redacted
  text.} 
  
  Instances are complete.

\dsquestionex{Are relationships between individual instances made
    explicit (e.g., users' movie ratings, social network links)?} {If
  so, please describe how these relationships are made explicit.} 
  
  Not applicable. 

\dsquestionex{Are there recommended data splits (e.g., training,
    development/validation, testing)?} {If so, please provide a
  description of these splits, explaining the rationale behind them.}
  
  Yes. We provide annotations for existing datasets with well-established splits.

\dsquestionex{Are there any errors, sources of noise, or redundancies
    in the dataset?} {If so, please provide a description.}
    
$UTD$-descriptions are generated by VLM and LLM models, which might introduce hallucinations, social biases, or imply information that is not actually visible in the frames. Since $UTD$-splits are derived using $UTD$-descriptions, it too may be susceptible to these errors.

\dsquestionex{Is the dataset self-contained, or does it link to or
    otherwise rely on external resources (e.g., websites, tweets,
    other datasets)?} {If it links to or relies on external resources,
    a) are there guarantees that they will exist, and remain constant,
    over time; b) are there official archival versions of the complete
    dataset (i.e., including the external resources as they existed at
    the time the dataset was created); c) are there any restrictions
    (e.g., licenses, fees) associated with any of the external
    resources that might apply to a \edit{dataset consumer}? Please provide
    descriptions of all external resources and any restrictions
    associated with them, as well as links or other access points, as
    appropriate.}

    The dataset is publicly available on our project webpage: \url{https://utd-project.github.io/}. This webpage includes links to data files hosted on Google Drive, and long-term public accessibility and maintenance will be ensured. The dataset will be released under the CC-4.0 license. However, certain parts of the upstream datasets may be subject to stricter licensing conditions from the corresponding video datasets.

\dsquestionex{Does the dataset contain data that might be considered
    confidential (e.g., data that is protected by legal privilege or
    by doctor\edit{--}patient confidentiality, data that includes the content
    of individuals' non-public communications)?} {If so, please provide
    a description.}

    No.

\dsquestionex{Does the dataset contain data that, if viewed directly,
    might be offensive, insulting, threatening, or might otherwise
    cause anxiety?} {If so, please describe why.}

    There is a small chance that the automatically generated text annotations can contain offensive language. However, with extensive manual checks, we have not encountered such a sample.

\dsquestionex{Does the dataset identify any subpopulations (e.g., by
    age, gender)?} {If so, please describe how these subpopulations are
  identified and provide a description of their respective
  distributions within the dataset.}

  Not applicable.

\dsquestionex{Is it possible to identify individuals (i.e., one or
    more natural persons), either directly or indirectly (i.e., in
    combination with other data) from the dataset?} {If so, please
    describe how.}

    Since $UTD$-splits contain only video IDs and $UTD$-descriptions provide textual descriptions of video frames, there is a very low chance that PID will be captured in the annotations.

\dsquestionex{Does the dataset contain data that might be considered
    sensitive in any way (e.g., data that reveals rac\edit{e} or ethnic
    origins, sexual orientations, religious beliefs, political
    opinions or union memberships, or locations; financial or health
    data; biometric or genetic data; forms of government
    identification, such as social security numbers; criminal
    history)?} {If so, please provide a description.}

    No.  %

\dssectionheader{Collection Process}

\dsquestionex{How was the data associated with each instance
    acquired?} {Was the data directly observable (e.g., raw text, movie
  ratings), reported by subjects (e.g., survey responses), or
  indirectly inferred/derived from other data (e.g., part-of-speech
  tags, model-based guesses for age or language)? If \edit{the} data was reported
  by subjects or indirectly inferred/derived from other data, was the
  data validated/verified? If so, please describe how.}

  $UTD$-descriptions are generated by VLM and LLM models from video frames. $UTD$-splits are derived using $UTD$-descriptions.

\dsquestionex{What mechanisms or procedures were used to collect the
    data (e.g., hardware apparatus\edit{es} or sensor\edit{s}, manual human
    curation, software program\edit{s}, software API\edit{s})?} {How were these
    mechanisms or procedures validated?} 

    Our dataset builds upon 12 existing datasets, namely UCF101~\cite{SoomroUCF101}, SomethingSomethingv2~\cite{Goyal_2017_ICCV}, Kinetics-400~\cite{Kay2017Kinetics}, -600~\cite{Carreira2018Kinetics600}, and -700~\cite{Carreira2019Kinetics700}, and Moments In Time~\cite{monfort2019moments}, MSRVTT~\cite{xu2016msr}, DiDeMo~\cite{anne2017localizing}, ActivityNet~\cite{Heilbron_2015_CVPR}, LSMDC~\cite{Rohrbach_2015_CVPR}, YouCook2~\cite{Das_2013_YouCook}, and Spoken Moments In Time~\cite{Monfort_2021_CVPR}. To compile our dataset, we first downloaded videos from these 12 datasets following their official instructions. We then generated $UTD$-descriptions using the officially released LLaVA-1.6-7B-mistral model and Mistral-7B-Instruct-v0.2 on an internal cluster. Additionally, we derived $UTD$-splits using the officially released SFR-Embedding-Mistral model. The detailed methodology is provided in the paper.

\dsquestionex{If the dataset is a sample from a larger set, what was
    the sampling strategy (e.g., deterministic, probabilistic with
    specific sampling probabilities)?}{}

    Not applicable. 

\dsquestionex{Who was involved in the data collection process (e.g.,
    students, crowdworkers, contractors) and how were they compensated
    (e.g., how much were crowdworkers paid)?}{}

    Not applicable. 
    
\dsquestionex{Over what timeframe was the data collected?} {Does this
  timeframe match the creation timeframe of the data associated with
  the instances (e.g., recent crawl of old news articles)?  If not,
  please describe the timeframe in which the data associated with the
  instances was created.}

  The dataset was created in 2024. 

\dsquestionex{Were any ethical review processes conducted (e.g., by an
    institutional review board)?} {If so, please provide a description
  of these review processes, including the outcomes, as well as a link
  or other access point to any supporting documentation.
}

No.

\dsquestionex{Did you collect the data from the individuals in
    question directly, or obtain it via third parties or other sources
    (e.g., websites)?}{}

    We did not perform any new data collection process but utilized data from 12 existing datasets, using corresponding official instructions to access the data. 

\dsquestionex{Were the individuals in question notified about the data
    collection?} {If so, please describe (or show with screenshots or
  other information) how notice was provided, and provide a link or
  other access point to, or otherwise reproduce, the exact language of
  the notification itself.}

  Our dataset is a meta-dataset and thus, by itself, does not collect any new data.
  All 12 considered datasets are publicly available and contain videos sourced from publicly accessible resources such as YouTube and other internet platforms, consisting of user uploads, however, we are not aware whether consent was obtained from the users. 

\dsquestionex{Did the individuals in question consent to the
    collection and use of their data?} {If so, please describe (or show
  with screenshots or other information) how consent was requested and
  provided, and provide a link or other access point to, or otherwise
  reproduce, the exact language to which the individuals consented.}

   Please see the previous answer.
 
\dsquestionex{If consent was obtained, were the consenting individuals
    provided with a mechanism to revoke their consent in the future or
    for certain uses?} {If so, please provide a description, as well as
  a link or other access point to the mechanism (if appropriate).}

  Not applicable.

\dsquestionex{Has an analysis of the potential impact of the dataset
    and its use on data subjects (e.g., a data protection impact
    analysis) been conducted?} {If so, please provide a description of
  this analysis, including the outcomes, as well as a link or other
  access point to any supporting documentation.}

  No.

\dssectionheader{Preprocessing/cleaning/labeling}

\dsquestionex{Was any preprocessing/cleaning/labeling of the data done
    (e.g., discretization or bucketing, tokenization, part-of-speech
    tagging, SIFT feature extraction, removal of instances, processing
    of missing values)?} {If so, please provide a description. If not,
  you may skip the remain\edit{ing} questions in this section.}

  We provided a simple post-processing of objects, activities, and verbs textual descriptions, such as removing numeration and text in brackets.  

\dsquestionex{Was the ``raw'' data saved in addition to the preprocessed/cleaned/labeled data (e.g., to support unanticipated future uses)?} {If so, please provide a link or other access point to the ``raw'' data.}

Yes, we provide a link to a raw version in the project webpage.

\dsquestionex{Is the software \edit{that was} used to preprocess/clean/label the \edit{data} available?} {If so, please provide a link or other access point.}

We used only simple Python scripts for this which we release.

\dssectionheader{Uses}

\dsquestionex{Has the dataset been used for any tasks already?} {If so, please provide a description.}

In our paper, we demonstrate the intended use of $UTD$-descriptions by deriving $UTD$-splits. We also use $UTD$-splits to benchmark various video backbones and analyze their robustness to object bias. All details can be found in the paper.

\dsquestionex{Is there a repository that links to any or all papers or systems that use the dataset?} {If so, please provide a link or other access point.}

All links are provided in the paper.

\dsquestionex{What (other) tasks could the dataset be used for?}{}

We believe that $UTD$-descriptions, which contain dense textual descriptions (for 8 uniformly sampled frames) of different concept categories—namely objects, activities, verbs, and objects+composition+activities—for 12 popular video datasets, could be widely used by the community for various tasks. Examples of other uses include deriving new datasets or models for understanding object relationships in videos or creating new challenging VQA datasets that require temporal understanding.

\dsquestionex{Is there anything about the composition of the dataset or the way it was collected and preprocessed/cleaned/labeled that might impact future uses?} {For example, is there anything that a \edit{dataset consumer} might need to know to avoid uses that could result in unfair treatment of individuals or groups (e.g., stereotyping, quality of service issues) or other \edit{risks or} harms (e.g., \edit{legal risks,} financial harms\edit{)?} If so, please provide a description. Is there anything a \edit{dataset consumer} could do to mitigate these \edit{risks or} harms?}

Our dataset provides annotations for existing datasets and is intended to be used in conjunction with those datasets. Therefore, while using videos and other data from the original datasets, users should comply with the licenses and terms of usage of these datasets, which are mostly restricted to research purposes. Additionally, since our annotations are generated using models, users should be aware of potential biases and inaccuracies and take appropriate measures to mitigate any risks or harms.

\dsquestionex{Are there tasks for which the dataset should not be used?} {If so, please provide a description.}

The dataset should be used for research only.

\dssectionheader{Distribution}

\dsquestionex{Will the dataset be distributed to third parties outside of the entity (e.g., company, institution, organization) on behalf of which the dataset was created?} {If so, please provide a description.}

No.

\dsquestionex{How will the dataset will be distributed (e.g., tarball on website, API, GitHub)?} {Does the dataset have a digital object identifier (DOI)?}

The dataset will be provided on our project webpage: \url{https://utd-project.github.io/}. This repository includes links to JSON data files hosted on Google Drive. It does not currently have a DOI.

\dsquestionex{When will the dataset be distributed?}{}

The dataset will be distributed starting in March 2025.

\dsquestionex{Will the dataset be distributed under a copyright or other intellectual property (IP) license, and/or under applicable terms of use (ToU)?} {If so, please describe this license and/or ToU, and provide a link or other access point to, or otherwise reproduce, any relevant licensing terms or ToU, as well as any fees associated with these restrictions.}

The dataset will be released under the Creative Commons Attribution 4.0 (CC BY 4.0) license. The terms of this license can be found at: \url{http://creativecommons.org/licenses/by/4.0}. However, certain parts of the underlying dataset may be subject to stricter licensing conditions from the corresponding video datasets.

\dsquestionex{Have any third parties imposed IP-based or other restrictions on the data associated with the instances?} {If so, please describe these restrictions, and provide a link or other access point to, or otherwise reproduce, any relevant licensing terms, as well as any fees associated with these restrictions.}

No. 

\dsquestionex{Do any export controls or other regulatory restrictions apply to the dataset or to individual instances?} {If so, please describe these restrictions, and provide a link or other access point to, or otherwise reproduce, any supporting documentation.}

Not that we are aware of.

\dssectionheader{Maintenance}

\dsquestionex{Who \edit{will be} supporting/hosting/maintaining the dataset?}{}

The dataset will be supported and maintained by the authors of the paper. The main contact person is Nina Shvetsova.

\dsquestionex{How can the owner/curator/manager of the dataset be contacted (e.g., email address)?}{}

The authors can be contacted via the following email addresses: \textit{ \{shvetsov,kuehne\}@uni-frankfurt.de}, \textit{arsha@robots.ox.ac.uk}, \textit{schiele@mpi-inf.mpg.de},  \textit{christian.rupprecht@cs.ox.ac.uk}.

\dsquestionex{Is there an erratum?} {If so, please provide a link or other access point.}

Errata will be posted on the project's webpage.

\dsquestionex{Will the dataset be updated (e.g., to correct labeling
    errors, add new instances, delete instances)?} {If so, please
  describe how often, by whom, and how updates will be communicated to
  \edit{dataset consumers} (e.g., mailing list, GitHub)?}

Updates will be communicated through the project's webpage and will be versioned. We will strive to correct errors promptly and may add or delete instances as necessary.

\dsquestionex{If the dataset relates to people, are there applicable
    limits on the retention of the data associated with the instances
    (e.g., were \edit{the} individuals in question told that their data would
    \edit{be} retained for a fixed period of time and then deleted)?} {If so,
    please describe these limits and explain how they will be
    enforced.}

Yes, we will delete instances upon request.

\dsquestionex{Will older versions of the dataset continue to be
    supported/hosted/maintained?} {If so, please describe how. If not,
  please describe how its obsolescence will be communicated to \edit{dataset
  consumers}.}

The older versions of the dataset will continue to be hosted on Google Drive. They will remain accessible through the project's webpage where updates and newer versions will also be posted.

\dsquestionex{If others want to extend/augment/build on/contribute to
    the dataset, is there a mechanism for them to do so?} {If so,
  please provide a description. Will these contributions be
  validated/verified? If so, please describe how. If not, why not? Is
  there a process for communicating/distributing these contributions
  to \edit{dataset consumers}? If so, please provide a description.}

We welcome contributions and ideas from others who wish to extend, augment, or build upon our dataset. Interested parties can reach out to us via email to discuss their ideas.

\twocolumn

\begin{table*}
    \small
    \setlength{\tabcolsep}{3pt}
    \centering
    \resizebox{1.0\linewidth}{!}{
    
    \begin{tabular}{@{}m{6cm}|m{12cm}@{}}
    	\toprule
                  & Prompt \\ 
                \midrule
                Prompting LLaVA-1.6-Mistral-7B: Obtaining Objects+Composition+Activities $d_{n,i}$  &  \lstinline{Describe the objects relationships in the photo.} \\
                \midrule
                Prompting Mistral-7B-Instruct-v0.2: Obtaining Objects $o_{n,i}$ & 
\lstinline{<s>[INST] You are an intelligent chatbot designed to extract requested information from the textual description of an image. I will give you a textual description of the image. List ALL objects visible in the image. An object is anything that has a fixed shape or form, that you can touch or see. Name each object with one noun or a maximum of two words. Skip uncertain objects. The textual description of the image: "<INPUT TEXTUAL DESCRIPTION>" DO NOT PROVIDE ANY EXTRA INFORMATION ABOUT OBJECT PROPERTIES OR RELATIONSHIPS TO OTHER OBJECTS IN PARENTHESES. DO NOT PROVIDE ANY OTHER OUTPUT TEXT OR EXPLANATION. [/INST] Comprehensive enumerated list of objects:} \\
\midrule
Prompting Mistral-7B-Instruct-v0.2: Obtaining Activities $a_{n,i}$ & 
\lstinline{<s>[INST] You are an intelligent chatbot designed to extract requested information from the textual description of an image. I will give you a textual description of the image. List all VISIBLE activities in the image. Activity is lively action or movement. Name each activity with a concise phrase SKIP possible or implied activities that are not visible. If no activity is visible, reply "No activity is visible." DO NOT PROVIDE ANY OTHER OUTPUT TEXT OR EXPLANATION. The textual description of the image: "<INPUT TEXTUAL DESCRIPTION>" [/INST] Comprehensive enumerated list of activities:} \\
\midrule
Prompting Mistral-7B-Instruct-v0.2: Obtaining Verbs $\nu_{n,i}$ & 
\lstinline{<s>[INST] You are an intelligent chatbot designed to extract requested information from the textual description of an image. I will give you a list of visible activities of the image. Your task is to delete information about objects from this description. Replace all objects in this list with "someone" or "something," but keep the activity. If you have to, you may delete some details, but delete ALL object information. If the input is "No activity is visible.", keep it "No activity is visible." DO NOT PROVIDE ANY OTHER OUTPUT TEXT OR EXPLANATION. The list of visible activities: "<INPUT ACTIVITIES DESCRIPTION>" [/INST] Post-processed enumerated list of activities:} \\
\midrule
Prompting Mistral-7B-Instruct-v0.2: Obtaining 15-words Summaries $d'_{n,i}$ & 
\lstinline{<s>[INST] You are an intelligent chatbot designed to extract requested information from the textual description of an image. Summarize the following image description in 15 words: "<INPUT TEXTUAL DESCRIPTION>" [/INST] 15-words summary:} \\
         	\arrayrulecolor{black}\bottomrule
    \end{tabular}
    }
    \caption{Prompts used in our UTD method to obtain textual descriptions of frames with respect to different concepts categories: objects+composition+activities $d_{n,i}$, objects $o_{n,i}$, activities $a_{n,i}$, and verbs $\nu_{n,i}$. \label{tab:prompts_concepts}
    }
\end{table*} 
\begin{table*}
    \small
    \setlength{\tabcolsep}{3pt}
    \centering
    \resizebox{1.0\linewidth}{!}{
    \begin{tabular}{@{}m{4.6cm}|m{2.6cm}|m{11cm}@{}}
    	\toprule
                 Textual description & Setup & \lstinline{<instruction>} \\ 
                \midrule
                \multirow{3}{*}{objects+composition+activities $d_{n,i}$} &  single-frame  & \lstinline{Given a video frame description, retrieve the activity depicted in this video.} \\
                 \cmidrule{2-3}
                & sequence-of-frames & \lstinline{Given descriptions of video frames, retrieve the activity depicted in this video.} \\
                \midrule
                \multirow{3}{*}{objects $o_{n,i}$} &  single-frame  & \lstinline{Given a list of objects visible on the video frame, retrieve the activity depicted in this video.} \\
                 \cmidrule{2-3}
                & sequence-of-frames & \lstinline{Given lists of objects visible on the video frames, retrieve the activity depicted in this video.} \\
                \midrule
                \multirow{3}{*}{activities $a_{n,i}$ / verbs $\nu_{n,i}$ } &  single-frame  & \lstinline{Given a description of actions visible on the video frame, retrieve the activity depicted in this video.} \\
                 \cmidrule{2-3}
                & sequence-of-frames & \lstinline{Given a description of actions visible on the video frames, retrieve the activity depicted in this video.} \\
                \midrule
                \multirow{3}{*}{activity class name} &  single-frame  & \lstinline{Given an activity, retrieve a video frame description that may depict this activity.} \\
                 \cmidrule{2-3}
                & sequence-of-frames & \lstinline{Given an activity, retrieve a video description that may depict this activity.} \\
         	\arrayrulecolor{black}\bottomrule
    \end{tabular}
    }
    \caption{Instructions used to prompt the SFR-Embedding-Mistral model for action classification. \label{tab:prompt_classification}}
\end{table*} 

\begin{table*}
    \small
    \setlength{\tabcolsep}{3pt}
    \centering
    \resizebox{1.0\linewidth}{!}{
    \begin{tabular}{@{}m{4.6cm}|m{2.6cm}|m{11cm}@{}}
    	\toprule
                 Textual description & Setup & \lstinline{<instruction>} \\ 
                \midrule
                \multirow{3}{*}{objects+composition+activities $d_{n,i}$} &  single-frame  & \lstinline{Given a description of a single video frame, retrieve a short description of the full video.} \\
                 \cmidrule{2-3}
                
                & sequence-of-frames & \lstinline{Given descriptions of video frames, retrieve a short description of the full video.} \\
                \midrule
                \multirow{3}{*}{objects $o_{n,i}$} &  single-frame  & \lstinline{Given a list of objects visible on the video frame, retrieve a short video description.} \\
                 \cmidrule{2-3}
                
                & sequence-of-frames & \lstinline{Given lists of objects visible on the video frames, retrieve a short video description.} \\
                \midrule
                \multirow{3}{*}{activities $a_{n,i}$ / verbs $\nu_{n,i}$} &  single-frame  & \lstinline{Given a description of actions visible on the video frame, retrieve a short video description.} \\
                 \cmidrule{2-3}
                
                & sequence-of-frames & \lstinline{Given a description of actions visible on the video frames, retrieve a short video description.} \\
                \midrule
                \multirow{2}{*}{caption (from dataset)} &  single-frame  & \lstinline{Given a short video description, retrieve a description of a specific frame within that video.} \\
                 \cmidrule{2-3}
                & sequence-of-frames & \lstinline{Given a short video description, retrieve another description of this video.} \\
         	\arrayrulecolor{black}\bottomrule
    \end{tabular}
    }
    \caption{Instructions used to prompt the SFR-Embedding-Mistral model for text-to-video retrieval. \label{tab:prompt_retrieval}}
\end{table*}

\begin{table*}
    \small
    \setlength{\tabcolsep}{3pt}
    \centering
    \resizebox{1.0\linewidth}{!}{
    \begin{tabular}{@{}m{2cm}|m{4.2cm}|m{11cm}@{}}
    	\toprule
                 Task & Textual Description & \lstinline{<instruction>} \\ 
                \midrule
                \multirow{13}{*}{retrieval} &  \multirow{6}{*}{objects $o_n$ (seq-of-frames setup) }   & \lstinline{Given lists of objects visible on the video frames, retrieve a short video description.} \\
                \cmidrule{3-3}
                & & \lstinline{Using lists of objects seen in video frames, retrieve a brief description of the video.} \\
                \cmidrule{3-3}
                & & \lstinline{From lists of objects present in video frames, retrieve a concise video description.} \\
                \cmidrule{2-3}
                &  \multirow{6}{*}{caption (from dataset)}  & \lstinline{Given a short video description, retrieve another description of this video.} \\
                \cmidrule{3-3}
                & & \lstinline{Use a brief video description as a query to retrieve an alternative description of the same video.} \\
                \cmidrule{3-3}
                & & \lstinline{Given a concise video description, retrieve another description for that video.} \\
                \midrule
                \multirow{5}{*}{classification} &  \multirow{6}{*}{objects $o_n$ (seq-of-frames setup)}   & \lstinline{Given lists of objects visible on the video frames, retrieve the activity depicted in this video.} \\
                \cmidrule{3-3}
                & & \lstinline{Using lists of objects seen in video frames, retrieve the activity captured in the video.} \\
                \cmidrule{3-3}
                & & \lstinline{From lists of objects present in video frames, retrieve the activity that the video shows.} \\
         	\arrayrulecolor{black}\bottomrule
    \end{tabular}
    }
    \caption{Multiple instructions for robust datasets unbiasing. We prompt the text embedding models using three different prompts for captions and object textual descriptions, generating three different embeddings for each. During the unbiasing process, we exclude samples from the test sets only if the sample is correctly classified or retrieved using any combination of these embeddings. \label{tab:instructions_unbiasing} }
\end{table*} 


\end{document}